%% file: neurips_2026.tex
\documentclass{article}
\PassOptionsToPackage{numbers}{natbib}

 \usepackage[main, final]{neurips_2026}

\usepackage[utf8]{inputenc}
\usepackage[T1]{fontenc}
\usepackage{hyperref}
\usepackage{url}
\usepackage{booktabs}
\usepackage{amsfonts}
\usepackage{nicefrac}
\usepackage{microtype}
\usepackage[dvipsnames,table]{xcolor}
\usepackage{graphicx}
\usepackage{subcaption}
\usepackage{wrapfig}

\usepackage{enumitem}
\usepackage{amsmath}
\usepackage{amssymb}

\usepackage{algorithm}
\usepackage{algpseudocode}
\usepackage{amsmath}
\usepackage{amssymb}

\title{Self-Supervised Keyframe Discovery for Horizon-Invariant Behavior Cloning}

\author{
  Prabin Kumar Rath \\
  Arizona State University \\
  \texttt{prath4@asu.edu}
  \And
  Omkar Patil \\
  Arizona State University \\
  \texttt{opatil3@asu.edu}
  \And
  Nakul Gopalan \\
  Arizona State University \\
  \texttt{ng@asu.edu}
}

\begin{document}

\maketitle

\begin{abstract}
Behavior cloning (BC) in non-Markovian environments is a challenging problem because policies have to reason over contextual information over long horizons.
Existing policy architectures rely on recurrent or attention-based mechanisms to capture long-term dependencies.
However, recurrent models suffer from hidden-state collapse and gradient instability under backpropagation through time while attention-based models are fundamentally limited by context length.
To address these issues, we propose Keyframe Mnemonics, a novel self-supervised method that \textit{discovers} a set of information-critical observations (\textit{mnemonics}) by learning an objective from randomly sampled past observations and using it as a reward for keyframe selection. We then train a BC policy that conditions on the discovered keyframes to model the action distribution.
Under the simplicity bias assumption~\cite{valle2018deep, mingard2025deep}, our formulation provides context retention guarantees over an infinite horizon, while maintaining a small set of decision-relevant keyframes in the policy's working memory.
We evaluate our method on synthetic memory domains, where mnemonic-conditioned BC policies achieve $100\%$ success rates (SR) and generalize to horizons orders of magnitude beyond training without performance degradation. Moreover, we evaluate on a memory-intensive robot manipulation benchmark, achieving a $13.9\%$ average absolute SR improvement over the strongest baseline across $23$ tasks. Finally, we demonstrate generalization with $80\%$ SR at $20\times$ longer time horizons on a real robot. Code and videos available at \url{https://keyframe-mnemonics.github.io}.
\end{abstract}

\section{Introduction}
The Markov assumption for decision processes is that the agent’s current state contains sufficient information to determine the optimal action with respect to its goal or objective \cite{puterman2014markov}. In practice, however, real-world environments are non-Markovian and partially observable, where decision-making often requires retaining (``remembering'') information over time to provide context for future actions \cite{kaelbling1998planning, lauri2022partially}. Recent advancements in policy learning have increasingly relied on behavior cloning (BC) \cite{zhao2023learning, chi2023diffusionpolicy, kawaharazuka2025vla-survey}, in which policies are highly dependent on short observation contexts with a working memory spanning only a few seconds into the past. This makes such BC policies brittle on tasks that require remembering information cues over long horizons \cite{cherepanov2025memory}. Recurrent models have been the architecture of choice for behavior cloning in such memory intensive partially observable environments and are typically trained using Back-Propagation Through Time (BPTT) \cite{werbos2002backpropagation, chevalier2018babyai}. However, BPTT is difficult to scale, as it requires sequential gradient accumulation over episode timesteps \cite{pascanu2013difficulty}.
Truncated-BPTT tries to mitigate this issue but trades off information from distant past observations \cite{tallec2017unbiasing}. Moreover, recurrent models suffer from hidden-state collapse due to repeated representation updates over time \cite{hochreiter2001gradient, pascanu2013difficulty}. Attention-based architectures such as transformers \cite{vaswani2017attention} address many representation learning challenges, but their finite context length limits retention of the full history in long horizon domains. Furthermore, increasing context through frame stacking incurs quadratic compute complexity \cite{vaswani2017attention}, degrades at extreme horizons \cite{cherepanov2025elmur}, and is prone to overfitting spurious correlations with visual observations \cite{de2019causal}. Many hybrid architectures have been proposed that improve information retention through dedicated memory mechanisms \cite{cherepanov2025elmur, fang2025sam2act, koo2025hamlet, jang2025contextvla, shi2025memoryvla, chung2026rethinking, tan2026memoact} or by learning recurrent summaries \cite{sorokin2022explain, zhou2025mtil, lei2026vpwem}. However, these methods remain susceptible to state collapse beyond the horizons encountered during training, often leading to severe performance degradation on tasks that require generalization to longer horizons.

\begin{figure}
  \centering
  \includegraphics[width=\linewidth]{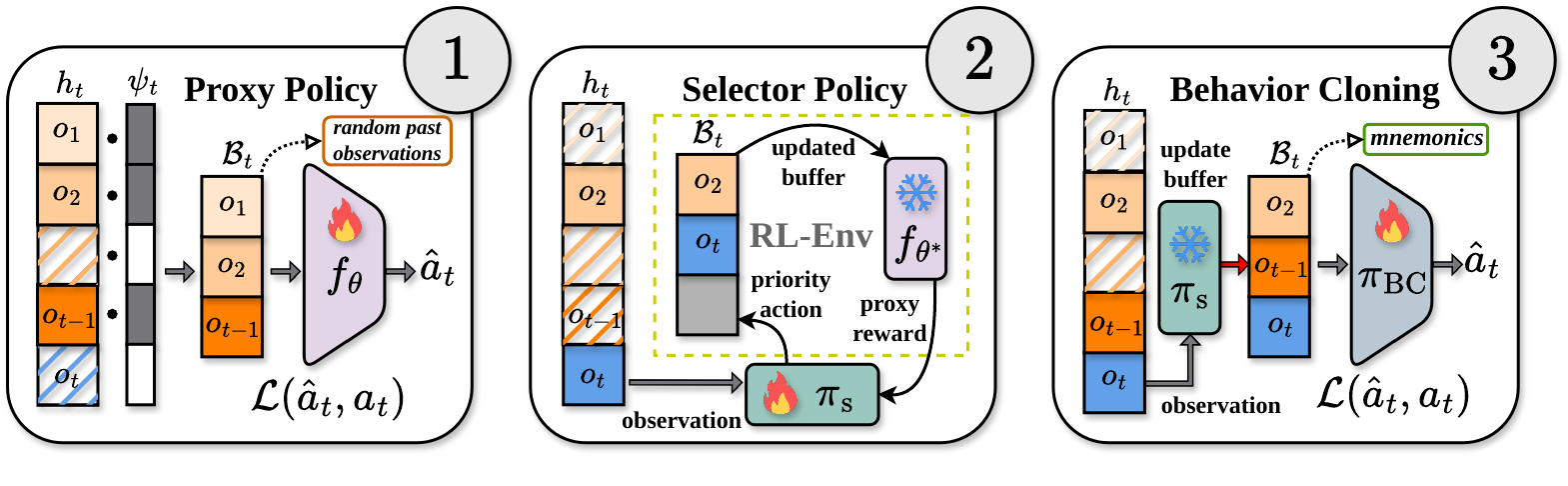}
  \caption{\textbf{Keyframe Mnemonics three-stage training for horizon-invariant behavior cloning.}
  \textbf{(1)~Proxy policy training.} A random binary mask $\Psi_k$ is
  applied to the observation history $h_t$ to assemble a memory buffer
  $\mathcal{B}_t$ of size at most $k$, on which the proxy
  $f_\theta:\mathcal{B}\!\to\!\mathcal{A}$ is trained to predict the
  expert action $a_t$ under a negative log-likelihood loss
  $\mathcal{L}$. Combinatorial support of training data (Lemma 1) biases $f_\theta$ toward keyframe-based hypotheses
  (Theorem~1, Proposition~1), so the frozen objective
  $J_{\theta^{\ast}}(\mathcal{B}_t) = \mathcal L(f_{\theta^*}(\mathcal B_t), a_t)$ scores keyframe-rich buffers
  favourably (Corollary~1; Section~\ref{sec:keyframe_discovery}).
  \textbf{(2)~Selector policy training.} The selector $\pi_\mathrm{s}$ is trained using PPO to populate $\mathcal{B}_t$ with keyframes using $J_{\theta^{\ast}}(\mathcal{B}_t)$ as reward function
  (Section~\ref{sec:selector_optimization}).
  \textbf{(3)~Behavior cloning.} The trained selector is used to construct a keyframe augmented dataset $\mathcal{D}^{\dagger}$, which is then used to train the BC policy $\pi_\mathrm{BC}(a_t \mid o_t,\mathcal{B}_t)$ (Section \ref{sec:behavior_cloning}).
  Throughout the figure, blue tiles denote the
  current observation $o_t$, orange tiles denote past frames in $h_t$,
  hatched tiles are frames not selected into $\mathcal{B}_t$.}
  \label{fig:method}
\end{figure}

To address these challenges in behavior cloning, we propose \textbf{Keyframe Mnemonics (KM)}, a policy learning
method that decomposes the original non-Markovian decision-making problem into two subproblems: (i) discovering observations (\textit{keyframes}) that contain information relevant to the action distribution, and (ii) training a BC policy on demonstrations augmented with the identified keyframes. Our keyframe discovery process is self-supervised and leverages the intrinsic simplicity bias of neural networks~\cite{valle2018deep, mingard2025deep, arpit2017closer} to train a reward function that assigns high rewards to information-rich keyframes in the policy's context buffer. The reward learning process is driven by the mutual information advantage of keyframes with the expert action distribution, which is amplified through a novel training method that uses randomly sampled past observations as training samples for the reward function. We then train a selector policy using reinforcement learning to optimize the learned reward function and identify a set of decision-relevant keyframes, which we call \textit{Mnemonics} (``something that assists memory'').
Finally, we train BC policies conditioned on these mnemonics to match the expert action distribution.
The trained policy is deployed together with the selector, which identifies keyframes at runtime and provides context to disambiguate actions under partial observability. For environments with sparse keyframes, this architectural decomposition enables \textit{horizon-invariance} i.e., memory retention beyond the training horizon, allowing BC policies to generalize on longer time scales where existing hybrid memory methods fail. We extensively evaluate Keyframe Mnemonics on synthetic memory benchmarks (Add~\cite{hochreiter1997long}, Scattered-Copy~\cite{sorokin2022explain}, T-Maze~\cite{bakker2001reinforcement}), grid worlds (MiniGrid Memory~\cite{chevalier2023minigrid}, LTMB-Hallway~\cite{yue2024learning}), memory-intensive robot manipulation benchmark MIKASA-Robo~\cite{cherepanov2025memory}, and additionally on a real robot, demonstrating behavior generalization without performance degradation far beyond the training horizon.
Our contributions are:
\begin{itemize}[leftmargin=2em, topsep=0pt, itemsep=2pt]
    \item We propose a novel self-supervised algorithm to discover decision-relevant keyframes in non-Markovian expert demonstrations without explicit keyframe annotations (Figure \ref{fig:method}, Algorithm~\ref{alg:masked-sampling}).

    \item We provide a theoretical intuition of our method for environments with sparse keyframes, derive information retention guarantees, and support our hypotheses with empirical evidence.

    \item We demonstrate horizon-invariant performance of mnemonic-conditioned BC policies, with MLP and CNN policies achieving $100\%$ SR in synthetic and grid memory domains, and Transformer policies improving absolute SR by $13.9\%$ on the MIKASA-Robo benchmark. Additionally, we show a flow-matching policy retaining $80\%$ SR at $20\times$ longer test timesteps on a real robot.

\end{itemize}

\section{Methodology}
\label{sec:problem_setup}
We assume a partially observable Markov decision process (POMDP) \cite{kaelbling1998planning} $\mathcal{M} = (\mathcal{S}, \mathcal{A}, \mathcal{O}, \mathcal{T}, \Omega, \rho_0)$, where $\rho_0 \in \Delta(\mathcal{S})$ denotes the initial state distribution. At timestep $t \in \{1, \ldots, T\}$ with horizon $T \in \mathbb{N}^+$, the unobserved state is denoted by $s_t \in \mathcal{S}$, the agent receives an observation $o_t \in \mathcal{O}$, and takes an action $a_t \in \mathcal{A}$ based on a belief over the state, $b_t = P(s_t \mid h_t)$, where $h_t = \{o_1, \ldots, o_t\}$ denotes the history of observations up to time $t$. Here, $\mathcal{T}(s_{t+1} \mid s_t, a_t)$ and $\Omega(o_{t+1} \mid s_{t+1}, a_t)$ denote the transition dynamics and observation function respectively. We omit the reward function $\mathcal{R}$ and discount factor $\gamma$ from the notation, as they are not available in standard behavior cloning. Additionally, we assume access to an expert demonstration dataset $\mathcal{D} = \{\tau_i\}_{i=1}^N$, where each trajectory is given by $\tau_i = \{(o_t, a_t)\}_{t=1}^{T_i}$. We make the following assumptions over the dataset (Assumptions 1 and 2) and training dynamics for gradient descent  (Assumptions 3 and 4), where $I(\cdot;\cdot)$ and $H(\cdot)$ denote mutual information and entropy, respectively.

\textbf{Assumption 1 (Approximate belief sufficiency).} The trajectory observations admit a partition into two disjoint subsets: a sparse subset $\mathcal{X}_{\mathrm{key}}$ of size $C \ll T$ (\textit{keyframes}) and the remaining $T - C$ observations $\mathcal{X}_{\mathrm{bg}}$ (\textit{background frames}). The keyframes $\mathcal{X}_{\mathrm{key},t} \subset h_t$ up to time $t$ contain observations with invariant task information and causal cues that remain consistent across the demonstration dataset, thereby forming an approximate sufficient statistic for the belief state $b_t$. In contrast, the background subset $\mathcal{X}_{\mathrm{bg},t} \subset h_t$ contributes minimal information toward the belief state. This assumption imposes a sparse keyframe structure on observations in the demonstration dataset, which we empirically observe in several memory intensive partially observable domains evaluated in Section~\ref{sec:experiments}.

\textbf{Assumption 2 (Action disambiguation by keyframes).} Any ambiguity in the action distribution is primarily captured by the keyframes $\mathcal{X}_{\mathrm{key}}$, with the following relationship in the mutual information $I(a_t ; \mathcal{X}_{\mathrm{key},t}) > I(a_t ; \mathcal{X}_{\mathrm{bg},t})$. That is, keyframes strictly provide more information for disambiguating the action than the background frames. Section \ref{sec:experiments} provides empirical evidence consistent with this assumption in the evaluated memory-intensive POMDPs.

\textbf{Assumption 3 (Realizability of conditional programs by interpolating hypotheses).} We combine the standard interpretation of conditional Kolmogorov
complexity as the shortest program mapping inputs to outputs
\cite{grunwald2008algorithmic,li2008introduction} with the finite-sample
expressivity of overparameterized neural networks, which can interpolate or
memorize arbitrary training labels
\cite{zhang2016understanding, arpit2017closer, yun2019small}. (Details in Appendix~\ref{appendix:proof}).

\textbf{Assumption 4 (Simplicity-biased SGD approximation).} We assume SGD is
biased towards selecting Kolmogorov simple functions, where low-complexity functions receive larger prior mass~\cite{valle2018deep, mingard2025deep, mingard2021sgd}. (Details in Appendix~\ref{appendix:proof}).

We formally define \textit{Mnemonics} as the set of keyframes up to time $t$, denoted by $\mathcal{X}_{\mathrm{key},t} \subset h_t$, which provide sufficient information for a behavior cloning policy to generate actions unambiguously. Our goal is to learn a selector policy $\pi_\mathrm{s}$ that identifies keyframes from a continuous stream of observations and adds them to a memory buffer $\mathcal{B}_t$, maximizing its overlap with $\mathcal{X}_{\mathrm{key},t}$. These mnemonics are then used to condition a behavior cloning policy $\pi_{\mathrm{BC}}$ that models the expert action distribution.

\subsection{Leveraging Simplicity-bias for Keyframe Discovery}
\label{sec:keyframe_discovery}
We formulate keyframe \textit{discovery} as a self-supervised buffer optimization problem. Specifically, we define an objective $J(\cdot,f)$ over a proxy policy $f$ mapping observation frames in the buffer, which collectively represent the belief state, to expert actions, $f: \mathcal{B} \rightarrow \mathcal{A}$, and optimize the buffer $\mathcal{B}$ subject to a memory budget $k$ much smaller than the observation history ($k \ll T$).
For
$J(\cdot,f)$ to serve as an effective objective, it must provide a stronger learning signal (i.e., lower objective value) when $\mathcal{B}$ contains a higher proportion of keyframes than background frames. In this section, we introduce a training strategy for the proxy policy $f$ that increases the complexity of the input--output mapping, thereby exploiting the simplicity bias of deep neural networks to encourage learning a hypothesis that favors keyframes over background frames due to their greater informational content. The resulting biased model can then be used to learn a selector $\pi_\mathrm{s}$ for keyframe discovery.

\textbf{Proxy Policy Training:} To train the proxy policy $f_\theta$, a neural network parameterized by $\theta$, we construct input buffers using a random binary mask $\Psi_k \in \{0,1\}^{|h_t|}$ that independently selects past observations from the history $h_t$, with cardinality bounded by the memory budget $k$. We call this \textit{masked-sampling} past observations that constructs the training-time memory buffer $\mathcal{B}_t = \{ o_i \in h_t \mid \Psi_k(i) = 1 \}$, which serves as input to the proxy policy $f_\theta : \mathcal{B} \rightarrow \mathcal{A}$ with $|\mathcal{B}_t| \le k$, where $\mathcal{B}$ denotes the space of all possible memory buffers. Algorithm~\ref{alg:masked-sampling} shows the masked-sampling method for training the proxy policy $f_\theta$. Treating $(\mathcal{B}_t, a_t)$ as i.i.d.\ samples, we now analyze the hypothesis space $\mathcal{H}$ explored by $f_\theta$ during training through the lens of Kolmogorov complexity \cite{li2008introduction}, denoted $K(f)$ for $f \in \mathcal{H}$ \footnote{For notational simplicity, we use the same variable $f$ to denote both the proxy policy and the hypothesis learned by the proxy policy, and use Kolmogorov complexity as a measure of hypothesis description length \cite{grunwald2007minimum}.
}. Let $\mathcal{H}^* \subset \mathcal{H}$ denote the set of interpolating hypotheses achieving zero empirical risk on the training dataset induced by masked-sampling. Assume $\mathcal{H}^*$ partitions into an invariant class
$\mathcal{H}_{\mathrm{in}}$, corresponding to hypotheses conditioned only on keyframes, and a non-invariant class $\mathcal{H}_{\neg \mathrm{in}}$, which depend on background frames. Let $f_{\mathrm{key}} \in \mathcal{H}_{\mathrm{in}}$ denote keyframe-based and $f_{\mathrm{bg}} \in \mathcal{H}_{\neg\mathrm{in}}$ denote background-based hypotheses defined by mappings $f_{\mathrm{key}} : \mathcal{X}_{\mathrm{key}} \rightarrow \mathcal{A}$ and $f_{\mathrm{bg}} : \mathcal{X}_{\mathrm{bg}} \rightarrow \mathcal{A}$, where the inputs are constrained to at most $k$ frames, consistent with $f_\theta$.

\begin{algorithm}[t]
\caption{Masked-sampling based training of the proxy policy $f_\theta$}
\label{alg:masked-sampling}
\begin{algorithmic}[1]

\Require dataset $\mathcal{D}$, training memory budget $k$, proxy policy $f_\theta$
\For{each training step}
  \State Sample $\tau \sim \mathcal{D}$ and timestep $t \in \{1,\ldots,|\tau|\}$ \Comment{uniform sampling of timestep}
  \State $n \gets \min\!\big(\mathrm{Binomial}(t,\, 0.5),\; k\big)$ \Comment{buffer occupancy under budget $k$}
  \State Draw $\mathcal{I} \subset \{1,\ldots,t\}$ with $|\mathcal{I}|=n$ \Comment{uniform sampling without replacement}
  \State $\Psi_k(i) \gets \mathbf{1}[\,i \in \mathcal{I}\,]$ for $i=1,\ldots,t$ \Comment{induced binary mask on $h_t$}
  \State $\mathcal{B}_t \gets \{\, o_i \in h_t \mid \Psi_k(i)=1 \,\}$ ordered by $i$ \Comment{$|\mathcal{B}_t| \le k$}
  \State $\hat{a}_t \gets f_\theta(\mathcal{B}_t)$ \Comment{predict expert action}
  \State $\theta \gets \theta - \eta\, \nabla_\theta\, \mathcal{L}(\hat{a}_t,\, a_t)$ \Comment{update proxy using NLL loss}
\EndFor
\end{algorithmic}
\end{algorithm}

\textbf{Overview:} First, we show that masked-sampling of past observations during training of $f_\theta$ induces combinatorially large support of training pairs (Lemma 1), which artificially increases the complexity of background-based hypotheses $f_{\mathrm{bg}}$ and effectively restricts the feasible hypothesis space (Theorem 1). Assuming that deep neural networks exhibit an implicit bias toward simpler hypotheses, we show that keyframe-based hypotheses $f_{\mathrm{key}}$ are learned with high probability (Proposition 1). Finally, we demonstrate that the prediction error of the trained proxy policy provides an optimizable signal to identify keyframe observations (Corollary 1).

\textbf{Lemma 1 (Combinatorial cardinality of the sample space).} \emph{The total cardinality $M$ of the space of distinct timestep-indexed training pairs
$(\mathcal{B}_t, a_t)$ is given by:}
\begin{equation}
M = \sum_{i=1}^N \left( T_i + \sum_{j=1}^k \binom{T_i + 1}{j + 1} \right).
\end{equation}

Masked-sampling induces a combinatorially large finite support of training pairs, whose cardinality $M$ admits a closed-form expression via the Hockey-stick identity \cite{pascal1978traite}. (Proof in Appendix~\ref{appendix:proof}). \hfill $\square$

\textbf{Theorem 1 (Expected gap in hypothesis complexity).} \emph{Under the assumption that training pairs are sampled i.i.d.\ from a sufficiently large support of size $M$ (Lemma~1), the expected Kolmogorov complexity of the minimum-complexity background-based hypothesis $f^*_{\mathrm{bg}}$ is strictly greater than that of the keyframe-based hypothesis $f^*_{\mathrm{key}}$.}
\begin{equation}
    \mathbb{E}[K(f_{\mathrm{bg}}^*)] > \mathbb{E}[K(f_{\mathrm{key}}^*)].
\end{equation}
Using results from \cite{grunwald2008algorithmic, li2008introduction}, we show that, under the stated assumptions, the complexity gap between background-based and keyframe-based hypotheses grows linearly with the masking-induced support size $M$ according to $M \big( I(A ; \mathcal{X}_{\mathrm{key}}) - I(A ; \mathcal{X}_{\mathrm{bg}}) \big)$. Thus, any information advantage of keyframes is amplified by the large support size $M$, rendering background-based hypotheses intractable for the proxy policy $f_\theta$. (Proof in Appendix~\ref{appendix:proof}). \hfill $\square$

\textbf{Proposition 1 (Conditional simplicity-bias preference).}
\emph{Under the simplicity-bias assumption and the complexity gap established in
Theorem 1, Stochastic Gradient Descent (SGD) is biased toward the invariant hypothesis class $\mathcal H_{\mathrm{in}}$. Given a masking-induced sample distribution with support size $M$ (Lemma 1), the probability of selecting an invariant hypothesis is approximated by}
\begin{equation}
P_{\mathrm{SGD}}(\mathcal H_{\mathrm{in}} \mid \mathcal D)
\approx
\frac{1}{
1 + 2^{-\left[
M \big(
I(A ; \mathcal X_{\mathrm{key}})
-
I(A ; \mathcal X_{\mathrm{bg}})
\big)
-
\mathcal O(1)
\right]}
}.
\end{equation}

Neural networks and SGD are biased toward Kolmogorov-simple functions \cite{valle2018deep, mingard2025deep, mingard2021sgd}. Hence, the positive hypothesis complexity gap from Theorem 1 drives $P_{\mathrm{SGD}}(\mathcal{H}_{\mathrm{in}} \mid \mathcal{D})$ close to $1$, biasing $f_\theta$ toward learning an invariant keyframe-based hypothesis. (Proof in Appendix~\ref{appendix:proof}). \hfill $\square$

\textbf{Corollary 1 (Keyframe recovery from the frozen loss objective).} \emph{
Let us define a subset-selection objective function
\begin{equation}
J_{\theta^*}(\mathcal B)
=
\mathbb E\!\left[
\mathcal L(f_{\theta^*}(\mathcal B), A)
\right]
\end{equation}
over a frozen predictor $f_{\theta^*}$ obtained under Proposition~1 after training with negative log-likelihood loss $\mathcal L$, where $f_{\theta^*}$ represents a near-optimal invariant hypothesis in $\mathcal H_{\mathrm{in}}$. Consider candidate buffer subsets $\mathcal C_k=\{\mathcal B_t\subseteq h_t:|\mathcal B_t|\le k\}$, and let $\mathcal B_{\mathrm{opt}} \in \arg\min_{\mathcal B\in\mathcal C_k} J_{\theta^*}(\mathcal B)$. If keyframe subsets $\mathcal B \subset \mathcal X_{\mathrm{key}}$ have a positive action-information margin over all non-keyframe candidate subsets, and this margin dominates the approximation error of the frozen predictor $f_{\theta^*}$, then the loss objective $J_{\theta^*}(\mathcal B)$ recovers the keyframes up to information-equivalent subsets. In particular, if $\mathcal X_{\mathrm{key}}$ is unique under the memory budget $k$, then $\mathcal B_{\mathrm{opt}}=\mathcal X_{\mathrm{key}}$.
}

Optimizing $J_{\theta^*}(\mathcal B)$ is equivalent to minimizing the information gap $I(A;\mathcal X_{\mathrm{key}}) - I(A;\mathcal B)$ over $\mathcal C_k$. Under Assumption 2, this drives $\mathcal B$ toward $\mathcal X_{\mathrm{key}}$. (Proof in Appendix~\ref{appendix:proof}). \hfill $\square$

\textbf{The insight of our approach is that by leveraging the natural simplicity bias of neural networks, we can automatically discover informative keyframes by partitioning a dataset based on the mutual information gap between observations and the action distribution.}
Our method is analogous to masked data augmentation~\cite{devlin2019bert, he2022masked} but along the temporal axis, where our masking strategy increases the difficulty of learning spurious correlations and biases the model toward invariant temporal mappings. Theorem~1 provides theoretical intuition for this phenomenon; see Section~\ref{sec:empirical_analysis}, where we provide empirical evidence across synthetic benchmark domains.

\subsection{Selector Optimization for Keyframe Identification}
\label{sec:selector_optimization}
Given the learned objective function, we train a selector policy $\pi_\mathrm{s}$ to identify keyframes. The selector serves two purposes: (i) extract keyframes from the demonstration dataset for constructing mnemonics that can be used to train a downstream policy $\pi_\mathrm{BC}$, and (ii) identify keyframes from a continuous stream of observations during inference to populate the memory buffer for $\pi_\mathrm{BC}$. However, note that the objective $J_{\theta^*}(\mathcal{B})$ is defined over buffer inputs rather than individual observations. Furthermore, the number of possible buffer states is combinatorially large (Lemma~1), making exhaustive evaluation non-trivial. To address this, we formulate keyframe identification as a deterministic Markov Decision Process (MDP) over the demonstration dataset, modeling the buffer $\mathcal{B}$ as a priority queue and using a selector policy $\pi_\mathrm{s}$ to assign priority scores that update the buffer. This reformulates the problem from optimizing over buffer states to learning a selector policy that prioritizes keyframe observations.

The selection MDP is defined by the tuple $\langle \mathcal{O}, \mathcal{P}, \mathcal{K}, R, \gamma \rangle$ over offline demonstration trajectories $\tau_i \in \mathcal{D}$, where $\tau_i = \{(o_t, a_t)\}_{t=1}^{T_i}$. At each timestep $t \in \{1, \ldots, T_i\}$, the selector policy $\pi_{\mathrm{s}}(p_t \mid o_t)$ receives an observation $o_t \in \mathcal{O}$ and outputs a priority action $p_t \in \mathcal{P}$, which is used to update the buffer $\mathcal{B}$, modeled as a priority queue with memory budget $k$.
The MDP tuple is specified as follows.
\textbf{State ($\mathcal{O}$)} is the current observation $o_t$ from a trajectory $\tau_i \in \mathcal{D}$. An \textbf{Action ($\mathcal{P}$)} is a priority score $p_t \in [0,1]$. \textbf{Transition ($\mathcal{K}$)}: The transition is deterministic, with the next state given by the subsequent observation $o_{t+1}$ in $\tau_i$. The buffer is updated as $\mathcal{B}_t = \mathcal{B}_{t-1}(o_t, p_t)$ according to an eviction rule that determines whether $o_t$ is inserted into or rejected from the buffer.  We use a \textbf{Discount Factor ($\gamma$)} $= 0.99$. The
     \textbf{Reward ($R$)} function  depends on the updated buffer $\mathcal{B}_t$ and a normalization constant $\lambda \in (0,1)$ (typically a value close to $0$).
        \begin{equation}
        \label{eq:reward}
        R(\mathcal{B}_t) =
        \underbrace{
        T_i \cdot\frac{e^{J^{j=t}_{\theta^*}/\kappa}}{\sum_{j=1}^{T_i} e^{J^j_{\theta^*}/\kappa}}
        }_{\text{surprise}}
        \cdot
        \underbrace{
        \frac{\log\big(J_{\theta^*}(\mathcal{B}_t) + \lambda\big)}{\log\big(\lambda\big)}
        }_{\text{reward}}.
        \end{equation}

We use a normalized logarithmic reward which increases sharply toward $1$ as the objective value approaches $0$, providing a strong optimization signal even for small changes in the memory buffer. This formulation encourages $\pi_\mathrm{s}$ to assign higher priorities to observations that minimize the learned objective. The reward is further scaled by a surprise \cite{sorokin2022explain, wen2021keyframe} coefficient weighted by episode length, which captures action uncertainty. Intuitively, this assigns higher reward weight to timesteps with larger objective values, which corresponds to higher action entropy under the proxy policy. Such timesteps are typically decision-critical, where actions depend strongly on the memory buffer. Hence, emphasizing them focuses optimization on resolving keyframes where they matter most.

The surprise coefficients are precomputed for each trajectory using a temperature ($\kappa$)-scaled softmax over the expected objective values across timesteps. The expected objective values are computed over a few epochs of masked-sampled input buffers by evaluating the objective $J_{\theta^*}(\mathcal B)$ at each timestep in the demonstration dataset $\mathcal{D}$. We optimize $\pi_\mathrm{s}$ using Proximal Policy Optimization (PPO) \cite{schulman2017proximal}, which enables stable training while promoting exploration through entropy regularization. The memory buffer employs different priority strategies during training and inference. During training, a FIFO queue penalizes the selector policy for assigning low priority to observations that have long-term influence on actions, whereas during inference, a LIFO queue preserves high-priority observations, ensuring their retention over long horizons. See Appendix~\ref{app:buffer-queue} for algorithms on priority queue eviction rules used during training and inference.

\subsection{Training BC Policy with Keyframe Mnemonics}
\label{sec:behavior_cloning}
A trained selector policy is used to generate mnemonics, represented by the buffer state $\mathcal{B}_t$ with $|\mathcal{B}_t| \le k$, which approximates the keyframes up to time $t$, i.e., $\mathcal{X}_{\mathrm{key},t} \subset h_t$. We now present a conditional guarantee for context retention in the memory buffer assuming an optimal selector, which enables mnemonics-conditioned BC policies to generalize to an infinite horizon (Theorem~2).

\textbf{Theorem 2 (Infinite-horizon keyframe retention).}
\emph{Assume a POMDP in which the expert behavior strictly satisfies Assumptions~1 and 2. Let $\pi_{\mathrm{s}}^*$ be an optimal selector minimizing the frozen objective $J_{\theta^*}(\mathcal{B}_t)$ (Corollary~1). Suppose a unique keyframe set $\mathcal{X}_{\mathrm{key}}$ exists that has the highest mutual information with the action and fits within the memory budget $k$. Then, as $T \to \infty$, $\pi_{\mathrm{s}}^*$ assigns higher priority to keyframes than to background frames, ensuring that the priority queue retains the keyframe set for the entire horizon $T$ without eviction, i.e., $\mathcal{B}_T = \mathcal{X}_{\mathrm{key}}$.}

An optimal selector minimizes the learned objective $J_{\theta^*}(\mathcal{B}_t)$ at every time step $t$, which corresponds to recovering the keyframe set $\mathcal{X}_{\mathrm{key},t}$ at each $t$. (Proof in Appendix~\ref{appendix:proof}). \hfill $\square$

Ideally, the memory buffer should retain the full context needed to disambiguate actions at each timestep $t$. However, in complex domains such as robot manipulation, our assumptions may not hold strongly, and actions may have some dependency on the background observation set. To address this, we additionally include the current observation $o_t$ and construct an augmented dataset $\mathcal{D}^\dagger$ with trajectories $\tau^\dagger_i = \{(\mathcal{B}_t, o_t, a_t)\}_{t=1}^{T_i}$, which is used to train the BC policy $\pi_\mathrm{BC}(a_t \mid o_t, \mathcal{B}_t)$. See Appendix~\ref{appendix:toy_implementation} and \ref{appendix:robot_implementation} for details on policy training.

\setlength{\parskip}{5.5pt plus 3pt}
\section{Experiments and Result Analysis}
\label{sec:experiments}
In this section, we present experiments validating the effectiveness of the Keyframe Mnemonics (KM) method. We first provide empirical evidence supporting our theoretical analysis of keyframe discovery from Section~\ref{sec:keyframe_discovery}. We then evaluate mnemonic-conditioned BC policies along two central questions: \textbf{(1)} How effectively does the selector identify keyframes across diverse memory benchmarks, ranging from synthetic domains to robot manipulation tasks? \textbf{(2)} Can mnemonic-conditioned BC policies generalize beyond the training horizon and exhibit horizon-invariant behavior cloning?

\begin{figure}[t]
  \centering
  \includegraphics[width=\linewidth]{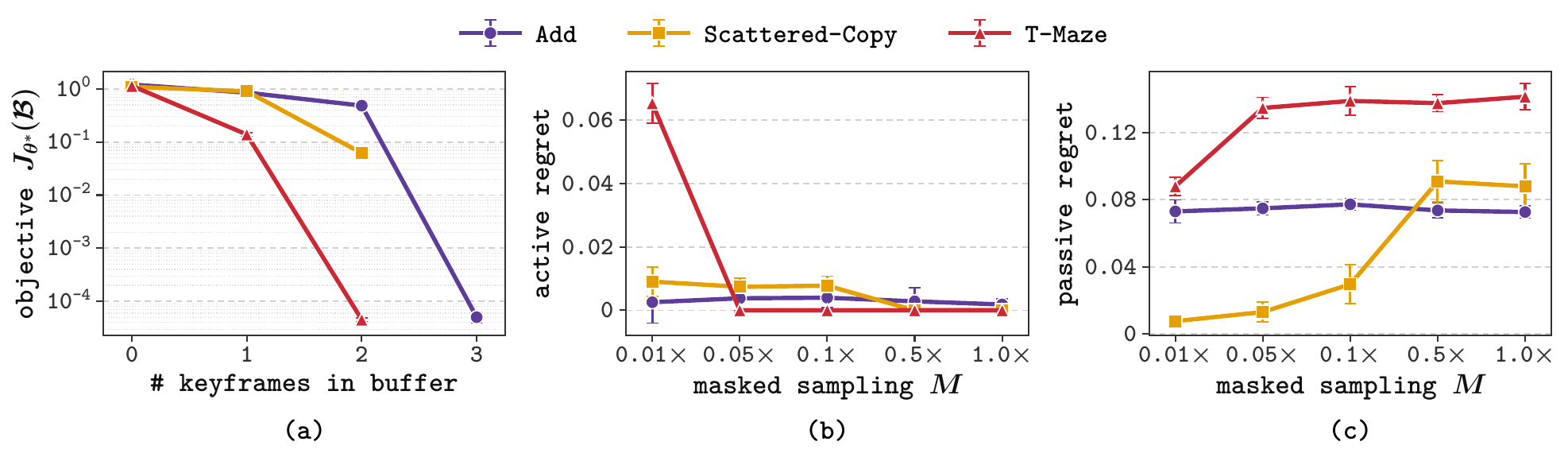}
  \vspace{-10pt}
  \caption{\textbf{Empirical evidence for Keyframe Mnemonics.}
    (a) Adding \textit{oracle} keyframes to the context buffer of the proxy policy leads to decrease in objective value across three problem domains (Corollary~1).
    (b) Active memory regret remains high for small masked-sampling support sizes $M$ and converges to zero as the support size increases (Theorem~1).
    (c) Passive memory regret grows with $M$ as the proxy policy fits the keyframe-based hypothesis. Note: Here, $M=1.0\times$ corresponds to a calibrated value, chosen empirically (Appendix~\ref{appendix:km_training_details}), that is sufficient according to Theorem~1 and yields mnemonics for horizon-invariant BC policies, as shown in Figure~\ref{fig:horizon_all5}.}
  \label{fig:keyframe-empirics}
\end{figure}

\begin{figure*}[t]
    \centering
    \includegraphics[width=\textwidth]{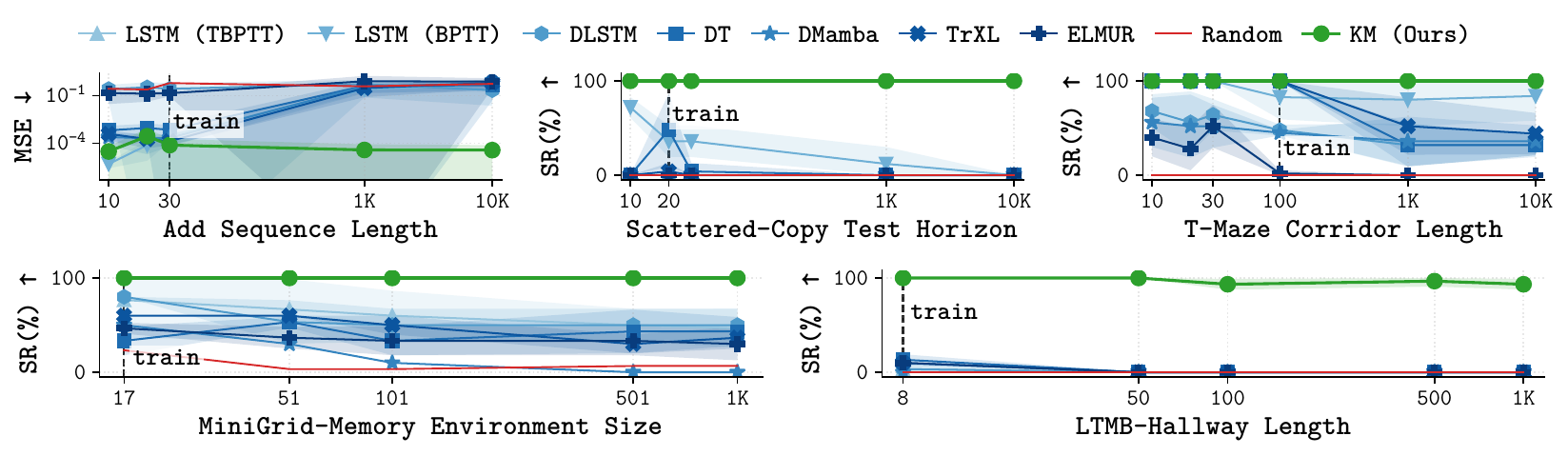}
    \caption{\textbf{Horizon-invariance in synthetic and grid domains.}
    Top row: Add, Scattered-Copy, T-Maze.
    Bottom row: MiniGrid-Memory, LTMB-Hallway.
    Each curve shows mean $\pm$ std over 5 seeds.
    Band width is one std-dev. Vertical dashed line marks the maximum training horizon. Lower is better for MSE; higher is better for Success Rate. KM maintains near-perfect performance across orders of magnitude of horizon scaling. Most baselines degrade outside the training range.}
    \label{fig:horizon_all5}
    \vspace{-8pt}
\end{figure*}

\subsection{Synthetic Domain Benchmarks and Convergence Analysis}
\label{sec:empirical_analysis}
\textbf{Domains.} We evaluate on three synthetic memory sequence-to-sequence tasks and two grid-based environments. These domains satisfy Assumptions~1 and~2, making them suitable for empirical analysis of our method under the stated assumptions.
\textbf{Add}~\cite{hochreiter1997long} requires the policy to emit the sum of two marked floating point numbers from a sequence.
\textbf{Scattered Copy}~\cite{sorokin2022explain} interleaves a small set of $n$ relevant tokens
with noise tokens, and the policy must recover the
relevant tokens in the same order as observed. In \textbf{Active T-Maze}~\cite{bakker2001reinforcement} the agent must remember an oracle observation from the start of a long corridor to make a binary decision at the final junction.
\textbf{MiniGrid-Memory}~\cite{chevalier2023minigrid} is a similar memory task with a more diverse grid-based observation space. \textbf{LTMB-Hallway}~\cite{yue2024learning} is a visual memory benchmark where the agent must match a key observation seen at the episode start to one of several targets uniformly placed along a long hallway. The policy must compare each observed door with the remembered visual cue and select the matching one. See Appendix~\ref{app:synth-env} for details on ground truth keyframes in each domain.

\textbf{Proxy policy objective.} Figure~\ref{fig:keyframe-empirics}(a) shows the learned objective value $J_{\theta^*}(\mathcal{B})$ as a function of the number of keyframes in the input buffer. Across three synthetic tasks, we observe a decrease in the objective value as more keyframes are added to the buffer, supporting Corollary~1.

\textbf{Memory regret metrics.} Let $\mathcal{L}^{\pi}_t$ denote the per-timestep proxy policy loss when the buffer is populated by a selection policy $\pi$. We compare the trained selector $\pi_\mathrm{s}$ against an \emph{oracle} selector $\pi^\star$ that assigns priority $1$ to keyframes and $0$ otherwise, and a \emph{passive} baseline $\pi_{\mathrm{rnd}}$ that samples priorities from $\mathcal{U}[0,1]$. For memory retention analysis, we define active memory regret as $r_{\mathrm{active}} = \mathbb{E}\!\left[\frac{1}{T}\sum_{t=1}^{T}(\mathcal{L}^{\pi_\mathrm{s}}_t - \mathcal{L}^{\pi^\star}_t)\right]$ and passive memory regret as $r_{\mathrm{passive}} = \mathbb{E}\!\left[\frac{1}{T}\sum_{t=1}^{T}(\mathcal{L}^{\pi_{\mathrm{rnd}}}_t - \mathcal{L}^{\pi^\star}_t)\right]$. $r_{\mathrm{passive}}$ characterizes the penalty incurred by not actively selecting what to retain, while $r_{\mathrm{active}}$ measures how close the trained selector is to the oracle. The desired regime for long horizon information retention is $r_{\mathrm{passive}} \gg r_{\mathrm{active}} \approx 0$. We vary the masked-sampling support size $M$ across three synthetic domains and report the memory regret incurred using the learned objective in Figure~\ref{fig:keyframe-empirics}(b,c). With sufficiently large calibrated support, the active regret drops to zero, while passive regret gradually increases with $M$, providing empirical evidence for Theorem~1.

\textbf{Baselines and horizon-invariance experiment.} We compare KM (mnemonic-conditioned MLP/Transformer policies) against seven competitive sequence models, all trained on BC datasets with favorable parameter budgets. The set comprises
LSTM (BPTT/TBPTT)~\cite{hochreiter1997long}, an LSTM trained with backpropagation through
time, Decision-LSTM (DLSTM)~\citep{siebenborn2022crucial}, Decision Transformer (DT)~\citep{chen2021decision}, Decision-Mamba (DMamba) a DT-like state-space model~\citep{lv2024decision}, Transformer-XL (TrXL) with relative position encoding and a recurrent memory cache~\citep{dai2019transformer}, ELMUR, a long-horizon mixture-of-experts memory
transformer~\citep{cherepanov2025elmur}, and Random, a stochastic
baseline that we include as a noise floor. For each task domain, we evaluate across a range of horizons that includes both in-distribution and out-of-distribution settings, spanning orders of magnitude beyond training distribution. Results in Figure~\ref{fig:horizon_all5} show that \emph{KM maintains near-perfect performance across all horizons on every task, whereas all baseline policies collapse beyond the training range}. To probe the limits of generalization we also evaluated KM at two further
horizons $10^{5}$ and $10^{6}$ time steps. KM retained \emph{equivalent performance}
to its in-distribution behavior on all three synthetic tasks; see Appendix~\ref{appendix:experiments}, Table~\ref{tab:horizon-scaling}.

\subsection{Robot Manipulation Benchmarks and Real-world Experiment}

\setlength{\columnsep}{12pt}
\begin{wrapfigure}{r}{0.4\textwidth}
    \vspace{-24pt}
    \centering
    \includegraphics[width=\linewidth]{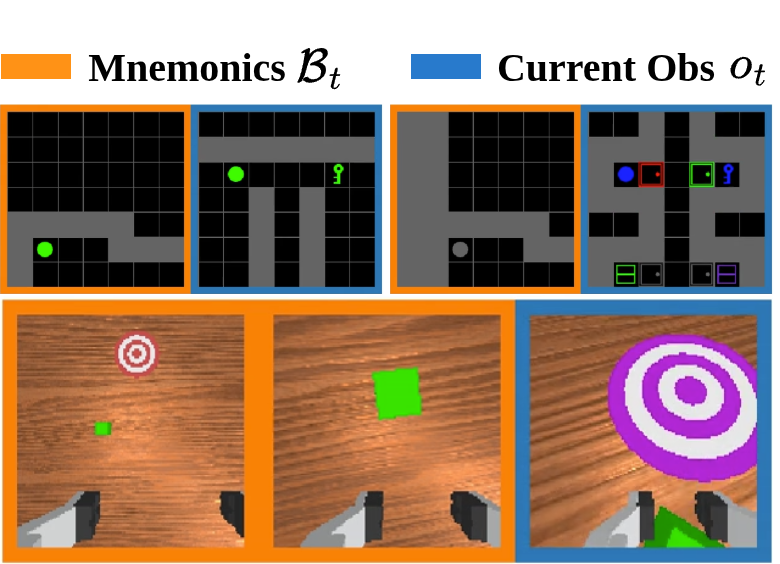}
    \caption{\textbf{Discovered mnemonics representing keyframes in grid and robot manipulation domains.}}
    \label{fig:mnemonics}
    \vspace{-4pt}
\end{wrapfigure}

\textbf{Memory benchmark.} We evaluate the performance of our method on memory-intensive robot manipulation using the \texttt{MIKASA-Robo} \citep{cherepanov2025memory} benchmark which has 32 tabletop manipulation tasks that evaluate object, spatial, sequential, and capacity memory capabilities of visuomotor policies. In each task the policy observes critical information at moments when the cues are visible, and must then act on that information after the cues are no longer in the observation. For example, in \texttt{RememberColor3}, the agent is briefly shown a colored cube as a cue; after the cue is removed from the observation, the robot must reach to and touch the matching cube among the three differently-colored cube options, requiring the policy to retain the cue across the time gap between observation and choice action. \texttt{MIKASA-Robo} thus evaluates the policy's ability to develop a relevant context from past observations to inform its action at the current state. To further evaluate the memory-retention capabilities of policies, that is, how long the policy can hold the necessary information for decision-making, we benchmark the policies on \texttt{RememberColor3-$\Delta T$} with cue-to-choice time delays $\Delta T$ larger than what was seen during training.

\textbf{Baselines.} For memory benchmarking, we compare our method on all 32 tasks of the \texttt{MIKASA-Robo} against a broad set of baselines including sequence models (Decision Transformer \cite{chen2021decision}), offline RL (Conservative Q-Learning \cite{kumar2020conservative}), Behavior Cloning, Diffusion Policies \cite{chi2023diffusionpolicy} and the recent state-of-the-art method \texttt{ELMUR} \cite{cherepanov2025elmur}. Here, \texttt{RATE} \cite{cherepanov2023recurrent} and \texttt{ELMUR} \cite{cherepanov2025elmur} serve as strong hybrid memory architecture baselines for comparison on tasks that demand memory capabilities from policies. For evaluation of memory retention capabilities, we compare our method with \texttt{ELMUR} and \texttt{PTP} \cite{torne2025learning}. \texttt{PTP} sub-samples frames from past observations for policy conditioning during training, in comparison to our learned approach of selecting relevant keyframes.

\textbf{Experimental setup.} We train Keyframe Mnemonics on the default dataset released by \texttt{MIKASA-Robo}. All results are reported as mean and standard error over $3$ different seeds with $100$ evaluations per seed. For the task designed to evaluate memory retention in manipulation tasks \texttt{RememberColor3-$\Delta T$}, we train all baselines with the same dataset and a $\Delta T=15$. During evaluation, we evaluate each baseline from $\Delta T=15$ to $\Delta T=150$ at intervals of $20$ timesteps, and then at $\Delta T=200,250$. We refer the reader to the Appendix~\ref{appendix:robot_implementation} for more details on the training configuration and the hyperparameters.

\textbf{Results on \texttt{MIKASA-Robo}.} Figure~\ref{fig:miaksa-result} summarizes the absolute success-rate improvement of Keyframe Mnemonics over the strongest published baseline, \texttt{ELMUR}, on the $23$ \texttt{MIKASA-Robo} tasks. Keyframe Mnemonics outperforms \texttt{ELMUR} on 17 of the 23 memory-intensive visual robot manipulation tasks and improves average absolute SR by 13.9\%; see full table of results in Appendix \ref{appendix:robot_implementation}. These results show that our self-supervised approach to keyframe identification yields policies that effectively leverage past observational cues from the environment to inform their actions.

\begin{figure}[t]
      \centering
      \begin{subfigure}{0.49\linewidth}
          \centering
          \includegraphics[width=\linewidth]{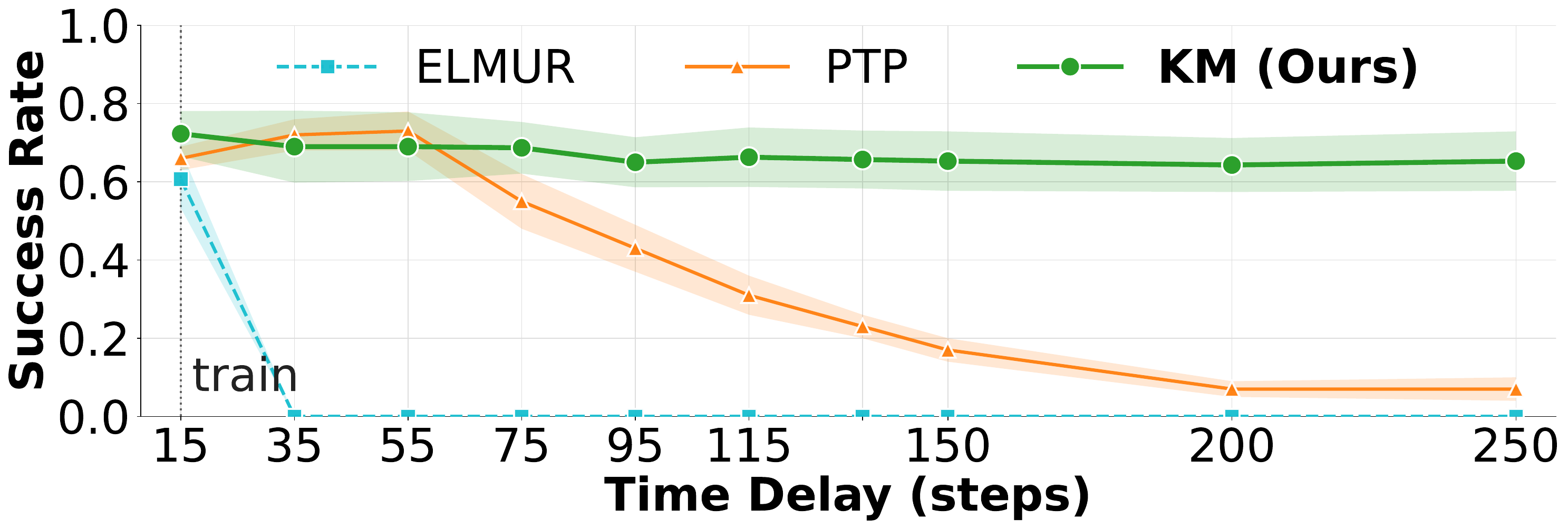}
          \caption{Memory retention on \texttt{RememberColor3-$\Delta T$} under cue-to-choice delays beyond the training horizon (marked at 15). \texttt{ELMUR} collapses immediately, \texttt{PTP} degrades gradually, while Keyframe Mnemonics sustains its training-time success rate up to 250 steps.}
          \label{fig:horizon}
      \end{subfigure}
      \hfill
      \begin{subfigure}{0.49\linewidth}
          \centering
          \includegraphics[width=\linewidth]{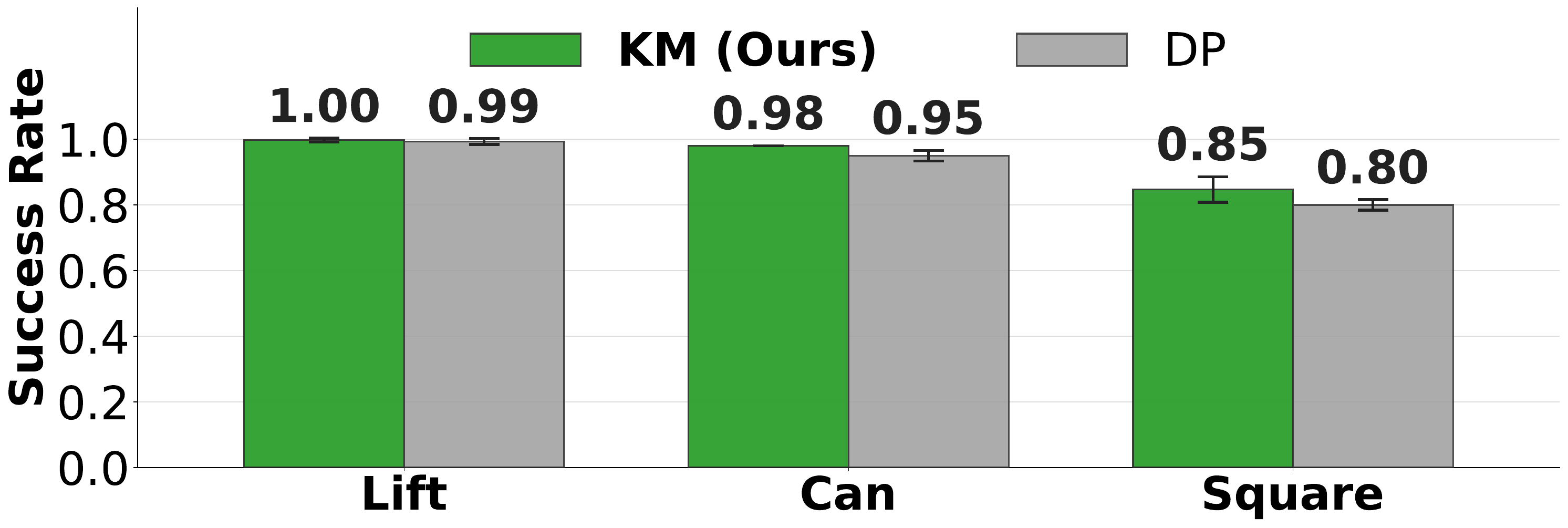}
          \caption{Keyframe Mnemonics matches the diffusion-policy (DP)~\cite{chi2023diffusionpolicy} baseline on three \texttt{Robomimic} tasks (\texttt{Lift}, \texttt{Can}, \texttt{Square}) where memory provides no benefit, confirming it does not regress in the no-memory regime.
}
          \label{fig:markovian}
      \end{subfigure}
      \caption{\textbf{Horizon-invariance in robot manipulation domain and results on Markovian tasks.}}
      \vspace{-6pt}
  \end{figure}

\textbf{Horizon-invariance experiment.} Figure~\ref{fig:horizon} evaluates how long each policy can hold the relevant cue once it is no longer visible. All three methods are trained at $\Delta T = 15$, but their out-of-distribution behavior diverges sharply: \texttt{ELMUR} collapses to near-zero success by $\Delta T = 35$, \texttt{PTP} retains roughly half its training performance through $\Delta T = 75$ before degrading, while Keyframe Mnemonics sustains its training-time success rate of $\sim\!70\%$ across the full evaluated range up to $\Delta T = 250$. This highlights the advantages of our learned approach to keyframe identification over sub-sampling of frames from the past, as performance of \texttt{PTP} deteriorates once the cue recedes from its observational context, while Keyframe Mnemonics keeps holding it in the policy's context.

\textbf{No regression on \texttt{Robomimic} tasks.} Figure~\ref{fig:markovian} shows that Keyframe Mnemonics matches the performance of diffusion-policy baseline trained without keyframes on three standard \texttt{Robomimic} manipulation tasks (\texttt{Lift}, \texttt{Can}, \texttt{Square}), confirming that the keyframe-conditioning does not deteriorate performance on Markovian benchmarks.

\begin{wrapfigure}{r}{0.28\textwidth}
    \vspace{-10pt}
    \centering
    \footnotesize
    \captionsetup{type=table}
    \caption{Training time.}
    \begin{tabular}{lc}
        \toprule
        \textbf{Method / Stage} & \textbf{Time} \\
        \midrule
        \textbf{KM (Ours)} & \textbf{10h 29m} \\
        \quad Proxy    & 1h 15m \\
        \quad Selector & 5h 46m \\
        \quad $\mathcal{D}^{\dagger}$ gen. & 1h 23m \\
        \quad BC policy & 2h 05m \\
        \midrule
        \texttt{PTP}~\cite{torne2025learning}   & \textbf{19h 34m} \\
        \texttt{ELMUR}~\cite{cherepanov2025elmur} & \textbf{1h 23m} \\
        \bottomrule
    \end{tabular}
    \label{tab:training_time}
    \vspace{-10pt}
\end{wrapfigure}
\begin{figure}[t]
    \centering
    \includegraphics[width=\linewidth]{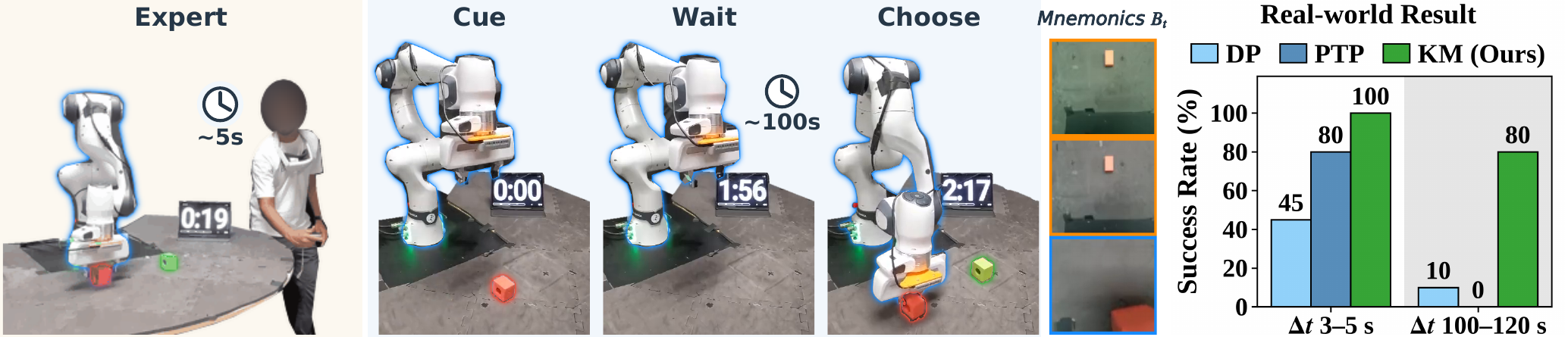}
    \caption{\textbf{Horizon-invariance in real-world \texttt{RememberColor2}-$\Delta T$ task.} From left to right: Expert user provides demonstrations with cue-to-choice delay of $\sim5$ seconds. During deployment, KM policy generalizes to a cue-to-choice delay of $\sim100$ seconds. Performance of \texttt{DP}~\cite{chi2023diffusionpolicy} and \texttt{PTP}~\cite{torne2025learning} deteriorates while KM retains average $80\%$ SR at $20\times$ longer test timesteps. See the webpage \url{https://keyframe-mnemonics.github.io} for policy rollout videos.}
    \label{fig:real-robot-result-and-keyframes}
    \vspace{0pt}
\end{figure}

\textbf{Real-world robot experiment}. We evaluate KM on the \texttt{RememberColor2}-$\Delta T$ task, where the robot observes a red or green cube for 3--5\,s. The cue is then removed, and after a delay, two cubes appear 30\,cm apart. The robot must reach for the previously observed color. We collect 30 VR-teleoperated demonstrations (15 per color) with ${\sim}5$\,s delays and evaluate at delays up to $20\times$ longer timesteps. Results shown in Figure~\ref{fig:real-robot-result-and-keyframes}.

\begin{figure}[t]
    \centering
    \includegraphics[width=\linewidth,trim=0 350 0 0,clip]{images/absolute_improvement_grouped.pdf}
    \caption{\textbf{Results across the 23 \texttt{MIKASA-Robo} tasks.} Keyframe Mnemonics improves on the strongest baseline (\texttt{ELMUR}) by an average of $+13.9$ absolute success rate, outperforming \texttt{ELMUR} on $17$ of the $23$ tasks. Results for tasks \texttt{RememberShape3/5/9}, \texttt{RememberColor3/5/9} and \texttt{RememberShapeColor} are aggregated, and the full table of results can be found in the Appendix~\ref{appendix:robot_implementation}.}
    \label{fig:miaksa-result}
    \vspace{-2pt}
\end{figure}

\textbf{Compute and training time.} Table~\ref{tab:training_time} reports training time for \texttt{RememberColor3}-$\Delta T$ on a single workstation with an NVIDIA RTX 3090 GPU and 12th-Gen Intel Core i9 CPU, with KM decomposed by training stages. KM trains faster than \texttt{PTP}, which conditions on subsampled frame buffers, while \texttt{ELMUR} is the fastest to train. We also trained \texttt{ELMUR} for $\sim$10 hours to match KM's training budget, but it remained at $0\%$ SR for $\Delta T > 15$ and failed to generalize beyond the training horizon even with further training.

\setlength{\columnsep}{10pt}
\section{Related Work}
\textbf{Explicit memory modeling.} Behavior cloning policies with dedicated memory have been developed using embedding caching \cite{cherepanov2025elmur}, learned memory compression models \cite{sorokin2022explain, zhou2025mtil, koo2025hamlet, fang2025sam2act, chung2026rethinking, tan2026memoact}, temporal state aggregation \cite{jang2025contextvla, torne2025learning}, and hybrid combinations of these approaches \cite{shi2025memoryvla, lei2026vpwem}. However, these methods generally construct compressed state representations that can blur information over long horizons. In contrast, our method preserves information-rich keyframes regardless of changes in horizon length and generalizes to timescales far beyond those seen during training.

\textbf{BC with keyframe annotations.} Prior work on keyframe-conditioned behavior cloning has relied on manual keyframe annotations \cite{yue2024learning, kou2024kisa, wu2025sage, chen2026non} or task-specific prompts for VLM-based selection \cite{mark2026bpp}. However, such approaches require additional supervision derived from oracle reasoning for each task. In contrast, our keyframe identification method is self-supervised, grounded in statistical learning principles, and requires no task-level oracle annotations. Therefore, we do not compare against methods that require additional data beyond expert demonstrations for policy training.

\textbf{Learning to manage policy context.} RL has been used in prior works to learn keyframe selection policies for optimizing context buffers \cite{demirbilek2025external, tasse2025beyond}. However, these methods have only been studied in simple grid and MuJoCo control environments, whereas we demonstrate that Keyframe Mnemonics generalizes to complex robot manipulation tasks in both simulation and the real world.

\textbf{Simplicity-bias in neural networks.} Simplicity-bias is a well-studied property of neural networks, reflecting their tendency to prefer simpler patterns over more complex ones \cite{arpit2017closer, valle2018deep, nakkiran2019sgd, shah2020pitfalls, mingard2025deep}. Recent work has leveraged simplicity bias to improve RL performance \cite{lee2024simba}, but this area remains largely underexplored, particularly in how such biases can be effectively utilized. We present one of the first methods that explicitly leverage this bias for behavior cloning in POMDP domains.

\section{Conclusion}
We presented a novel method for enabling horizon-invariant behavior cloning in partially observable memory intensive environments. Our approach uses a training strategy that samples random past observations to train a reward model, which is then optimized with reinforcement learning to learn a selector policy that discovers decision-relevant keyframes in a fully self-supervised manner. We provided theoretical analysis and intuition for our algorithm, along with empirical evidence supporting the proposed theory. Furthermore, we evaluated standard behavior cloning policy architectures conditioned on learned mnemonics, achieving state-of-the-art performance across diverse memory-intensive benchmarks, ranging from synthetic tasks to complex robot manipulation environments. We hope our work provides a step toward more reliable long-horizon behavior cloning by enabling policies to retain compact, decision-relevant context over extended horizons.

\textbf{Limitations of our approach.} The memory budget $k$ is a hyperparameter that is difficult to predetermine, which can lead to multiple information-equivalent buffer states and limit the number of correct keyframes retained in the buffer. In such cases, storing more keyframes in the inference-time priority queue can provide richer context for the policy. Each demonstration used in this work must contain at least one keyframe. This limits the current algorithm in multi-task or ``play demonstration'' settings, where keyframes may vary significantly across tasks or may not be present in every trajectory. Addressing \textit{horizon-invariance} for high-capacity memory remains an open problem, with the potential to enable more capable non-Markovian behavior cloning methods.

\begin{ack}
This material is based upon work supported by the Air Force Office of Scientific Research under award number FA9550-24-1-0239. Any opinions, findings, and conclusions or recommendations expressed in this material are those of the author(s) and do not necessarily reflect the views of the United States Air Force. We also want to thank our anonymous reviewers for providing great feedback and making the work stronger. Additionally, we thank Kunal Garg for providing computational resources for our simulation experiments, Prabhath Adireddi for helping with the robot experiments, Weiwei Gu for proofreading the theoretical results, and Yifan Zhou for the initial discussions.
\end{ack}

\setlength{\parskip}{5.5pt}
\bibliographystyle{unsrtnat}
\bibliography{references}

\newpage
\appendix
\input{appendix}

\newpage
\input{checklist.tex}

\end{document}

%% file: appendix.tex
\section*{Appendix}

\newcommand{\AppPart}[3]{%
  \par\addvspace{8pt}%
  \noindent\makebox[2em][l]{\textbf{#2}}%
  \textbf{\hyperref[#1]{#3}}\nobreak\leaders\hbox{$\cdot$\kern .35em}\hfill\nobreak\textbf{\pageref{#1}}\par
}
\newcommand{\AppSub}[3]{%
  \par\addvspace{1pt}%
  \noindent\hspace*{2em}\makebox[2.4em][l]{#2}%
  \hyperref[#1]{#3}\nobreak\leaders\hbox{$\cdot$\kern .35em}\hfill\nobreak\pageref{#1}\par
}

\AppPart{appendix:proof}{A}{Proofs}

\AppPart{appendix:algorithms}{B}{Algorithms}
\AppSub{app:buffer-queue}{B.1}{Buffer priority queue}
\AppSub{app:proxy-training}{B.2}{Proxy policy training}
\AppSub{app:selector-env}{B.3}{Selector RL environment}
\AppSub{app:e2e_training}{B.4}{End-to-end training and inference}

\AppPart{appendix:toy_implementation}{C}{Synthetic and Grid Domain Details}
\AppSub{app:synth-env}{C.1}{Synthetic environments}
\AppSub{app:grid-env}{C.2}{Grid environments}
\AppSub{app:training_dataset}{C.3}{Training dataset}
\AppSub{appendix:km_training_details}{C.4}{Training Keyframe Mnemonics policies}
\AppSub{app:syn-grid-arch}{C.5}{Baselines and benchmarking details}
\AppSub{app:param-budget}{C.6}{Parameter-budget comparison}
\AppSub{appendix:benchmark_details}{C.7}{Benchmark result analysis}
\AppSub{app:syn-grid-hardware}{C.8}{Hardware and wall-clock}

\AppPart{appendix:robot_implementation}{D}{Robot Manipulation Domain Details}
\AppSub{app:mikasa-env}{D.1}{\texttt{MIKASA-Robo} environments}
\AppSub{app:training:pipeline}{D.2}{Pipeline overview}
\AppSub{app:training:arch}{D.3}{Architectures and parameter counts}
\AppSub{app:training:optim}{D.4}{Optimizer and schedule}
\AppSub{app:training:proxy}{D.5}{Stage-1: Proxy policy $f_\theta$}
\AppSub{app:training:selector}{D.6}{Stage-2: Selector policy $\pi_\mathrm{s}$}
\AppSub{app:training:policy}{D.7}{Stage-3: BC policy $\pi_\mathrm{BC}$}
\AppSub{app:training:augment}{D.8}{Camera-image data augmentation}
\AppSub{app:mikasa-results}{D.9}{\texttt{MIKASA-Robo} all results}
\AppSub{app:robot-hardware}{D.10}{Hardware and wall-clock}

\AppPart{appendix:experiments}{E}{Experimental Details}
\AppSub{app:horizon-scaling}{E.1}{Horizon-invariance experiments on synthetic and grid domains}
\AppSub{app:training:hover}{E.2}{Horizon-invariance experiment on \texttt{RememberColor3-$\Delta T$}}
\AppSub{app:training:repro}{E.3}{Reproducibility}

\AppPart{app:ablations}{F}{Ablation Study}
\AppSub{app:abl-random}{F.1}{Learned selector against a random selector}
\AppSub{app:abl-gamma}{F.2}{Selector discount factor}
\AppSub{app:abl-surprise}{F.3}{Surprise coefficient}
\AppSub{app:abl-curobs}{F.4}{Current observation in proxy training}
\AppSub{app:abl-queue}{F.5}{Queue tie-break rule}
\AppSub{app:abl-buffer}{F.6}{Memory buffer size}

\AppPart{app:additional_analysis}{G}{Additional Analysis}
\AppSub{app:interpretability}{G.1}{Interpretability of discovered keyframes}
\AppSub{app:selector-priority}{G.2}{Selector priority behavior}
\AppSub{app:activation-map}{G.3}{BC Policy activation map}
\AppSub{app:memory-budget}{G.4}{Sensitivity to memory budget $k$}
\AppSub{app:reward-plots}{G.5}{PPO reward plots}
\AppSub{appendix:extended_limitations}{G.6}{Limitation discussion}

\AppPart{appendix:ext-rel-works}{H}{Extended Related Works}

\section{Proofs}
\label{appendix:proof}

\textbf{Lemma 1 (Combinatorial cardinality of the sample space).} \emph{The total cardinality $M$ of the space of distinct timestep-indexed training pairs $(\mathcal{B}_t, a_t)$ is given by:}
\[
M = \sum_{i=1}^N \left( T_i + \sum_{j=1}^k \binom{T_i + 1}{j + 1} \right).
\]

\emph{Proof.} Assume a stochastic training procedure that samples a trajectory $\tau_i$, a timestep $t \in [1, T_i]$, and a mask $\Psi_k$ selecting a subset of $j \le k$ observations from the history $h_t$ without replacement. The masked-sampling process generates an observation subset representing the buffer $\mathcal{B}$ by selecting $j$ observation frames, where $j \sim \min(\mathrm{Binomial}(t, 0.5), k)$. Because this spans $j \in \{0, \dots, \min(t, k)\}$, the sample space includes all subsets of size up to $k$, including the empty set for $j=0$.

For any timestep $t$, the number of unique combinations of size $j \ge 1$ is $\binom{t}{j}$ evaluating to $0$ if $t < j$. Summing these over all timesteps $t$ yields the total combinations for size $j$:
\[
\sum_{t=1}^{T_i} \binom{t}{j} = \binom{T_i + 1}{j + 1},
\]
which simplifies via the combinatorial Hockey-stick identity \cite{pascal1978traite}.

For the case where $j=0$, there is exactly $\binom{t}{0} = 1$ unique combination at each timestep. Summing this across all $T_i$ timesteps adds exactly $T_i$ combinations:
\[
\sum_{t=1}^{T_i} \binom{t}{0} = T_i.
\]
Summing the combinations for all possible mask sizes $j \in \{0, \dots, k\}$ gives the sample space cardinality for a single trajectory:
\[
M_i = T_i + \sum_{j=1}^k \binom{T_i + 1}{j + 1}.
\]
Finally, summing across all $N$ independent trajectories in $\mathcal{D}$ yields the total cardinality $M$ of the training sample space:
\[
M = \sum_{i=1}^N \left( T_i + \sum_{j=1}^k \binom{T_i + 1}{j + 1} \right).
\]
\hfill $\square$

\textbf{Assumption 3 (Realizability of conditional programs by interpolating hypotheses).}
For $Z \in \{\mathrm{key}, \mathrm{bg}\}$, let
$
f_Z^* = \arg\min_{f \in \mathcal{H}_Z:\, f(\mathcal{X}_{Z,\ell}) = A_\ell\ \forall \ell} K(f)
$
denote the minimum-complexity interpolating hypothesis using input source $Z$. We assume that the shortest conditional program mapping the training input sequence $\mathcal{X}_{Z,1:M}$ to the action sequence $A_{1:M}$ is realizable by the hypothesis class $\mathcal{H}$ up to an additive constant:
$
\left| K(f_Z^*) - K(A_{1:M} \mid \mathcal{X}_{Z,1:M}) \right| \le c_Z,
$
where $c_Z$ is an overhead constant independent of the sample size $M$. This assumption combines the standard interpretation of conditional Kolmogorov
complexity as the shortest program mapping inputs to outputs
\cite{grunwald2008algorithmic,li2008introduction} with the finite-sample
expressivity of overparameterized neural networks, which can interpolate or
memorize arbitrary training labels
\cite{zhang2016understanding, arpit2017closer, yun2019small}.

\textbf{Theorem 1 (Expected gap in hypothesis complexity).} \emph{Under the assumption that training pairs are sampled i.i.d.\ from a sufficiently large support of size $M$ (Lemma~1), the expected Kolmogorov complexity of the minimum-complexity background-based hypothesis $f^*_{\mathrm{bg}}$ is strictly greater than that of the keyframe-based hypothesis $f^*_{\mathrm{key}}$.}
\[
\mathbb{E}[K(f_{\mathrm{bg}}^*)] > \mathbb{E}[K(f_{\mathrm{key}}^*)].
\]

\emph{Proof.}
We first analyze complexity using established equivalence results from algorithmic information theory \cite{grunwald2008algorithmic}. Let $P(X, Y)$ denote a computable joint probability distribution generating data pairs $(X, Y)$ with marginal entropy $H(Y)$.

According to Theorem 10 in \citet{grunwald2008algorithmic}, the expected Kolmogorov complexity $K(\cdot)$ of a random variable equals its Shannon entropy up to an additive term bounded by the algorithmic complexity of the data-generating distribution itself:
\begin{equation}
    \mathbb{E}[K(Y)] = H(Y) + O(K(P)).
    \label{eq:expected_k}
\end{equation}

Similarly, Theorem 11 in \citet{grunwald2008algorithmic} establishes that expected algorithmic mutual information $I(Y:X)$ equals the probabilistic Shannon mutual information $I(Y;X)$ up to an additive term bounded by the same algorithmic complexity:
\begin{equation}
    \mathbb{E}[I(Y:X)] = I(Y;X) + O(K(P)).
    \label{eq:expected_mi}
\end{equation}

The algorithmic conditional complexity is defined via algorithmic mutual information (Equation 20 in \citet{grunwald2008algorithmic}) as $K(y \mid x^*) = K(y) - I(y:x)$. Taking the expectation over the data distribution and substituting Equations \ref{eq:expected_k} and \ref{eq:expected_mi}, we derive expected conditional Kolmogorov complexity in terms of conditional Shannon entropy \footnote{For notational simplicity, we suppress the distinction between conditioning on
$X$ and on its shortest program $X^*$, absorbing the resulting additive terms into
the $O(\cdot)$ overhead.}:
\begin{equation}
\begin{aligned}
\mathbb{E}[K(Y)] - \mathbb{E}[I(Y:X)]
&= H(Y) - I(Y;X) + O(K(P)), \\
\mathbb{E}[K(Y \mid X)]
&= H(Y \mid X) + O(K(P)).
\end{aligned}
\label{eq:conditional_equivalence}
\end{equation}

We now apply this to our setting, where $P = P(\mathcal{D})$ denotes the distribution induced by the POMDP and expert policy that generate dataset $\mathcal{D}$, and we take $K(\mathcal{D}) = K(P(\mathcal{D}))$ to represent the constant algorithmic overhead of this data-generating process. During training, $(\mathcal{X}, A)$ pairs are sampled from a finite support of cardinality $M$ induced by the masked-sampling procedure (Lemma 1) \footnote{Note that we use $\mathcal{X}$ to denote the input domain of interpolating hypotheses $f \in \mathcal{H}$, while $\mathcal{B}$ denotes a buffer subset used as input to the neural network $f_\theta$; in both cases, the input size is bounded by $k$.}. Extending Equation \ref{eq:conditional_equivalence} to the entire sequence of $M$ samples, the expected conditional Kolmogorov complexity of the sequence of actions given the sequence of sampled frames is bounded by the joint Shannon entropy plus the one-time algorithmic overhead of the data-generating distribution:
\begin{equation}
    \mathbb{E}[K(A_{1:M} \mid \mathcal{X}_{1:M})] = H(A_{1:M} \mid \mathcal{X}_{1:M}) + O(K(\mathcal{D})).
\end{equation}

Because the $M$ sampled training pairs are treated as i.i.d., the joint conditional Shannon entropy scales linearly, $H(A_{1:M} \mid \mathcal{X}_{1:M}) = M \cdot H(A \mid \mathcal{X})$. Substituting this yields:
\begin{equation}
    \mathbb{E}[K(A_{1:M} \mid \mathcal{X}_{1:M})] = M \cdot H(A \mid \mathcal{X}) + O(K(\mathcal{D})).
    \label{eq:sequence_complexity}
\end{equation}

The conditional Kolmogorov complexity $K(y \mid x)$ is strictly defined as the length of the shortest program on a universal conditional prefix machine that outputs $y$ when $x$ is provided as input \cite{grunwald2008algorithmic}. Therefore, for any interpolating hypothesis $f_Z$ satisfying $f_Z(\mathcal{X}_{Z,\ell}) = A_\ell$ for all $\ell$, where $Z \in \{\mathrm{key}, \mathrm{bg}\}$, there exists a fixed evaluator program which, given a description of $f_Z$ and the input sequence $\mathcal{X}_{Z,1:M}$, perfectly reconstructs the action sequence $A_{1:M}$. Hence, the conditional sequence complexity is upper-bounded by the hypothesis complexity:
\[
K(A_{1:M} \mid \mathcal{X}_{Z,1:M}) \le K(f_Z^*) + O(1).
\]
Assumption~3 states that, for each input source
$Z\in\{\mathrm{key},\mathrm{bg}\}$, the minimum-complexity interpolating
hypothesis $f_Z^*$ realizes the corresponding shortest conditional program up to
an additive constant:
\[
\left|
K(f_Z^*) -
K(A_{1:M}\mid \mathcal X_{Z,1:M})
\right|
\le c_Z.
\]
Equivalently, unpacking the absolute value gives the two-sided bound
\[
K(A_{1:M}\mid \mathcal X_{Z,1:M}) - c_Z
\le
K(f_Z^*)
\le
K(A_{1:M}\mid \mathcal X_{Z,1:M}) + c_Z.
\]
To analyze the complexity gap
$K(f_{\mathrm{bg}}^*)-K(f_{\mathrm{key}}^*)$, we derive a conservative lower
bound by using a lower bound for the background hypothesis and an upper bound for
the keyframe hypothesis. This yields
\begin{align*}
K(f_{\mathrm{bg}}^*)
&\ge
K(A_{1:M} \mid \mathcal{X}_{\mathrm{bg},1:M}) - c_{\mathrm{bg}}, \\
K(f_{\mathrm{key}}^*)
&\le
K(A_{1:M} \mid \mathcal{X}_{\mathrm{key},1:M}) + c_{\mathrm{key}}.
\end{align*}

Subtracting the keyframe upper bound from the background lower bound, and taking the expectation over the data-generating distribution, yields the expected hypothesis complexity gap:
\[
\mathbb{E}[K(f_{\mathrm{bg}}^*)] - \mathbb{E}[K(f_{\mathrm{key}}^*)] \ge \mathbb{E}[K(A_{1:M} \mid \mathcal{X}_{\mathrm{bg},1:M})] - \mathbb{E}[K(A_{1:M} \mid \mathcal{X}_{\mathrm{key},1:M})] - (c_{\mathrm{key}} + c_{\mathrm{bg}}).
\]
Substituting the sequence complexities from Equation \ref{eq:conditional_equivalence} and \ref{eq:sequence_complexity}, we obtain:
\[
\mathbb{E}[K(f_{\mathrm{bg}}^*)] - \mathbb{E}[K(f_{\mathrm{key}}^*)] \ge M \big( I(A ; \mathcal{X}_{\mathrm{key}}) - I(A ; \mathcal{X}_{\mathrm{bg}}) \big) - O(K(\mathcal{D})) - (c_{\mathrm{key}} + c_{\mathrm{bg}}).
\]
Under Assumption 2, the mutual information gap $I(A ; \mathcal{X}_{\mathrm{key}}) - I(A ; \mathcal{X}_{\mathrm{bg}})$ is strictly positive. Because the sample support $M$ is combinatorially massive (Lemma 1), this linear scaling term strictly dominates the combined constant overheads $O(K(\mathcal{D})) + (c_{\mathrm{key}} + c_{\mathrm{bg}})$. Thus, for sufficiently large $M$, we obtain:
\[
\mathbb{E}[K(f_{\mathrm{bg}}^*)] > \mathbb{E}[K(f_{\mathrm{key}}^*)].
\]
\hfill $\square$

\textbf{Assumption 4 (Simplicity-biased SGD approximation).}
Conditioned on reaching an interpolating solution in $\mathcal H^*$, SGD is
approximated as selecting hypotheses with probability biased toward lower
Kolmogorov complexity:
\[
P_{\mathrm{SGD}}(f \mid \mathcal D) \propto 2^{-K(f)},
\qquad f \in \mathcal H^* .
\]
This approximation follows the simplicity-bias view of neural networks, where
low-complexity functions receive larger prior mass
\cite{valle2018deep, mingard2025deep, mingard2021sgd}.

\textbf{Proposition 1 (Conditional simplicity-bias preference).}
\emph{Under the simplicity-bias assumption and the complexity gap established in
Theorem 1, Stochastic Gradient Descent (SGD) is biased toward the invariant hypothesis class $\mathcal H_{\mathrm{in}}$. Given a masking-induced sample distribution with support size $M$ (Lemma 1), the probability of selecting an invariant hypothesis is approximated by}
\[
P_{\mathrm{SGD}}(\mathcal H_{\mathrm{in}} \mid \mathcal D)
\approx
\frac{1}{
1 + 2^{-\left[
M \big(
I(A ; \mathcal X_{\mathrm{key}})
-
I(A ; \mathcal X_{\mathrm{bg}})
\big)
-
\mathcal O(1)
\right]}
}.
\]
\emph{Proof.}
Under the Empirical Risk Minimization (ERM) framework \cite{vapnik2013nature}, an overparameterized neural
network can realize hypotheses in the interpolating set $\mathcal H^*$ that achieve zero empirical risk on $\mathcal D$. To analyze which interpolating hypothesis is selected by SGD, we use the simplicity-bias approximation in Assumption 4, motivated by evidence that neural network function priors and SGD solutions are biased toward low-complexity functions \cite{valle2018deep, mingard2025deep, mingard2021sgd}. The probability of selecting an invariant hypothesis is obtained by marginalizing over the invariant class:
\[
P_{\mathrm{SGD}}(\mathcal H_{\mathrm{in}} \mid \mathcal D)
\approx
\frac{
\sum_{f \in \mathcal H_{\mathrm{in}}} 2^{-K(f)}
}{
\sum_{f \in \mathcal H^*} 2^{-K(f)}
}.
\]
Since $\mathcal H^*$ is partitioned into the invariant class and the
non-invariant class,
\[
\mathcal H^*
=
\mathcal H_{\mathrm{in}}
\cup
\mathcal H_{\neg\mathrm{in}},
\]
we can rewrite this as
\[
P_{\mathrm{SGD}}(\mathcal H_{\mathrm{in}} \mid \mathcal D)
\approx
\frac{
\sum_{f \in \mathcal H_{\mathrm{in}}} 2^{-K(f)}
}{
\sum_{f \in \mathcal H_{\mathrm{in}}} 2^{-K(f)}
+
\sum_{f \in \mathcal H_{\neg\mathrm{in}}} 2^{-K(f)}
}.
\]

Because the probability mass decays exponentially with complexity, we use the
standard minimum-description approximation that each class-level mass is
dominated, up to constant factors independent of $M$, by its minimum-complexity
interpolating hypothesis. Let
\[
f_{\mathrm{key}}^*
=
\arg\min_{f \in \mathcal H_{\mathrm{in}}} K(f),
\qquad
f_{\mathrm{bg}}^*
=
\arg\min_{f \in \mathcal H_{\neg\mathrm{in}}} K(f).
\]
Then
\[
\sum_{f \in \mathcal H_{\mathrm{in}}} 2^{-K(f)}
\approx
2^{-K(f_{\mathrm{key}}^*)},
\qquad
\sum_{f \in \mathcal H_{\neg\mathrm{in}}} 2^{-K(f)}
\approx
2^{-K(f_{\mathrm{bg}}^*)}.
\]
Substituting this approximation gives
\[
P_{\mathrm{SGD}}(\mathcal H_{\mathrm{in}} \mid \mathcal D)
\approx
\frac{
2^{-K(f_{\mathrm{key}}^*)}
}{
2^{-K(f_{\mathrm{key}}^*)}
+
2^{-K(f_{\mathrm{bg}}^*)}
}.
\]
Dividing numerator and denominator by $2^{-K(f_{\mathrm{key}}^*)}$ gives
\[
P_{\mathrm{SGD}}(\mathcal H_{\mathrm{in}} \mid \mathcal D)
\approx
\frac{1}{
1 + 2^{-\left(K(f_{\mathrm{bg}}^*) - K(f_{\mathrm{key}}^*)\right)}
}.
\]

By Theorem 1, the expected complexity gap between the non-invariant/background-based hypothesis and the invariant/keyframe-based hypothesis satisfies
\[
\mathbb E[K(f_{\mathrm{bg}}^*)]
-
\mathbb E[K(f_{\mathrm{key}}^*)]
\ge
M
\big(
I(A ; \mathcal X_{\mathrm{key}})
-
I(A ; \mathcal X_{\mathrm{bg}})
\big)
-
\mathcal O(1).
\]
Here, the constant overhead $O(K(\mathcal{D})) + c_{\mathrm{key}} + c_{\mathrm{bg}}$
is absorbed into the $\mathcal O(1)$ term. Therefore, for a typical dataset sampled from the masking-induced distribution, we use the
corresponding expected gap approximation
\[
K(f_{\mathrm{bg}}^*) - K(f_{\mathrm{key}}^*)
\approx
M
\big(
I(A ; \mathcal X_{\mathrm{key}})
-
I(A ; \mathcal X_{\mathrm{bg}})
\big)
-
\mathcal O(1).
\]
Substituting this into the class-probability expression yields
\[
P_{\mathrm{SGD}}(\mathcal H_{\mathrm{in}} \mid \mathcal D)
\approx
\frac{1}{
1 + 2^{-\left[
M \big(
I(A ; \mathcal X_{\mathrm{key}})
-
I(A ; \mathcal X_{\mathrm{bg}})
\big)
-
\mathcal O(1)
\right]}
}.
\]
Thus, when the keyframe information advantage is positive and the assumptions hold, the simplicity-bias approximation predicts a rapid increase in the probability of selecting an invariant hypothesis. This follows because the exponent scales linearly with the combinatorially large support size $M$. \footnote{If the assumptions fail, for example, when a lower-complexity non-invariant shortcut exists, this preference need not hold.}
\hfill $\square$

\textbf{Corollary 1 (Keyframe recovery from the frozen loss objective).} \emph{
Let us define a subset-selection objective function
\[
J_{\theta^*}(\mathcal B)
=
\mathbb E\!\left[
\mathcal L(f_{\theta^*}(\mathcal B), A)
\right]
\]
over a frozen predictor $f_{\theta^*}$ obtained under Proposition~1 after training with negative log-likelihood loss $\mathcal L$, where $f_{\theta^*}$ represents a near-optimal invariant hypothesis in $\mathcal H_{\mathrm{in}}$. Consider candidate buffer subsets $\mathcal C_k=\{\mathcal B_t\subseteq h_t:|\mathcal B_t|\le k\}$, and let $\mathcal B_{\mathrm{opt}} \in \arg\min_{\mathcal B\in\mathcal C_k} J_{\theta^*}(\mathcal B)$. If keyframe subsets $\mathcal B \subset \mathcal X_{\mathrm{key}}$ have a positive action-information margin over all non-keyframe candidate subsets, and this margin dominates the approximation error of the frozen predictor $f_{\theta^*}$, then the loss objective $J_{\theta^*}(\mathcal B)$ recovers the keyframes up to information-equivalent subsets. In particular, if $\mathcal X_{\mathrm{key}}$ is unique under the memory budget $k$, then $\mathcal B_{\mathrm{opt}}=\mathcal X_{\mathrm{key}}$.
}

\emph{Proof.}
For negative log-likelihood loss, the Bayes-optimal expected loss conditioned on
an input buffer subset $\mathcal B$ is the conditional entropy, since the logarithmic score is a strictly proper scoring rule and the expected negative log-likelihood decomposes into conditional entropy plus a nonnegative KL divergence
\cite{cover1999elements,gneiting2007strictly}:
\[
\inf_f
\mathbb E[
\mathcal L(f_{\theta^*}(\mathcal B),A)
]
=
H(A\mid \mathcal B).
\]
Using the Shannon identity
\[
H(A\mid \mathcal B)
=
H(A)-I(A;\mathcal B)
\]
minimizing the Bayes-optimal negative log-likelihood is
equivalent to maximizing the mutual information between the selected subset and
the action \cite{cover1999elements}. Define the action-information margin of the keyframe set over competing candidate subsets as
\[
\mu_k
=
\inf_{\mathcal B\in\mathcal C_k:\,
\mathcal B\not\sim \mathcal X_{\mathrm{key}}}
\left[
I(A;\mathcal X_{\mathrm{key}})
-
I(A;\mathcal B)
\right],
\]
where $\mathcal B\sim\mathcal X_{\mathrm{key}}$ denotes an information-equivalent
subset. The condition $\mu_k>0$ extends Assumption 2 from the background subset
to the full candidate search space $\mathcal C_k$, requiring that keyframes have
strictly larger action information than any non-equivalent competing subset.

Since Proposition 1 biases training toward the invariant class, suppose the
frozen model $f_{\theta^*}$ is a near-optimal keyframe-conditioned predictor.
Define its approximation error on the keyframe input as
\[
\epsilon_{\mathrm{app}}
=
J_{\theta^*}(\mathcal X_{\mathrm{key}})
-
H(A\mid \mathcal X_{\mathrm{key}}).
\]
Since negative log-likelihood risk is lower-bounded by conditional entropy, $\epsilon_{\mathrm{app}}\ge 0$. Thus,
\[
J_{\theta^*}(\mathcal X_{\mathrm{key}})
=
H(A\mid \mathcal X_{\mathrm{key}})
+
\epsilon_{\mathrm{app}}.
\]

For any competing subset
$\mathcal B\in\mathcal C_k$ with
$\mathcal B\not\sim\mathcal X_{\mathrm{key}}$, every predictor using only
$\mathcal B$ has negative log-likelihood risk at least its Bayes risk by
nonnegativity of the KL divergence \cite{cover1999elements,gneiting2007strictly}:
\[
J_{\theta^*}(\mathcal B)
\ge
H(A\mid \mathcal B).
\]
By the definition of $\mu_k$,
\[
I(A;\mathcal X_{\mathrm{key}})-I(A;\mathcal B)
\ge
\mu_k,
\]
which is equivalent to
\[
H(A\mid \mathcal B)
\ge
H(A\mid \mathcal X_{\mathrm{key}})
+
\mu_k.
\]
Therefore,
\[
J_{\theta^*}(\mathcal B)
\ge
H(A\mid \mathcal X_{\mathrm{key}})
+
\mu_k.
\]

If the information margin dominates the approximation error,
\[
\mu_k>\epsilon_{\mathrm{app}},
\]
then
\[
J_{\theta^*}(\mathcal X_{\mathrm{key}})
=
H(A\mid \mathcal X_{\mathrm{key}})
+
\epsilon_{\mathrm{app}}
<
H(A\mid \mathcal X_{\mathrm{key}})
+
\mu_k
\le
J_{\theta^*}(\mathcal B)
\]
for every non-equivalent competing subset $\mathcal B$. Hence, the minimizer of
the frozen loss objective belongs to the keyframe equivalence class. If
$\mathcal X_{\mathrm{key}}$ is unique under the memory budget $k$, then
\[
\mathcal B_{\mathrm{opt}}=\mathcal X_{\mathrm{key}}.
\]
This keyframe recovery guarantee naturally extends to continuous action spaces optimized via Mean Squared Error (MSE). Under the assumption that the conditional distribution of the actions is Gaussian with a fixed variance, minimizing the MSE is equivalent to minimizing the negative log-likelihood up to affine constants \cite{bishop2006pattern, goodfellow2016deep}. Consequently, the same argument applies with conditional differential entropy
and mutual information replacing their discrete counterparts, so the frozen MSE
objective recovers $\mathcal X_{\mathrm{key}}$ under the same positive-margin and
approximation-error conditions.
\hfill $\square$

\textbf{Theorem 2 (Infinite-horizon keyframe retention).}
\emph{Assume a POMDP in which the expert behavior strictly satisfies Assumptions~1 and 2. Let $\pi_{\mathrm{s}}^*$ be an optimal selector minimizing the frozen objective $J_{\theta^*}(\mathcal{B}_t)$ (Corollary~1). Suppose a unique keyframe set $\mathcal{X}_{\mathrm{key}}$ exists that has the highest mutual information with the action and fits within the memory budget $k$. Then, as $T \to \infty$, $\pi_{\mathrm{s}}^*$ assigns higher priority to keyframes than to background frames, ensuring that the priority queue retains the keyframe set for the entire horizon $T$ without eviction, i.e., $\mathcal{B}_T = \mathcal{X}_{\mathrm{key}}$.}

\emph{Proof.}
We analyze the behavior of an optimal selector policy $\pi_{\mathrm{s}}^*$ over
the temporal horizon $T$. The reward function $R(\mathcal{B}_t)$ is designed to
increase as the frozen loss objective $J_{\theta^*}(\mathcal{B}_t)$ decreases.
Thus, maximizing expected return requires $\pi_{\mathrm{s}}^*$ to assign priority
scores that configure the buffer $\mathcal{B}_t$ into the subset minimizing
$J_{\theta^*}(\mathcal{B})$.

By Assumption~2 and Corollary~1, the keyframe set
$\mathcal{X}_{\mathrm{key}}$ has a strictly positive action-information margin
$\mu_k > 0$ over all non-keyframe subsets. Consider a timestep $t$ at which the
keyframe set has already been inserted into the buffer. Let a new background
observation $o_t \in \mathcal{X}_{\mathrm{bg}}$ be evaluated. If $\pi_{\mathrm{s}}^*$
assigns a priority $p_t$ high enough to evict a keyframe, then the resulting buffer
$\mathcal{B}_t$ has strictly lower mutual information with the action, decreasing
by at least $\mu_k$.

By Corollary~1, this induces a strictly higher expected loss:
\[
J_{\theta^*}(\mathcal{B}_t)
\ge
H(A \mid \mathcal{X}_{\mathrm{key}}) + \mu_k
>
J_{\theta^*}(\mathcal{X}_{\mathrm{key}}),
\]
which penalizes the reward. Therefore, any such eviction is suboptimal,
and $\pi_{\mathrm{s}}^*$ must assign lower priority to background observations
than to keyframes. Once the keyframe set $\mathcal{X}_{\mathrm{key}}$ is fully inserted into the
buffer, this argument applies recursively. Hence, under the optimal selector policy,
background observations never receive sufficient priority to evict keyframes.
Thus, the priority queue preserves $\mathcal{X}_{\mathrm{key}}$ indefinitely,
yielding $\mathcal{B}_T = \mathcal{X}_{\mathrm{key}}$.
\hfill $\square$

\section{Algorithms}
\label{appendix:algorithms}

\begin{algorithm}[h!]
\caption{FIFO priority queue --- training-time eviction rule.}
\label{alg:fifo_queue}
\begin{algorithmic}[1]
\State \textbf{Parameters:} training memory budget $k$, rejection threshold $\eta_{\mathrm{r}}$
\State \textbf{State:} buffer $\mathcal{B}_t \gets \emptyset$, insertion counter $c \gets 0$
\Procedure{Push}{observation $o$, priority $p$}
    \If{$p < \eta_{\mathrm{r}}$}
        \State \Return \Comment{reject priority below threshold}
    \EndIf
    \State $c \gets c + 1$ \Comment{monotone insertion timestamp}
    \If{$|\mathcal{B}_t| < k$}
        \State $\mathcal{B}_{t+1} \gets \mathcal{B}_t \cup \{(o, p, c)\}$ \Comment{free slot, insert directly}
    \Else
        \State $(o^{\star}, p^{\star}, c^{\star}) \gets \arg\min_{(o', p', c') \in \mathcal{B}_t} (p', \, c')$
            \Comment{lowest priority; \emph{oldest} on a tie}
        \If{$p \ge p^{\star}$}
            \State $\mathcal{B}_{t+1} \gets \big(\mathcal{B}_t \setminus \{(o^{\star}, p^{\star}, c^{\star})\}\big)
                                    \cup \{(o, p, c)\}$
        \EndIf
    \EndIf
\EndProcedure
\end{algorithmic}
\end{algorithm}

\begin{algorithm}[h!]
\caption{LIFO priority queue --- inference-time eviction rule.}
\label{alg:lifo_queue}
\begin{algorithmic}[1]
\State \textbf{Parameters:} inference memory budget $k$, rejection threshold $\eta_{\mathrm{r}}$,
       no-repeat threshold $\eta_{\mathrm{n}}$
\State \textbf{State:} buffer $\mathcal{B}_t \gets \emptyset$, insertion counter $c \gets 0$,
       last priority $p_{\mathrm{last}} \gets 0$
\Procedure{Push}{observation $o$, priority $p$}
    \If{$p < \eta_{\mathrm{r}}$ \textbf{or} $|p - p_{\mathrm{last}}| < \eta_{\mathrm{n}}$}
        \State \Return \Comment{reject low or quasi-repeated priority}
    \EndIf
    \State $c \gets c + 1$
    \If{$|\mathcal{B}_t| < k$}
        \State $\mathcal{B}_{t+1} \gets \mathcal{B}_t \cup \{(o, p, c)\}$
    \Else
        \State $(o^{\star}, p^{\star}, c^{\star}) \gets \arg\min_{(o', p', c') \in \mathcal{B}_t} (p', \, -c')$
            \Comment{lowest priority; \emph{newest} on a tie}
        \If{$p \ge p^{\star}$}
            \State $\mathcal{B}_{t+1} \gets \big(\mathcal{B}_t \setminus \{(o^{\star}, p^{\star}, c^{\star})\}\big)
                                    \cup \{(o, p, c)\}$
        \EndIf
    \EndIf
    \State $p_{\mathrm{last}} \gets p$ \Comment{update even if no eviction occurred}
\EndProcedure
\end{algorithmic}
\end{algorithm}
\subsection{Buffer priority queue}
\label{app:buffer-queue}
The memory buffer $\mathcal{B}_t$ introduced in
Section~\ref{sec:selector_optimization} is realised by a bounded
priority queue of capacity $k$. The buffer design supports two key behaviors: (i) \textit{overflow forgetting}, where consistently assigning high priorities causes queue overflow and removes keyframes from the buffer, and (ii) \textit{background rejection}, where assigning priorities below a rejection threshold actively ignores background observations. Both training and inference queue use the
same insertion semantics - the selector emits a priority $p \in [0, 1]$
per observation, and the queue accepts or rejects based on $p$; the
two phases differ \emph{only} in how ties on the lowest priority are
broken at eviction time. Both preserve the temporal ordering of frames in the buffer. Algorithms~\ref{alg:fifo_queue}
and~\ref{alg:lifo_queue} state the two rules. We use $\eta_{\mathrm{r}}
= 0.5$ as the rejection threshold throughout, and
$\eta_{\mathrm{n}} = 0.05$ as the no-repeat threshold for the
inference queue.

\textbf{Why FIFO at training time.} The training rule
(Algorithm~\ref{alg:fifo_queue}) breaks ties in favour of evicting the
\emph{oldest} entry. Under this rule, a frame that has been assigned
a uniformly high priority will still age out as new high-priority
frames arrive, so the selector cannot rely on saturating priorities
to retain information. This sharpens the gradient signal driving
$\pi_{\mathrm{s}}$ towards genuinely discriminative scores rather
than constant-1 outputs.

\textbf{Why LIFO at inference time.} The inference rule (Algorithm~\ref{alg:lifo_queue}) breaks ties by evicting the \emph{newest} entry. Thus, initial frames assigned high priority by the selector are preserved, enabling robust keyframe retention and maintaining the keyframe set over arbitrarily long horizons. The no-repeat threshold $\eta_{\mathrm{n}}$ further prevents near-tied priority observations from entering the queue, ensuring that it retains diversity. This especially helps with retaining keyframes when there exists multiple information-equivalent keyframe subsets in the demonstration dataset.

Note that under a near-optimal selector and a sufficient memory budget $k$ (Theorem 2), such that the buffer can store all true keyframes, both queues exhibit identical behavior. We observed this empirically in synthetic domains, where the discovered mnemonics remained consistent regardless of the queue strategy used during inference.

\subsection{Proxy policy training}
\label{app:proxy-training}

Section~\ref{sec:keyframe_discovery} introduces the proxy policy $f_{\theta}:\mathcal{B}\to\mathcal{A}$
trained by masked-sampling of past observations (Algorithm~\ref{alg:masked-sampling}).
In practice, we implement two equivalent realizations of this training procedure,
selected based on the scale of the masked-sampling support $M$ (Lemma~1).

\paragraph{Online masked-sampling (robot manipulation).}
For \texttt{MIKASA-Robo} tasks we draw a fresh
mask $\Psi_{k}$ at every gradient step, as stated in
Algorithm~\ref{alg:masked-sampling}: a trajectory $\tau\sim\mathcal{D}$ and timestep
$t\in\{1,\dots,T_{\tau}\}$ are sampled, the masked-sampling distribution
$n\sim\min(\mathrm{Binomial}(t,0.5),k)$ selects a memory buffer
$\mathcal{B}_{t}=\{o_{i}\in h_{t}\mid\Psi_{k}(i)=1\}$ of cardinality
$\lvert\mathcal{B}_{t}\rvert\le k$, and the proxy regresses the expert action
$a_{t}$ under negative log-likelihood loss $\mathcal{L}$.

\paragraph{Offline masked-sampling rollouts (synthetic and grid domains).}
For the synthetic and grid domains (Add, Scattered-Copy, T-Maze, MiniGrid-Memory, LTMB-Hallway), the support size $M$ is small enough to precompute masked samples. We therefore run multiple offline passes over $\mathcal{D}$ and cache the random buffer-action pairs on disk, rather than resampling masks at every gradient step. In each pass, for each trajectory $\tau_i \in \mathcal{D}$, we iterate over $t=1,\dots,T_i$, and execute the masked-sampling procedure (Algorithm~\ref{alg:masked-sampling}) for each timestep. The resulting pairs $(\mathcal{B}_t,a_t)$ are written to the cache, and the proxy $f_\theta$ is trained for multiple epochs on the cached pairs. The number of passes trades off cache size against Monte Carlo coverage of the masked-sampling support and is chosen empirically so that the total number of cached pairs exceeds the sufficient support size $M$ required by Theorem~1. Although online masked-sampling provides the maximum possible cardinality support $M$, this offline counterpart helped us with more controlled experiments on $M$ to empirically verify Theorem~1 as shown in Figure~\ref{fig:keyframe-empirics}. See Appendix~\ref{app:training_dataset} for more details.

\paragraph{Lemma 2 (Equivalence to uniform selector priority queue).}
\emph{Let $\mathcal{P}_{\mathrm{mask}}$ denote the distribution over buffer states
$\mathcal{B}$ induced by masked-sampling (Algorithm~\ref{alg:masked-sampling}), and let
$\mathcal{P}_{\mathrm{queue}}$ denote the distribution induced by using a random selector $\pi_\mathrm{rand}$
that assigns uniform priorities $p\in[0,1]$ to observations retained in a priority queue. Both procedures induce the same distribution over buffer states.}

\emph{Proof.} Under $\mathcal{P}_{\mathrm{mask}}$, the buffer is drawn by picking a size
$n\sim\min(\mathrm{Binomial}(t,0.5),k)$ uniformly at random and then
selecting $n$ indices uniformly without replacement from $\{1,\dots,t\}$ (Lemma~1).
Under $\mathcal{P}_{\mathrm{queue}}$, each past observation $o_{i}\in h_{t}$
receives an i.i.d.\ priority $p_{i}\sim \mathrm{Unif}(0,1)$, and the queue
retains the top-$k$ priorities exceeding the rejection threshold
$\eta_{r}$. Because the priorities are exchangeable and independent of
$i$, every $k$-subset of $h_{t}$ is selected with equal probability
conditional on its cardinality, and the cardinality distribution is
$\min(\mathrm{Binomial}(t,\Pr[p>\eta_{r}]),k)$. With $\eta_{r}=0.5$ we have $\Pr[p>\eta_{r}]=0.5$, recovering the masked-sampling cardinality distribution of
Algorithm~\ref{alg:masked-sampling}. Hence
$\mathcal{P}_{\mathrm{queue}}=\mathcal{P}_{\mathrm{mask}}$ at the level of
marginal over $\mathcal{B}_{t}$. \hfill $\square$

The above result establishes an equivalence between the masked-sampling
procedure (Algorithm~\ref{alg:masked-sampling}) and selector-driven buffer
construction under a random selector policy. Intuitively, this shows that the
selector optimization (Section~\ref{sec:selector_optimization}) operates over
the same combinatorial space of buffer states explored during masked-sampling
training. Consequently, the learned objective $J_{\theta^*}$ remains aligned
with the optimization problem solved by the selector policy. Because
$J_{\theta^*}$ wraps a function that learns a keyframe-conditioned hypothesis favored by
Theorem~1, optimizing the selector is naturally biased toward recovering
keyframe subsets.

\subsection{Selector RL environment}
\label{app:selector-env}

We provide the algorithmic specification of the deterministic MDP
$\langle\mathcal{O},\mathcal{P},\mathcal{K},R,\gamma\rangle$ described in
Section~\ref{sec:selector_optimization}, on which the selector policy
$\pi_{\mathrm s}$ is trained with PPO. Algorithm~\ref{alg:selector-env} states the
per-step transition: the policy observes $o_{t}\in\mathcal{O}$, emits a
priority $p_{t}\in\mathcal{P}=[0,1]$, and the environment updates the
priority queue $\mathcal{B}_{t}$ via the transition rule
$\mathcal{K}$ (Algorithm~\ref{alg:fifo_queue}). The post-update buffer is then
scored by the frozen proxy $f_{\theta^{\ast}}$ to compute the surprise-weighted
log-reward $R(\mathcal{B}_{t})$ from Equation~\ref{eq:reward}. The episode
terminates after $T_{\tau}$ steps and the next trajectory is loaded from
$\mathcal{D}$.

\begin{algorithm}
\caption{Selector RL environment step}
\label{alg:selector-env}
\textbf{Require:} dataset $\mathcal{D}$, frozen proxy $f_{\theta^{\ast}}$,
memory budget $k$, normalisation constant $\lambda$, surprise coefficients $\{\beta_{\tau,t}\}$, queue eviction
rule \textsc{Push} (Algorithm~\ref{alg:fifo_queue})\\
\textbf{State:} current trajectory $\tau\in\mathcal{D}$, step counter $t$,
buffer $\mathcal{B}_{t-1}$
\begin{algorithmic}[1]
    \Procedure{Step}{$o_{t}, p_{t}$} \Comment{$p_{t}\sim\pi_{\mathrm s}(\cdot\mid o_{t})$}
        \State $a_{t}\gets\tau.\textsc{Action}(t)$ \Comment{expert action from $\mathcal{D}$}
        \State $\mathcal{B}_{t}\gets\textsc{Push}(\mathcal{B}_{t-1},o_{t},p_{t})$ \Comment{priority-queue update}
        \State $\ell_{t}\gets\mathcal{L}\bigl(f_{\theta^{\ast}}(\mathcal{B}_{t}),a_{t}\bigr)$ \Comment{frozen proxy objective $J_{\theta^*}(\mathcal{B}_t)$}
        \State $r_{t}\gets\beta_{\tau,t}\cdot\dfrac{\log(\ell_{t}+\lambda)}{\log(\lambda)}$ \Comment{Equation~\ref{eq:reward}}
        \State $t\gets t+1$
        \If{$t=T_{\tau}$} \Comment{episode terminated}
            \State $\tau\sim\mathcal{D}$ \Comment{load next trajectory}
            \State $\mathcal{B}_{t-1}\gets\emptyset$, $t\gets 1$
            \State $\mathrm{done}\gets\textsc{True}$
        \Else
            \State $\mathrm{done}\gets\textsc{False}$
        \EndIf
        \State $o_{t+1}\gets\tau.\textsc{Obs}(t)$
        \State \Return $(o_{t+1}, r_{t}, \mathrm{done})$
    \EndProcedure
\end{algorithmic}
\end{algorithm}

The surprise coefficient $\beta_{\tau,t}$ in Equation~\ref{eq:reward} is precomputed once
before PPO training as
$\beta_{\tau,t}=T_{\tau}\cdot
\mathrm{softmax}_{t}(J_{\theta^{\ast}}^{\tau,t}/\kappa)$,
where $J_{\theta^{\ast}}^{\tau,t}=\mathbb{E}_{\Psi_{k}}[\mathcal{L}(f_{\theta^{\ast}}(\mathcal{B}_{t}),a_{t})]$
is the expected masked-sampling loss evaluated at $(\tau,t)$. Surprise scaling is applied on a per-domain basis: we enable it for synthetic and grid domains, where the loss landscape over $t$ is sharply peaked at decision-critical timesteps, and disable it ($\beta_{\tau,t}\!\equiv\!1$) for robot manipulation domains, where the unscaled reward in Equation~\ref{eq:reward} empirically provides a more stable PPO training signal.
At training time, the environment is vectorized across $n_{\mathrm{envs}}$ parallel trajectories drawn from $\mathcal{D}$, and the proxy is evaluated on batches of buffer states for scalable RL training.

\subsection{End-to-end training and inference}
\label{app:e2e_training}

Algorithm~\ref{alg:e2e_training} describes the end-to-end training
procedure introduced in
Sections~\ref{sec:keyframe_discovery}--\ref{sec:behavior_cloning}.
The proxy $f_{\theta}$ is trained using the masked-sampling algorithm
(Algorithm~\ref{alg:masked-sampling}), then frozen and used to score the
reward $R(\mathcal{B}_{t})$ for selector RL
(Algorithm~\ref{alg:selector-env}). The trained selector is then used to produce an
augmented dataset $\mathcal{D}^{\dagger}$ by rolling $\pi_{\mathrm s}$ on
$\mathcal{D}$ with the LIFO eviction rule\footnote{For synthetic domains
where Theorem~2 holds strongly, we empirically observe
that either FIFO or LIFO eviction can be used at inference time without
affecting the discovered mnemonics; the LIFO rule is preferred for grid
and robot manipulation domains where the assumptions hold less tightly.},
which is then used to train the BC policy
$\pi_{\mathrm{BC}}(a_{t}\mid o_{t},\mathcal{B}_{t})$. See Appendix~\ref{appendix:toy_implementation} and \ref{appendix:robot_implementation} for more details.

\begin{algorithm}
\caption{Keyframe Mnemonics training}
\label{alg:e2e_training}
\textbf{Require:} dataset $\mathcal{D}$, training memory budget $k$, proxy
$f_{\theta}$, selector $\pi_{\mathrm s}$, BC policy $\pi_{\mathrm{BC}}$, training
epochs $E_{f},E_{\mathrm s},E_{\mathrm{BC}}$\\
\textbf{Returns:} trained $(f_{\theta},\pi_{\mathrm s},\pi_{\mathrm{BC}})$
\begin{algorithmic}[1]
    \Statex \texttt{// Stage 1 -- proxy policy training (Section~\ref{sec:keyframe_discovery})}
    \For{$e=1,\dots,E_{f}$}
        \State update $\theta$ on a batch sampled by Algorithm~\ref{alg:masked-sampling} on $\mathcal{D}$
    \EndFor
    \State freeze $\theta^{\ast}\gets\theta$; define
    $J_{\theta^{\ast}}(\mathcal{B})=\mathbb{E}[\mathcal{L}(f_{\theta^{\ast}}(\mathcal{B}),a)]$
    \Statex \texttt{// Stage 2 -- selector policy training (Section~\ref{sec:selector_optimization})}
    \State precompute surprise coefficients $\{\beta_{\tau,t}\}$ from $J_{\theta^{\ast}}$
    \For{$e=1,\dots,E_{\mathrm s}$}
        \State roll out the selector MDP (Algorithm~\ref{alg:selector-env}) with FIFO eviction
        \State update $\pi_{\mathrm s}$ by PPO on the surprise-weighted reward $R(\mathcal{B}_{t})$
    \EndFor
    \Statex \texttt{// Stage 3 -- BC policy training (Section~\ref{sec:behavior_cloning})}
    \State $\mathcal{D}^{\dagger}\gets\emptyset$
    \For{$\tau\in\mathcal{D}$}
        \State roll $\pi_{\mathrm s}$ on $\tau$ with LIFO eviction (Algorithm~\ref{alg:lifo_queue})
        \State append $\{(\mathcal{B}_{t},o_{t},a_{t})\}_{t=1}^{T_{\tau}}$ to $\mathcal{D}^{\dagger}$
    \EndFor
    \For{$e=1,\dots,E_{\mathrm{BC}}$}
        \State update $\pi_{\mathrm{BC}}$ on a batch from $\mathcal{D}^{\dagger}$ under the BC loss
    \EndFor
    \State \Return $(f_{\theta^*},\pi_{\mathrm s},\pi_{\mathrm{BC}})$
\end{algorithmic}
\end{algorithm}

\begin{algorithm}
\caption{Keyframe Mnemonics inference}
\label{alg:inference}
\textbf{Require:} environment $\mathcal{E}$, trained selector $\pi_{\mathrm s}$,
trained BC policy $\pi_{\mathrm{BC}}$, training memory budget $k$, max rollout steps
$T_{\max}$, queue eviction rule \textsc{Push} (Algorithm~\ref{alg:lifo_queue})
\begin{algorithmic}[1]
    \State $o_{1}\gets\mathcal{E}.\textsc{Reset}()$
    \State $\mathcal{B}_{0}\gets\emptyset$
    \For{$t=1,\dots,T_{\max}$}
        \State $p_{t}\gets\pi_{\mathrm s}(o_{t})$ \Comment{assign priority}
        \State $\mathcal{B}_{t}\gets\textsc{Push}(\mathcal{B}_{t-1},o_{t},p_{t})$ \Comment{LIFO buffer update}
        \State $a_{t}\gets\pi_{\mathrm{BC}}(\mathcal{B}_{t},o_{t})$
        \State $o_{t+1}\gets\mathcal{E}.\textsc{Step}(a_{t})$
    \EndFor
\end{algorithmic}
\end{algorithm}

At inference time the trained $(\pi_{\mathrm s},\pi_{\mathrm{BC}})$ pair is
rolled out in the target environment as specified in
Algorithm~\ref{alg:inference}. At each step the selector emits a
priority $p_{t}=\pi_{\mathrm s}(o_{t})$ that updates the buffer
$\mathcal{B}_{t}$ under the LIFO eviction rule of
Algorithm~\ref{alg:lifo_queue} (matching the rule used to construct
$\mathcal{D}^{\dagger}$), and the BC policy emits an action
$a_{t}=\pi_{\mathrm{BC}}(\mathcal{B}_{t},o_{t})$ that is applied to
the environment for at most $T_{\max}$ steps.

\section{Synthetic and Grid Domain Details}
\label{appendix:toy_implementation}

\subsection{Synthetic environments}
\label{app:synth-env}

\providecommand{\swatch}[1]{{\setlength{\fboxsep}{0pt}\fcolorbox{black!40}{#1}{\rule{0pt}{0.7em}\rule{0.9em}{0pt}}}}

In this section we show expert trajectories for the three synthetic
memory benchmarks. Cell colors group conceptually related
observations: \swatch{blue!20}\,/\,\swatch{red!20}\,/\,\swatch{orange!20}
mark the \emph{keyframes} that an oracle selector must retain,
\swatch{gray!15} marks background/noise observations that an optimal
selector should reject, and \swatch{green!20} marks the
decision-critical query timestep at which the policy emits its
action.

\paragraph{Add~\cite{hochreiter1997long}.}
A sequence of $T$ noisy real-valued observations. Two positions $t_1, t_2$ such that $t_1<t_2$ are marked with binary flags (the ``first $(1,0)$'' and ``second $(0,1)$''
summands); at the final timestep $t=T{-}1$ both flags are
simultaneously set $(1,1)$, signalling the policy to emit the sum of the two
marked values. The per-step observation is the triple
$o_t=(v_t,\;\mathbf{1}[t=t_1],\;\mathbf{1}[t=t_2])$ with
$v_t\!\sim\!\mathcal{U}[0,1]$, and the target action is the scalar
$a_t=v_{t_1}+v_{t_2}$ at $t=T{-}1$ and $a_t=0$ elsewhere. Example
with $T=6$, $t_1=1$, $t_2=3$:

\begin{center}
\setlength{\tabcolsep}{4pt}
\renewcommand{\arraystretch}{1.2}
\begin{tabular}{c|cccccc}
\toprule
$t$ & $0$ & $1$ & $2$ & $3$ & $4$ & $5$ \\
\midrule
$o_t$ & \cellcolor{gray!15}$(0.42,0,0)$ & \cellcolor{blue!20}$(0.83,1,0)$ & \cellcolor{gray!15}$(0.17,0,0)$ & \cellcolor{red!20}$(0.65,0,1)$ & \cellcolor{gray!15}$(0.91,0,0)$ & \cellcolor{green!20}$(0.0,1,1)$ \\
$a_t$ & $0$ & $0$ & $0$ & $0$ & $0$ & \cellcolor{green!20}$1.48$ \\
\bottomrule
\end{tabular}
\end{center}
The three keyframes
$\mathcal{X}_{\mathrm{key}}=\{o_1,o_3,o_5\}$ comprise the two
flag-marked summands (\swatch{blue!20}, \swatch{red!20}) and the
emit step (\swatch{green!20}); the latter is itself a keyframe
because the policy must use the emit token to know when to emit. The supervised target at the emit step
is $a_5=v_{t_1}+v_{t_2}=0.83+0.65=1.48$; all other timesteps carry
unmarked $(v_t,0,0)$ observations and are background frames.
Evaluation is done on sequences with larger $T$ obtained by increasing the number of unmarked $(v_t,0,0)$ background observations in the sequence.

\paragraph{Scattered-Copy~\cite{sorokin2022explain}.}
A subsequence of $n_\mathrm{relevant}$ relevant tokens is scattered among
$n_\mathrm{noise}$ noise tokens to form a source segment of length
$L_\mathrm{src}=n_\mathrm{relevant}+n_\mathrm{noise}$; the source is followed by a
recall segment of $n_\mathrm{relevant}$ delimiter tokens \texttt{E} scattered among
$n_\mathrm{blank}$ blank tokens $\square$ (length
$L_\mathrm{rec}=n_\mathrm{relevant}+n_\mathrm{blank}$). At each delimiter the policy
must emit the next relevant token in the order it originally appeared,
and a no-op token \texttt{0} at every other position. The set of tokens for relevant and noise are disjoint and remains fixed for the task during both training and evaluation. Example with
$n_\mathrm{relevant}=3$, $n_\mathrm{noise}=3$, $n_\mathrm{blank}=5$:

\begin{center}
\setlength{\tabcolsep}{3pt}
\renewcommand{\arraystretch}{1.1}
\begin{tabular}{c|cccccc|cccccccc}
\toprule
$t$    & $0$ & $1$ & $2$ & $3$ & $4$ & $5$ & $6$ & $7$ & $8$ & $9$ & $10$ & $11$ & $12$ & $13$ \\
\midrule
$o_t$  & \cellcolor{gray!15}\texttt{N$_1$} & \cellcolor{blue!20}\texttt{A} & \cellcolor{gray!15}\texttt{N$_2$} & \cellcolor{red!20}\texttt{B} & \cellcolor{gray!15}\texttt{N$_3$} & \cellcolor{orange!20}\texttt{C} & \cellcolor{gray!15}$\square$ & \cellcolor{gray!15}$\square$ & \cellcolor{green!20}\texttt{E} & \cellcolor{gray!15}$\square$ & \cellcolor{green!20}\texttt{E} & \cellcolor{gray!15}$\square$ & \cellcolor{gray!15}$\square$ & \cellcolor{green!20}\texttt{E} \\
$a_t$ & \texttt{0} & \texttt{0} & \texttt{0} & \texttt{0} & \texttt{0} & \texttt{0} & \texttt{0} & \texttt{0} & \cellcolor{blue!20}\texttt{A} & \texttt{0} & \cellcolor{red!20}\texttt{B} & \texttt{0} & \texttt{0} & \cellcolor{orange!20}\texttt{C} \\
\bottomrule
\end{tabular}
\end{center}
The six
keyframes
$\mathcal{X}_{\mathrm{key}}=\{o_1,o_3,o_5,o_8,o_{10},o_{13}\}$
comprise the three relevant source tokens
(\swatch{blue!20}\,/\,\swatch{red!20}\,/\,\swatch{orange!20}) and the
three delimiter tokens (\swatch{green!20}); the delimiters are
themselves keyframes because the policy must use an \texttt{E}
token to know it is at an emit position. The noise
tokens \texttt{N}$_i$ in the source and blank tokens in the
recall are background frames. Evaluation is done by increasing $n_\mathrm{noise}$ and $n_\mathrm{blank}$ in the sequence.

\paragraph{Active T-Maze~\cite{bakker2001reinforcement}.}
The agent is placed at the start position $x=1$ in a corridor of
length $L$ with the oracle at $x=0$ and the T-junction at $x=L+1$.
The goal $\texttt{g}\in\{+1,-1\}$ is observable only at the oracle
position (cue observed at $x=0 \rightarrow (0,\texttt{g})$); elsewhere along the corridor the
observation is the ambiguous $(0,0)$. At the T-junction the
observation is $(1,0)$ and the agent must choose to turn up or down
based on the remembered cue. The action is the 2D displacement
$a_t\in\{(1,0),(-1,0),(0,1),(0,-1)\}$ (\texttt{right}/\texttt{left}/\texttt{up}/\texttt{down}).
\begin{center}
\setlength{\tabcolsep}{2pt}
\renewcommand{\arraystretch}{1.0}
\begin{tabular}{cccccc}
& & & & & \cellcolor{blue!20}$1{+}$ \\
\cellcolor{blue!20}\texttt{g=${\pm}1$} & $s$ & $\cdot$ & $\cdot$ & $\cdot$ & \cellcolor{green!20}\texttt{J} \\
& & & & & \cellcolor{blue!20}$1{-}$ \\
\end{tabular}
\end{center}
where $s$ is the start, \texttt{J} the T-junction, $\{1+, 1-\}$ the
two goal arms, and the three dots between $s$ and \texttt{J} are a
schematic ellipsis denoting a variable-length corridor of $L$
ambiguous $(0,0)$ cells; the agent must traverse $s\!\to\!\texttt{g}
\!\to\!s\!\to\!\cdots\!\to\!\texttt{J}\!\to\! 1{\pm}$ correctly to
succeed. Example with $L=4$ and $\texttt{g}=+1$ is

\begin{center}
\setlength{\tabcolsep}{4pt}
\renewcommand{\arraystretch}{1.2}
\begin{tabular}{c|ccccccc}
\toprule
$t$ & $0$ & $1$ & $2$ & $3$ & $4$ & $5$ & $6$ \\
\midrule
$o_t$ & \cellcolor{gray!15}$(0,0)$ & \cellcolor{blue!20}$(0,+1)$ & \cellcolor{gray!15}$(0,0)$ & \cellcolor{gray!15}$(0,0)$ & \cellcolor{gray!15}$(0,0)$ & \cellcolor{gray!15}$(0,0)$ & \cellcolor{green!20}$(1,0)$ \\
$a_t$ & $(-1,0)$ & $(1,0)$ & $(1,0)$ & $(1,0)$ & $(1,0)$ & $(1,0)$ & \cellcolor{blue!20}$(0,+1)$ \\
\bottomrule
\end{tabular}
\end{center}
The two keyframes $\mathcal{X}_{\mathrm{key}}=\{o_1,o_6\}$ are
the cue exposure at $t=1$ (\swatch{blue!20}) and the T-junction
observation at $t=6$ (\swatch{green!20}); the junction is itself a
keyframe because the policy must use the $(1,0)$ token to know when to turn into a goal arm. At the junction the
agent must emit $(0,+1)$ if $\texttt{g}=+1$ and $(0,-1)$ if $\texttt{g}=-1$. The
ambiguous $(0,0)$ corridor observations at all other timesteps
(\swatch{gray!15}) are background frames. Evaluation is done by stretching
the gap of $(0,0)$ corridor observations between the cue \texttt{g} and the junction \texttt{J}.

\subsection{Grid environments}
\label{app:grid-env}
Both grid benchmarks share the MiniGrid~\cite{chevalier2023minigrid} interface.

\begin{figure}[h]
    \centering
    \includegraphics[width=0.5\linewidth]{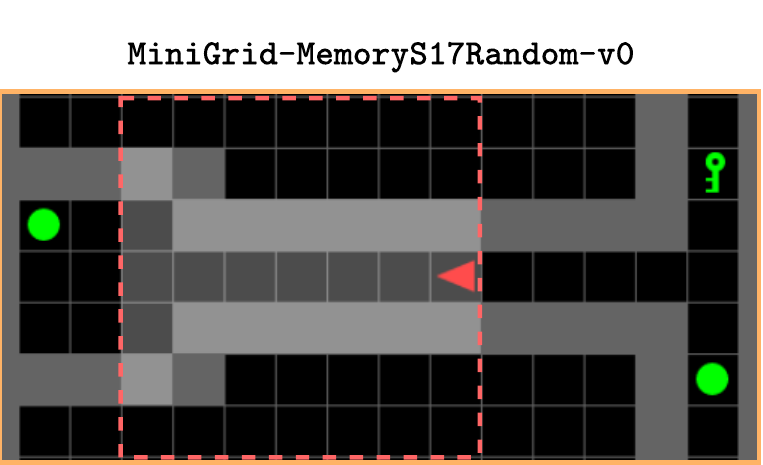}
\caption{\textbf{MiniGrid-Memory environment.} The agent (red triangle) observes only a local ($7\times7$) view of the grid environment (dotted red box). It must navigate to the left room to observe the cue (green ball), then make a U-turn and traverse the corridor to the junction, where it must decide whether to move up or down. To succeed, the agent must move toward the same object observed in the left room.}
    \label{fig:minigridmem_env}
\end{figure}

\begin{figure}[h]
    \centering
    \includegraphics[width=0.75\linewidth]{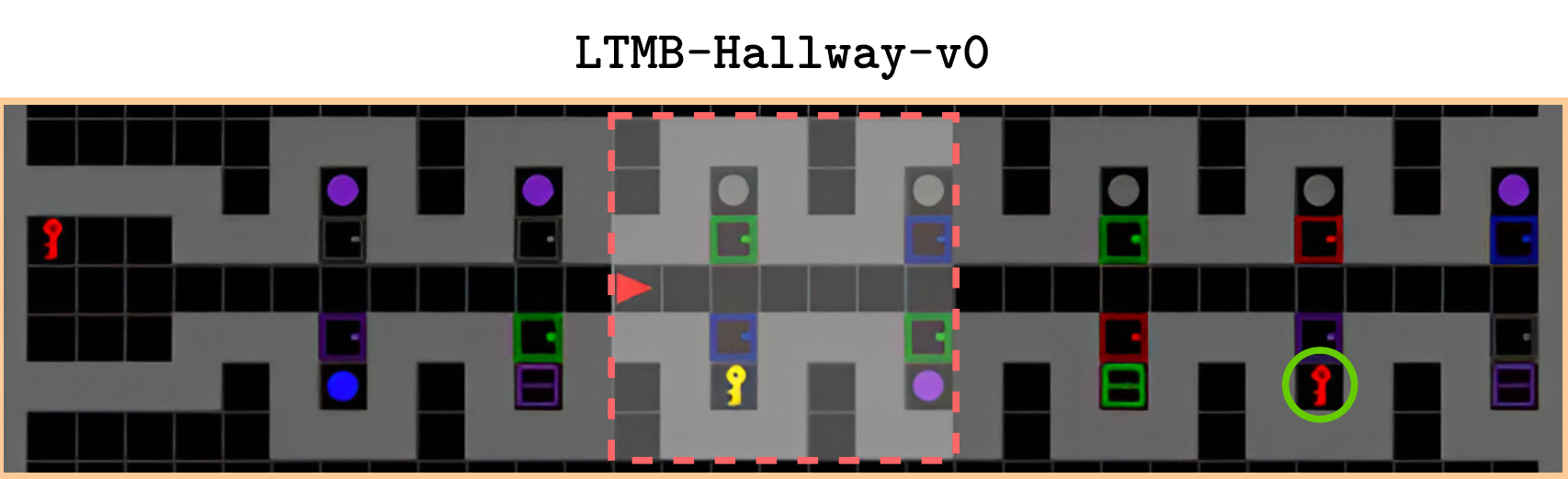}
\caption{\textbf{LTMB-Hallway environment.} The agent (red triangle) observes only a local ($7\times7$) view of the grid environment (dotted red box). It must remember the cue in the left room and then traverse the hallway while matching the remembered cue to the objects behind the locked doors. To succeed, it must open and enter the door containing the same object as the one in the left room. In the example above, the correct target is highlighted with a green circle.}
    \label{fig:ltmb_env}
\end{figure}

\paragraph{MiniGrid-Memory~\cite{chevalier2023minigrid}.}
A green \texttt{Key} or \texttt{Ball} is placed in a start room
attached to a long corridor; the corridor ends in a T-junction with
one \texttt{Key} arm and one \texttt{Ball} arm (order randomised) (Figure~\ref{fig:minigridmem_env}).
The agent spawns at a random cell along the corridor and must reach
the arm matching the start-room object; stepping into the wrong arm
terminates with task failure. The keyframe is the start-room object
observation (cue); all corridor
observations are background frames.
Evaluation is done stretching the corridor length.

\paragraph{LTMB-Hallway~\cite{yue2024learning}.}
A colored target object (\texttt{Key}, \texttt{Ball}, or \texttt{Box} in one of six colors) is placed in the start room (Figure~\ref{fig:ltmb_env}). A horizontal corridor leads past several branching vertical hallways, each containing random objects behind colored doors. Exactly one branch object matches the target in both shape and color, and the agent must open and enter the corresponding hallway door to succeed. Entering any other door results in task failure. The keyframe is the start-room target observation (cue), which must be remembered and matched with hallway targets, while the other corridor observations are background frames. Evaluation is performed by increasing the number of distractor hallways.

\subsection{Training dataset}
\label{app:training_dataset}
All
training datasets contain expert demonstrations only, with no failure
trajectories. This matches the requirements of Keyframe Mnemonics, which needs an expert dataset for training. For each baseline we trained two variants per
benchmark, differing only in the training-distribution scale: a
\emph{small} variant matching the original demonstration distribution
of $\mathcal{D}$ used for training Keyframe
Mnemonics, and a \emph{large} variant in which the
trajectory horizon and the number of variations are scaled up.
Training sequence lengths and number of demonstrations ($|\mathcal{D}|$, $N$ in Lemma~1) used in both variants are summarized in
Table~\ref{tab:dataset-seqlens}. Policies trained on both variants are evaluated on the
same out-of-distribution test distribution, and the variant that
achieves the higher test success rate (or lower test MSE for Add) is
considered for benchmarking. Some baselines benefit from
the larger distribution and from more diverse training data (e.g.\ DT
and TrXL on Scattered-Copy), while others struggle on the \emph{large}
variant and generalize better when trained on the \emph{small} dataset (e.g.\
LSTM-BPTT on T-Maze). See Appendix~\ref{appendix:benchmark_details} for more details.

As discussed in Appendix~\ref{app:proxy-training}, we run multiple offline passes over the dataset $\mathcal{D}$ during masked-sampling training of the proxy policy. The number of passes is denoted by $\mathrm{pass}_{\mathrm{ms}}$ in Table~\ref{tab:dataset-seqlens}. We determine this value by starting from a small constant and increasing it by orders of $10$ until the requirement on $M$ in Theorem~1 is satisfied. The x-axis multipliers in Figure~\ref{fig:keyframe-empirics} scale $\mathrm{pass}_{\mathrm{ms}}$; for example, $M=0.1\times$ for the Add domain means that $\mathrm{pass}_{\mathrm{ms}}=0.1\times300=30$ passes were used to train the proxy policy.

\begin{table}[h]
\centering
\small
\caption{Training and test sequence lengths and number of demonstrations
$|\mathcal{D}|$ per dataset variant. MiniGrid-Memory and LTMB-Hallway
use a single dataset (no \emph{large} variant). In the ``Seq len''
columns, ``$a$--$b$'' denotes a range (lower--upper bound) and
``${\sim}n$'' denotes the maximum sequence length. The last column $\mathrm{pass}_{\mathrm{ms}}$
reports the number of offline rollout passes used to create the
masked-sampling cache for proxy training
(Appendix~\ref{app:proxy-training}).}
\label{tab:dataset-seqlens}
\begin{tabular}{l cc cc c c}
\toprule
& \multicolumn{2}{c}{\textbf{Train} (\emph{small})$^{*}$}
& \multicolumn{2}{c}{\textbf{Train} (\emph{large})}
& \textbf{Test} & \\
\cmidrule(lr){2-3} \cmidrule(lr){4-5} \cmidrule(lr){6-6} \cmidrule(lr){7-7}
\textbf{Benchmark}   & Seq len    & $|\mathcal{D}|$ & Seq len     & $|\mathcal{D}|$ & Seq len      & $\mathrm{pass}_{\mathrm{ms}}$ \\
\midrule
Add             & $5$--$10$  & $1000$     & $20$--$30$  & $5000$     & $50$--$100$  & $300$ \\
Scattered-Copy   & $4$        & $1000$     & $40^{*}$    & $5000$     & $220$        & $30$  \\
T-Maze      & $5$--$10$  & $100$      & $55$--$105$ & $5000$     & ${\sim}1005$ & $300$ \\
MiniGrid-Memory & ${\sim}21$ & $2000$   & ---         & ---        & ${\sim}2012$ & $300$ \\
LTMB-Hallway       & ${\sim}46$ & $50{,}000$ & ---         & ---        & ${\sim}425$  & $10$  \\
\bottomrule
\addlinespace[2pt]
\multicolumn{7}{l}{\footnotesize $^{*}$ Dataset used for training Keyframe Mnemonics.} \\
\end{tabular}
\end{table}

\subsection{Training Keyframe Mnemonics policies}
\label{appendix:km_training_details}
Keyframe Mnemonics (KM) uses three networks per benchmark (Algorithm~\ref{alg:e2e_training}): a proxy policy $f_{\theta}$ trained by
masked-sampling, a selector $\pi_{\mathrm s}$ policy trained with PPO, and a BC policy $\pi_{\mathrm{BC}}$.
Per-benchmark architecture and PPO hyperparameters are summarised in
Tables~\ref{tab:km-arch} and~\ref{tab:km-ppo}.

\begin{table}[h]
\centering
\small
\caption{KM module architectures per benchmark.
$k$: memory budget; $d_{\mathrm{feat}}$: selector
feature dimension; $d_{\mathrm{hid}}$: proxy / BC hidden width;
$H$: action-chunk horizon; $L$: proxy / BC hidden layers. Cells with two
values $a/b$ are proxy / BC policy. Architecture of proxy and BC policy can be different per-domain basis.}
\label{tab:km-arch}
\begin{tabular}{l cccccc}
\toprule
\textbf{Benchmark} & $k$ & $d_{\mathrm{feat}}$ & Proxy / BC arch                          & $d_{\mathrm{hid}}$ & $H$  & $L$ \\
\midrule
Add                & $3$ & $32$                & ReLU MLP                            & $256$              & $1$  & $3$ \\
Scattered-Copy     & $2/20$ & $32$             & ReLU MLP / Transformer       & $256/64$           & $1$  & $3/4$ \\
T-Maze             & $2$ & $16$                & ReLU MLP                            & $256$              & $1$  & $3$ \\
MiniGrid-Memory    & $2/1$ & $64$              & CNN and Transformer                  & $128/64$           & $1$  & $4$ \\
LTMB-Hallway       & $2/1$ & $64$              & CNN and Transformer                  & $128/64$           & $8/1$ & $4$ \\
\bottomrule
\end{tabular}
\end{table}

\begin{table}[h]
\centering
\small
\caption{KM selector optimization settings per
benchmark. PPO uses Adam with the SB3~\cite{stable-baselines3} defaults except where shown
($n_{\mathrm{envs}}$ parallel selector environments, total environment steps$_{\mathrm{ppo}}$, rollout length $n_{\mathrm{steps}}$, batch
size, $n_{\mathrm{epochs}}$ per update, learning rate $\eta_{\pi}$,
and reward normalisation constant $\lambda$ from Equation~\ref{eq:reward}). The value of $\lambda$ depends on the type of loss function used for reward objective. The proxy is trained with AdamW (learning rate $\eta_{f}$, batch
size).}
\label{tab:km-ppo}
\begin{tabular}{l ccccccc cc}
\toprule
& \multicolumn{7}{c}{\textbf{Selector PPO}} & \multicolumn{2}{c}{\textbf{Proxy}} \\
\cmidrule(lr){2-8} \cmidrule(lr){9-10}
\textbf{Benchmark} & $n_{\mathrm{envs}}$ & steps$_{\mathrm{ppo}}$ & $n_{\mathrm{steps}}$ & batch & $n_{\mathrm{epochs}}$ & $\eta_{\pi}$ & $\lambda$ & $\eta_{f}$ & batch \\
\midrule
Add             & $4$  & $2\!\times\!10^{5}$ & $2048$ & $64$  & $10$ & $3\!\times\!10^{-4}$ & $10^{-6}$ & $10^{-3}$ & $256$ \\
Scattered-Copy  & $4$  & $2\!\times\!10^{5}$ & $2048$ & $64$  & $10$ & $3\!\times\!10^{-4}$ & $10^{-3}$ & $10^{-3}$ & $256$ \\
T-Maze          & $16$ & $2\!\times\!10^{5}$ & $512$  & $64$  & $10$ & $3\!\times\!10^{-4}$ & $10^{-6}$ & $10^{-3}$ & $256$ \\
MiniGrid-Memory & $32$ & $3\!\times\!10^{5}$ & $512$  & $128$ & $5$  & $3\!\times\!10^{-4}$ & $10^{-2}$ & $10^{-3}$ & $512$ \\
LTMB-Hallway    & $32$ & $2\!\times\!10^{5}$ & $1024$ & $256$ & $5$  & $3\!\times\!10^{-4}$ & $10^{-2}$ & $10^{-3}$ & $512$ \\
\bottomrule
\end{tabular}
\end{table}

In Table~\ref{tab:km-arch}, note that our formulation allows flexibility in the memory budget $k$ used for proxy and BC policy training, since the inference buffer size can be adjusted after selector training to store a different number of keyframes than what was used during proxy training. The memory budget for the proxy and BC policies is selected empirically for each benchmark, guided by the expected number of task-relevant keyframes. We try a small set of candidate values around this estimate and choose the smallest $k$ that yields stable proxy policy convergence and produces qualitatively valid mnemonics after selector training. See Appendix~\ref{app:additional_analysis} for detailed analysis.

Also note that in Table~\ref{tab:km-arch}, the proxy policy uses different network architectures across benchmarks. We find that Theorem~1 holds generally, irrespective of network architecture; for instance, the synthetic domains use MLPs whereas grid-world models use CNNs and Transformers. Our choice of ReLU MLPs in synthetic domains is motivated by prior empirical evidence showing that such models exhibit a stronger simplicity bias \cite{de2019random}.

\subsection{Baselines and benchmarking details}
\label{app:syn-grid-arch}

The five benchmarks (Add, Scattered-Copy, T-Maze, MiniGrid-Memory,
LTMB-Hallway) share a common trainer pipeline; only the per-baseline
hyperparameters and per-task input/output dimensions differ. All baselines
are trained with AdamW, a linear warm-up cosine learning-rate schedule
(10\%\ of total steps warmup, decaying to $1\%$ of the peak rate), and
gradient clipping. Our baseline design and implementation choices are derived from the open-source code of RATE~\cite{cherepanov2023recurrent}. Additionally, we tuned the training recipe to obtain the best possible baseline results on our benchmarking domains.

\begin{table}[h]
\centering
\small
\caption{Trainable parameter counts per benchmark. KM totals sum the
proxy, selector and BC policy modules. The baseline architectural choices that determine these
counts (layer widths, head counts, embedding dimensions) are inherited
from the RATE~\cite{cherepanov2023recurrent} open-source
implementation. Units: K = $10^{3}$, M = $10^{6}$.}
\label{tab:params-comparison}
\begin{tabular}{l rrrrrrr}
\toprule
\textbf{Benchmark} & LSTM & DLSTM & DT & DMamba & TrXL & ELMUR & \textbf{KM (Ours)} \\
\midrule
Add             & $100\mathrm{K}$ & $740\mathrm{K}$ & $182\mathrm{K}$ & $394\mathrm{K}$ & $182\mathrm{K}$ & $4.4\mathrm{M}$ & $\mathbf{268\mathrm{K}}$ \\
Scattered-Copy  & $102\mathrm{K}$ & $742\mathrm{K}$ & $184\mathrm{K}$ & $396\mathrm{K}$ & $184\mathrm{K}$ & $4.4\mathrm{M}$ & $\mathbf{366\mathrm{K}}$ \\
T-Maze          & $100\mathrm{K}$ & $740\mathrm{K}$ & $246\mathrm{K}$ & $395\mathrm{K}$ & $246\mathrm{K}$ & $4.4\mathrm{M}$ & $\mathbf{266\mathrm{K}}$ \\
MiniGrid-Memory & $641\mathrm{K}$ & $1.9\mathrm{M}$ & $2.0\mathrm{M}$ & $1.5\mathrm{M}$ & $997\mathrm{K}$ & $4.7\mathrm{M}$ & $\mathbf{1.44\mathrm{M}}$ \\
LTMB-Hallway    & $641\mathrm{K}$ & $1.9\mathrm{M}$ & $2.0\mathrm{M}$ & $1.5\mathrm{M}$ & $997\mathrm{K}$ & $4.7\mathrm{M}$ & $\mathbf{1.45\mathrm{M}}$ \\
\bottomrule
\end{tabular}
\end{table}

\begin{table}[h]
\centering
\small
\caption{Decomposition of the Keyframe Mnemonics parameter budget into
the proxy $f_{\theta}$, the selector policy $\pi_{\mathrm s}$, and the BC policy $\pi_{\mathrm{BC}}$.}
\label{tab:km-params-decomp}
\begin{tabular}{l rrr r}
\toprule
\textbf{Benchmark} & Proxy $f_{\theta}$ & Selector $\pi_{\mathrm s}$ & BC policy $\pi_{\mathrm{BC}}$ & \textbf{Total} \\
\midrule
Add             & $134\mathrm{K}$ & $128$           & $134\mathrm{K}$             & $\mathbf{268\mathrm{K}}$ \\
Scattered-Copy  & $157\mathrm{K}$ & $2.7\mathrm{K}$ & $206\mathrm{K}$ & $\mathbf{366\mathrm{K}}$ \\
T-Maze          & $133\mathrm{K}$ & $48$            & $133\mathrm{K}$             & $\mathbf{266\mathrm{K}}$ \\
MiniGrid-Memory & $957\mathrm{K}$ & $242\mathrm{K}$ & $245\mathrm{K}$ & $\mathbf{1.44\mathrm{M}}$ \\
LTMB-Hallway    & $959\mathrm{K}$ & $242\mathrm{K}$ & $245\mathrm{K}$ & $\mathbf{1.45\mathrm{M}}$ \\
\bottomrule
\end{tabular}
\end{table}

\subsection{Parameter-budget comparison}
\label{app:param-budget}
Table~\ref{tab:params-comparison} compares the total trainable parameter
count of every method per benchmark. For Keyframe Mnemonics we report
the sum of the proxy, selector, and BC policy
parameters.
Despite using three networks, the KM total remains well below most of the baselines demonstrating that the
architectural decomposition does not come at a parameter cost. Table~\ref{tab:km-params-decomp} breaks the KM total down by module. The
selector policy is compact (under $1\%$ of the total on
the synthetic tasks and roughly $15\%$ on the grid tasks), since it
only encodes a single observation at a time; most of the budget sits in
the proxy and, the BC policy.

\subsection{Benchmark result analysis}
\label{appendix:benchmark_details}
We report per-benchmark in-distribution and out-of-distribution
performance for every method in Table~\ref{tab:results}. For each baseline we
report the better of policy performance trained on \emph{small} /
\emph{large} dataset variants (Appendix~\ref{app:training_dataset}); Keyframe Mnemonics is trained only on
the \emph{small} variant (except Scattered-Copy that uses
\emph{large}\footnote{For Scattered-Copy, the proxy policy (Stage~1)
and selector policy (Stage~2) are trained on the \emph{small} variant, while the BC policy (Stage~3) is trained on the
\emph{large} variant.}). The selected variant per (benchmark, baseline) pair is as follows. On
Add, the \emph{small} variant is used for LSTM(BPTT), DLSTM,
and the \emph{large} variant for LSTM(TBPTT), DT, DMamba, TrXL, and
ELMUR. On Scattered-Copy, the \emph{small} variant is used for
LSTM(BPTT) and LSTM(TBPTT), and the \emph{large} variant for all
remaining baselines. On T-Maze, the \emph{large} variant is used for
LSTM(TBPTT), DT, and TrXL, and the \emph{small} variant for all
remaining baselines. The grid benchmarks (MiniGrid-Memory and
LTMB-Hallway) use a single dataset.

\begin{table}[h]
\centering
\caption{Aggregated benchmark results across all five tasks. Cells show mean~$\pm$~std across $5$ seeds for Add, Scattered-Copy, and T-Maze, and $3$ seeds for MiniGrid-Memory and LTMB-Hallway. For Add, the metric is final-step MSE ($\downarrow$ better); for all others, it is success rate ($\uparrow$ better). ``---'' marks the LSTM~(BPTT) variant on the grid tasks, which we did not train because it required significantly more compute than the other baselines; only TBPTT was trained. See Table~\ref{tab:dataset-seqlens} for train and test sequence lengths and the number of demonstrations $|\mathcal{D}|$ used for training.}
\label{tab:results}
\providecommand{\E}[1]{{\scriptscriptstyle\,\cdot\,10^{#1}}}
\resizebox{\textwidth}{!}{%
\setlength{\tabcolsep}{3pt}
\begin{tabular}{l rrrrrrrr}
\toprule
& LSTM(B) & LSTM(T) & DLSTM & DT & DMamba & TrXL & ELMUR & \textbf{KM (Ours)} \\
\midrule
\multicolumn{9}{l}{\textbf{Add}} \\
Train MSE $\downarrow$ & $\mathbf{4\E{-6} \!\pm\! 1\E{-6}}$ & $0.006 \!\pm\! 0.003$ & $0.167 \!\pm\! 0.106$ & $\mathbf{8.1\E{-4} \!\pm\! 6.8\E{-4}}$ & $\mathbf{2.0\E{-4} \!\pm\! 1.3\E{-4}}$ & $\mathbf{2.2\E{-4} \!\pm\! 1.2\E{-4}}$ & $0.114 \!\pm\! 0.014$ & $\mathbf{3.4\E{-5} \!\pm\! 3.6\E{-5}}$ \\
Test MSE $\downarrow$  & $0.008 \!\pm\! 0.004$ & $0.040 \!\pm\! 0.024$ & $0.368 \!\pm\! 0.186$ & $0.148 \!\pm\! 0.083$ & $0.160 \!\pm\! 0.086$ & $0.133 \!\pm\! 0.064$ & $0.157 \!\pm\! 0.047$ & $\mathbf{3.2\E{-5} \!\pm\! 3.3\E{-5}}$ \\
\midrule
\multicolumn{9}{l}{\textbf{Scattered-Copy}} \\
Train Copy SR $\uparrow$ & $\mathbf{1.000 \!\pm\! 0.000}$ & $0.100 \!\pm\! 0.035$ & $0.000 \!\pm\! 0.000$ & $0.538 \!\pm\! 0.303$ & $0.000 \!\pm\! 0.000$ & $0.018 \!\pm\! 0.016$ & $0.000 \!\pm\! 0.000$ & $\mathbf{1.000 \!\pm\! 0.000}$ \\
Test Copy SR $\uparrow$  & $0.160 \!\pm\! 0.021$ & $0.106 \!\pm\! 0.030$ & $0.000 \!\pm\! 0.000$ & $0.000 \!\pm\! 0.000$ & $0.000 \!\pm\! 0.000$ & $0.000 \!\pm\! 0.000$ & $0.000 \!\pm\! 0.000$ & $\mathbf{1.000 \!\pm\! 0.000}$ \\
\midrule
\multicolumn{9}{l}{\textbf{T-Maze}} \\
Train SR $\uparrow$ & $\mathbf{1.000 \!\pm\! 0.000}$ & $0.330 \!\pm\! 0.032$ & $0.530 \!\pm\! 0.055$ & $\mathbf{1.000 \!\pm\! 0.000}$ & $0.550 \!\pm\! 0.032$ & $\mathbf{1.000 \!\pm\! 0.000}$ & $0.682 \!\pm\! 0.156$ & $\mathbf{1.000 \!\pm\! 0.000}$ \\
Test SR $\uparrow$  & $0.828 \!\pm\! 0.236$ & $0.314 \!\pm\! 0.078$ & $0.478 \!\pm\! 0.058$ & $0.518 \!\pm\! 0.043$ & $0.450 \!\pm\! 0.029$ & $0.500 \!\pm\! 0.047$ & $0.018 \!\pm\! 0.035$ & $\mathbf{1.000 \!\pm\! 0.000}$ \\
\midrule
\multicolumn{9}{l}{\textbf{MiniGrid-Memory}} \\
Train SR $\uparrow$ & --- & $0.823 \!\pm\! 0.214$ & $0.747 \!\pm\! 0.067$ & $0.553 \!\pm\! 0.050$ & $0.203 \!\pm\! 0.006$ & $0.483 \!\pm\! 0.035$ & $0.653 \!\pm\! 0.025$ & $\mathbf{0.993 \!\pm\! 0.012}$ \\
Test SR $\uparrow$  & --- & $0.497 \!\pm\! 0.072$ & $0.530 \!\pm\! 0.069$ & $0.547 \!\pm\! 0.025$ & $0.030 \!\pm\! 0.000$ & $0.487 \!\pm\! 0.031$ & $0.300 \!\pm\! 0.100$ & $\mathbf{0.997 \!\pm\! 0.006}$ \\
\midrule
\multicolumn{9}{l}{\textbf{LTMB-Hallway}} \\
Train SR $\uparrow$ & --- & $0.037 \!\pm\! 0.064$ & $0.077 \!\pm\! 0.031$ & $0.217 \!\pm\! 0.112$ & $0.000 \!\pm\! 0.000$ & $0.123 \!\pm\! 0.015$ & $0.070 \!\pm\! 0.000$ & $\mathbf{1.000 \!\pm\! 0.000}$ \\
Test SR $\uparrow$  & --- & $0.003 \!\pm\! 0.006$ & $0.003 \!\pm\! 0.006$ & $0.013 \!\pm\! 0.006$ & $0.000 \!\pm\! 0.000$ & $0.010 \!\pm\! 0.000$ & $0.010 \!\pm\! 0.000$ & $\mathbf{0.993 \!\pm\! 0.006}$ \\
\bottomrule
\end{tabular}%
}
\vspace{2pt}
\\{\scriptsize Random baseline omitted; its test values are
$0.499\!\pm\!0.061$ test MSE on Add, $0.000\!\pm\!0.000$ test copy SR
on Scattered-Copy, $0.000\!\pm\!0.000$ test SR on T-Maze,
$0.030\!\pm\!0.010$ test SR on MiniGrid-Memory, and
$0.000\!\pm\!0.000$ test SR on LTMB-Hallway.}
\end{table}

\paragraph{Memory blur issue.} In T-Maze, sequence models such as LSTM~(BPTT) and TrXL fit the training distribution well and achieve a train SR of $1.0$, but collapse to near-chance performance on longer test horizons. We observe a similar trend in the Add domain. Recurrent memory baselines struggle in these domains because their hidden states or external memories are repeatedly updated over long horizons. Keyframe Mnemonics avoids these failure modes by retaining only keyframes which are relevant for the task which remain unaffected to change in horizon length.

A more severe case is Scattered-Copy, which requires remembering multiple numbers and emitting them in the order they were observed. This task is particularly difficult for baselines that repeatedly update hidden states or external memories, as such updates blur information over time and fail to generalize to longer OOD sequences. In contrast, the KM selector learns to identify relevant keyframes during Stage~2 of training and stores only these keyframes in the buffer, making it easier for the BC policy to learn the copy operation over buffer elements while ignoring noise tokens in the sequence. This highlights a key advantage of our method; since the buffer stores individual tokens, it avoids information blur.

\paragraph{Non-memory baseline.} Among the baselines, Decision Transformer (DT) and DMamba do not include an explicit memory mechanism. We include them for completeness in our benchmark evaluation. The observed results are expected, as these models are limited by the context window length used during training; once the sequence length exceeds this window, relevant information falls out of context and the models can no longer predict the correct action. Consequently, they perform well on the training distribution but degrade to near-chance performance on the test distribution.

\subsection{Hardware and wall-clock}
\label{app:syn-grid-hardware}
All training runs were executed on a single NVIDIA RTX~3090
($24$\,GB) GPU. The three-stage Keyframe Mnemonics pipeline (proxy training, selector
PPO, BC policy training) completes across all five benchmarks in $\sim3$\,h per seed, with the
grid benchmarks accounting for the majority of the wall-clock time. The cumulative training time for all baselines on a single synthetic benchmark is
under $\sim30$\,minutes per seed. The cumulative baseline training time is
$\sim4$\,h per seed on MiniGrid-Memory and $\sim6$\,h per seed on
LTMB-Hallway.

\section{Robot Manipulation Domain Details}
\label{appendix:robot_implementation}

This section provides details on the hyperparameters and architectural choices
used to produce the \texttt{MIKASA-Robo} results in
Section~\ref{sec:experiments}. For the \texttt{Robomimic} domain experiments (Figure~\ref{fig:markovian}), we use the same configuration as \texttt{MIKASA-Robo}. Values not stated here take the
standard defaults of the underlying libraries.

\subsection{\texttt{MIKASA-Robo} environments}
\label{app:mikasa-env}

\begin{figure}
    \centering
    \includegraphics[width=0.9\linewidth]{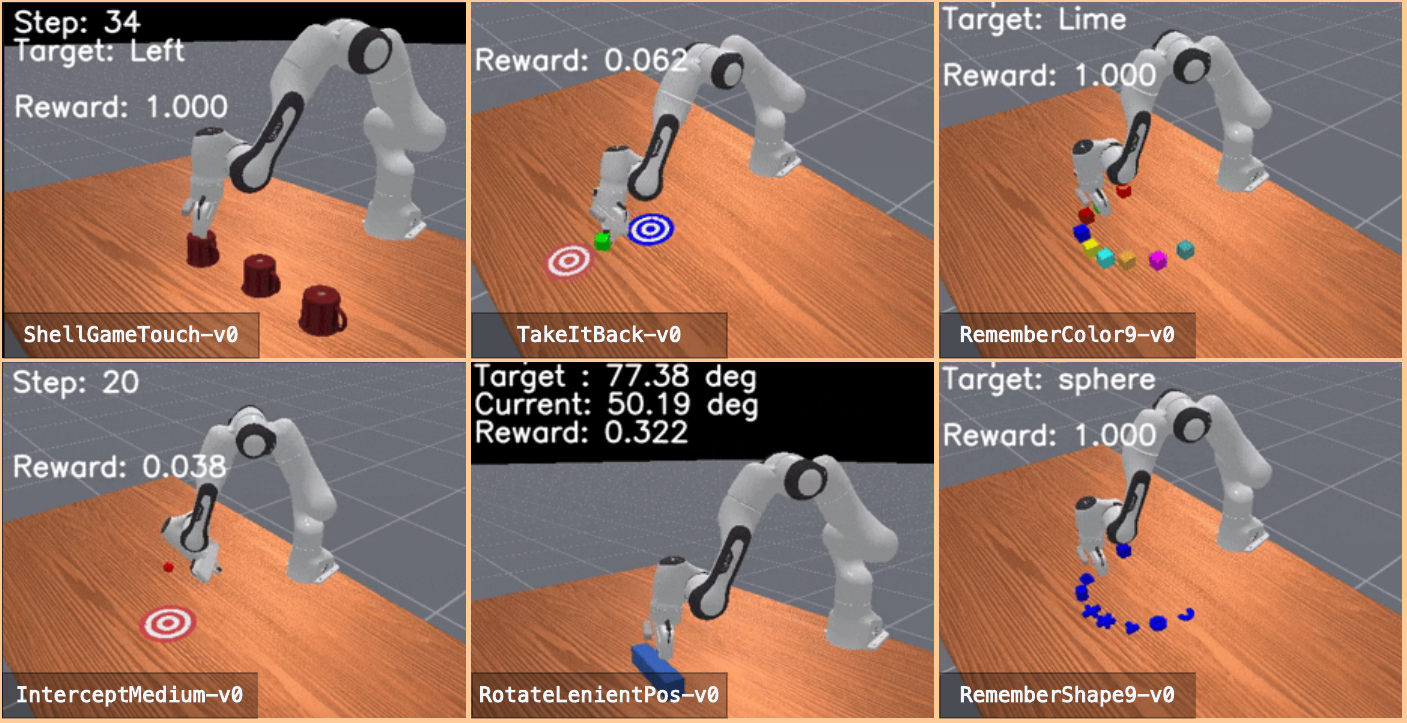}
\caption{\textbf{MIKASA-Robo Environments.} A collage of six representative tasks from the \texttt{MIKASA-Robo} memory-intensive robot manipulation benchmark. The benchmark evaluates robot policies across comprehensive memory categories, including object memory, spatial memory, sequential memory, and memory capacity. More details can be found in the reference paper~\cite{cherepanov2025memory}.}
    \label{fig:mikasa_env}
\end{figure}

A set of six representative tasks from the benchmark is shown in Figure~\ref{fig:mikasa_env}.

\subsection{Pipeline overview}
\label{app:training:pipeline}

Each \texttt{MIKASA-Robo} task is trained in three stages (Algorithm~\ref{alg:e2e_training}), following
Sections~\ref{sec:keyframe_discovery}--\ref{sec:behavior_cloning}
of the main paper. \textbf{(i)}~The \emph{proxy policy} $f_\theta$ is
trained under masked-sampling of past observations, after which its
frozen prediction loss serves as the keyframe-discovery objective
$J_{\theta^*}(\mathcal B)$. \textbf{(ii)}~The \emph{selector
policy} $\pi_\mathrm{s}$ is trained with PPO against the reward derived from $J_{\theta^*}$, learning
to assign high priority to keyframes. \textbf{(iii)}~The \emph{BC
policy} $\pi_\mathrm{BC}(a_t \mid o_t, \mathcal{B}_t)$ is trained on
the augmented dataset $\mathcal D^\dagger$, in which every observation
$o_t$ is paired with the priority queue $\mathcal B_t$ produced by
$\pi_\mathrm{s}$. All three stages use only the offline expert
demonstrations $\mathcal{D}$; the simulator is never queried during
proxy or selector training. The memory
budget is fixed at $k = 3/2$ (proxy / BC policy) throughout, so that $\pi_\mathrm{BC}$ sees
the current observation alongside two mnemonics drawn from the queue.

\subsection{Architectures and parameter counts}
\label{app:training:arch}

All modules share a \textsc{ResNet}-18 visual encoder, initialised from
\textsc{ImageNet} weights and then trained end-to-end. Camera frames are
$128{\times}128$ RGB; the proprioceptive state concatenates the TCP
pose, joint positions, and joint velocities into a $25$-dimensional
vector. Both overhead and wrist cameras are used.
Table~\ref{tab:arch} summarises the architecture and parameter count
of each module.

\begin{table}[h]
\caption{Module architectures, dimensions and parameter counts. $d$:
hidden width, $h$: attention heads, $L$: transformer layers, $H$:
action-chunk length. The
selector and proxy are training-only; only one BC-policy variant is
deployed per task.}
\centering
\small
\setlength{\tabcolsep}{4pt}
\begin{tabular}{l l c c c c r}
\toprule
\textbf{Module} & \textbf{Architecture} & $d$ & $h$ & $L$ & $H$ & \textbf{Params} \\
\midrule
$\pi_\mathrm{s}$ (selector)            & ResNet-18 + MLP\,$[500,\,500]$ & 256  & --- & --- & --- & 15.80\,M \\
$f_\theta$ (proxy)                     & ResNet-18 + Transformer        & 512  & 4   & 4   & 8   & 24.13\,M \\
$\pi_\mathrm{BC}^{\text{SST}}$         & Regression transformer         & 1024 & 16  & 4   & 1   & 62.13\,M \\
$\pi_\mathrm{BC}^{\text{FM}}$          & Flow-matching cross-attn transformer & 1024 & 16  & 4   & 8   & 120.20\,M \\
\bottomrule
\end{tabular}

\label{tab:arch}
\end{table}

The single-step regression head ($\pi_\mathrm{BC}^{\text{SST}}$) is
the default BC-policy. We substitute a flow-matching transformer
($\pi_\mathrm{BC}^{\text{FM}}$) only on the four \texttt{Intercept*}
dynamics tasks (\texttt{InterceptMedium} and
\texttt{InterceptGrab}\,\{\texttt{Slow},\texttt{Medium},\texttt{Fast}\}),
where the higher-frequency control profile benefits from action
chunking of length $H = 8$. The flow-matching backbone is a
cross-attention transformer in which action tokens
self-attend and then cross-attend to visual and proprioceptive context
tokens.

\subsection{Optimizer and schedule}
\label{app:training:optim}

The proxy and BC policies are trained with
\textsc{AdamW}~\citep{kingma2014adam}; the selector is trained with
\textsc{PPO}~\citep{schulman2017proximal} using Adam (Table~\ref{tab:optim}). Actions
are predicted directly in the simulator's $[-1,\,1]$ range without
output normalization.

\begin{table}[h]
\caption{Optimizer settings for the three training stages.}
\centering
\small
\begin{tabular}{l l l l l}
\toprule
\textbf{Stage} & \textbf{Optimizer} & \textbf{LR} & \textbf{Weight decay} & \textbf{Schedule} \\
\midrule
Selector policy $\pi_\mathrm{s}$ (PPO)             & Adam   & $5\,{\times}\,10^{-5}$  & $0$    & constant \\
Proxy policy $f_\theta$                            & AdamW  & $1\,{\times}\,10^{-4}$  & $0$    & constant \\
BC policy $\pi_\mathrm{BC}^{\text{SST}}$           & AdamW  & $5\,{\times}\,10^{-5}$  & $0.05$ & constant \\
BC policy $\pi_\mathrm{BC}^{\text{FM}}$            & AdamW  & $5\,{\times}\,10^{-5}$  & $0.05$ & constant \\
\bottomrule
\end{tabular}
\label{tab:optim}
\end{table}

\subsection{Stage-1: Proxy policy $f_\theta$}
\label{app:training:proxy}

The proxy policy is trained for $60$ epochs over the demonstration
dataset $\mathcal{D}$ with batch size $128$. We use online masked-sampling (Appendix~\ref{app:proxy-training}) and at every gradient step we
sample a random binary mask $\Psi_k$ over the past observations $h_t$
with cardinality bounded by $k$, forming the masked memory buffer
$\mathcal{B}_t = \{o_i \in h_t \mid \Psi_k(i) = 1\}$
(Section~\ref{sec:keyframe_discovery}). Conditioned on $\mathcal{B}_t$,
the proxy is trained to reconstruct
the expert action chunk at horizon $H = 8$ under L1 loss.

\subsection{Stage-2: Selector policy $\pi_\mathrm{s}$}
\label{app:training:selector}

The selector policy is a PPO actor-critic with the hyperparameters of
Table~\ref{tab:ppo}. The reward at every step is the normalised log objective defined in
Section~\ref{sec:selector_optimization}, evaluated on the post-update
memory buffer $\mathcal{B}_t$. We do not use surprise coefficient for robot manipulation domains. We clip the proxy loss inside the
log-reward at $\lambda = 10^{-5}$ to prevent unbounded rewards near
the proxy's convergence. The selector observes the same RGB and
proprioceptive state input as the proxy policy, and emits a continuous
priority score $p_t \in [0, 1]$ per observation; the priority queue
uses this score to update $\mathcal{B}_t$ from $\mathcal{B}_{t-1}$ (Algorithm~\ref{alg:fifo_queue}).
We train the selector for $5\,{\times}\,10^{6}$ environment steps
across $n_{\mathrm{envs}}=32$ parallel selector environments.

\begin{table}[h]
  \caption{PPO hyperparameters used to train the selector policy
  $\pi_\mathrm{s}$.}
  \centering
  \small
  \setlength{\tabcolsep}{6pt}
  \begin{tabular}{l c | l c | l c}
  \toprule
  \textbf{Hyperparameter} & \textbf{Value} & \textbf{Hyperparameter} & \textbf{Value} &
   \textbf{Hyperparameter} & \textbf{Value} \\
  \midrule
  Total environment steps          & $5\,{\times}\,10^{6}$ & PPO epochs per update &
  $5$       & Discount $\gamma$                & $0.99$ \\
  Parallel envs ($n_\text{envs}$)  & $32$                  & Entropy coefficient   &
  $0.01$    & GAE $\lambda_{\text{GAE}}$       & $0.95$ \\
  Rollout length ($n_\text{steps}$)& $1024$                & Clip range            &
  $0.2$     & Value-function coef.\ $c_\text{vf}$ & $0.5$ \\
  Minibatch size                   & $256$                 & Reward floor $\lambda$&
  $10^{-5}$ & Max grad norm                    & $0.5$  \\
  \bottomrule
  \end{tabular}
  \label{tab:ppo}
\end{table}

\begin{table}[t]
  \centering
  \caption{Success rate (mean$\pm$sem over 3 seeds, 100 episodes per seed) on the MIKASA-Robo benchmark.
  \textbf{Bold} marks every method whose mean is within standard error of the best per
  row.
  }
  \label{tab:mikasa_results}
  \resizebox{\linewidth}{!}{%
  \begin{tabular}{lccccccc}
  \toprule
  Environment & RATE & DT & BC & CQL & DP & ELMUR & KM (Ours) \\
  \midrule
  Shell Game Touch & 0.92 $\pm$ 0.01 & 0.53 $\pm$ 0.07 & 0.28 $\pm$ 0.01 & 0.16 $\pm$ 0.04 &
  0.18 $\pm$ 0.02 & \textbf{0.96 $\pm$ 0.02} & \textbf{0.94 $\pm$ 0.02} \\
  Shell Game Push & 0.78 $\pm$ 0.06 & 0.62 $\pm$ 0.14 & 0.27 $\pm$ 0.01 & 0.25 $\pm$ 0.01 &
  0.22 $\pm$ 0.03 & 0.87 $\pm$ 0.04 & \textbf{0.99 $\pm$ 0.01} \\
  Shell Game Pick & 0.02 $\pm$ 0.01 & 0.00 $\pm$ 0.00 & 0.01 $\pm$ 0.01 & 0.00 $\pm$ 0.00 &
  0.01 $\pm$ 0.00 & 0.66 $\pm$ 0.01 & \textbf{0.78 $\pm$ 0.08} \\
  Intercept Slow & 0.23 $\pm$ 0.02 & 0.40 $\pm$ 0.02 & 0.37 $\pm$ 0.06 & 0.25 $\pm$ 0.01 & 0.33
   $\pm$ 0.05 & \textbf{0.61 $\pm$ 0.04} & 0.48 $\pm$ 0.06 \\
  Intercept Medium & 0.32 $\pm$ 0.02 & 0.56 $\pm$ 0.01 & 0.31 $\pm$ 0.14 & 0.03 $\pm$ 0.01 &
  0.68 $\pm$ 0.02 & 0.64 $\pm$ 0.03 & \textbf{0.78 $\pm$ 0.01} \\
  Intercept Fast & 0.30 $\pm$ 0.04 & 0.36 $\pm$ 0.04 & 0.03 $\pm$ 0.02 & 0.02 $\pm$ 0.02 & 0.21
   $\pm$ 0.05 & 0.54 $\pm$ 0.02 & \textbf{0.59 $\pm$ 0.04} \\
  Intercept Grab Slow & 0.09 $\pm$ 0.03 & 0.00 $\pm$ 0.00 & \textbf{0.28 $\pm$ 0.18} & 0.03
  $\pm$ 0.00 & 0.03 $\pm$ 0.01 & 0.08 $\pm$ 0.03 & \textbf{0.30 $\pm$ 0.04} \\
  Intercept Grab Medium & \textbf{0.09 $\pm$ 0.03} & 0.00 $\pm$ 0.00 & \textbf{0.11 $\pm$ 0.02}
   & \textbf{0.08 $\pm$ 0.04} & 0.03 $\pm$ 0.01 & \textbf{0.10 $\pm$ 0.04} & \textbf{0.12 $\pm$
   0.08} \\
  Intercept Grab Fast & 0.14 $\pm$ 0.03 & 0.11 $\pm$ 0.03 & 0.09 $\pm$ 0.02 & 0.08 $\pm$ 0.03 &
   0.18 $\pm$ 0.02 & 0.20 $\pm$ 0.01 & \textbf{0.37 $\pm$ 0.02} \\
  Rotate Lenient Pos & 0.11 $\pm$ 0.04 & 0.01 $\pm$ 0.01 & 0.15 $\pm$ 0.03 & 0.16 $\pm$ 0.02 &
  0.11 $\pm$ 0.02 & 0.24 $\pm$ 0.02 & \textbf{0.52 $\pm$ 0.05} \\
  Rotate Lenient Pos Neg & 0.29 $\pm$ 0.03 & 0.05 $\pm$ 0.02 & 0.22 $\pm$ 0.01 & 0.12 $\pm$
  0.02 & 0.14 $\pm$ 0.05 & 0.26 $\pm$ 0.00 & \textbf{0.71 $\pm$ 0.06} \\
  Rotate Strict Pos & 0.03 $\pm$ 0.02 & 0.05 $\pm$ 0.04 & 0.01 $\pm$ 0.00 & 0.03 $\pm$ 0.01 &
  0.06 $\pm$ 0.02 & 0.16 $\pm$ 0.00 & \textbf{0.52 $\pm$ 0.02} \\
  Rotate Strict Pos Neg & 0.08 $\pm$ 0.01 & 0.05 $\pm$ 0.03 & 0.04 $\pm$ 0.02 & 0.04 $\pm$ 0.02
   & 0.15 $\pm$ 0.01 & 0.12 $\pm$ 0.02 & \textbf{0.42 $\pm$ 0.07} \\
  Take It Back & 0.42 $\pm$ 0.24 & 0.08 $\pm$ 0.04 & 0.33 $\pm$ 0.10 & 0.04 $\pm$ 0.01 & 0.05
  $\pm$ 0.02 & 0.78 $\pm$ 0.03 & \textbf{0.90 $\pm$ 0.04} \\
  Remember Color3 & 0.65 $\pm$ 0.04 & 0.01 $\pm$ 0.01 & 0.27 $\pm$ 0.03 & 0.29 $\pm$ 0.01 &
  0.32 $\pm$ 0.01 & 0.89 $\pm$ 0.07 & \textbf{0.99 $\pm$ 0.01} \\
  Remember Color5 & 0.13 $\pm$ 0.03 & 0.07 $\pm$ 0.05 & 0.12 $\pm$ 0.01 & 0.15 $\pm$ 0.02 &
  0.10 $\pm$ 0.02 & 0.19 $\pm$ 0.03 & \textbf{0.46 $\pm$ 0.07} \\
  Remember Color9 & 0.09 $\pm$ 0.02 & 0.01 $\pm$ 0.01 & 0.12 $\pm$ 0.02 & 0.15 $\pm$ 0.01 &
  0.17 $\pm$ 0.01 & 0.23 $\pm$ 0.02 & \textbf{0.33 $\pm$ 0.03} \\
  Remember Shape3 & 0.21 $\pm$ 0.04 & 0.05 $\pm$ 0.04 & 0.31 $\pm$ 0.04 & 0.20 $\pm$ 0.10 &
  0.32 $\pm$ 0.05 & \textbf{0.76 $\pm$ 0.01} & 0.63 $\pm$ 0.15 \\
  Remember Shape5 & 0.17 $\pm$ 0.04 & 0.04 $\pm$ 0.04 & 0.18 $\pm$ 0.01 & 0.15 $\pm$ 0.00 &
  0.21 $\pm$ 0.04 & 0.27 $\pm$ 0.02 & \textbf{0.49 $\pm$ 0.08} \\
  Remember Shape9 & 0.05 $\pm$ 0.00 & 0.05 $\pm$ 0.02 & 0.10 $\pm$ 0.02 & 0.14 $\pm$ 0.01 &
  0.11 $\pm$ 0.02 & \textbf{0.17 $\pm$ 0.03} & \textbf{0.17 $\pm$ 0.02} \\
  Remember Shape Color3x2 & 0.14 $\pm$ 0.02 & 0.04 $\pm$ 0.02 & 0.13 $\pm$ 0.02 & 0.11 $\pm$
  0.05 & 0.14 $\pm$ 0.02 & 0.19 $\pm$ 0.02 & \textbf{0.52 $\pm$ 0.07} \\
  Remember Shape Color3x3 & 0.08 $\pm$ 0.03 & 0.06 $\pm$ 0.06 & 0.09 $\pm$ 0.02 & 0.09 $\pm$
  0.02 & 0.16 $\pm$ 0.01 & 0.19 $\pm$ 0.02 & \textbf{0.27 $\pm$ 0.03} \\
  Remember Shape Color5x3 & 0.07 $\pm$ 0.02 & 0.01 $\pm$ 0.01 & \textbf{0.09 $\pm$ 0.01} &
  \textbf{0.09 $\pm$ 0.02} & \textbf{0.11 $\pm$ 0.03} & \textbf{0.13 $\pm$ 0.01} & \textbf{0.14
   $\pm$ 0.06} \\
  Bunch Of Colors3 & 0.00 $\pm$ 0.00 & 0.00 $\pm$ 0.00 & 0.00 $\pm$ 0.00 & 0.00 $\pm$ 0.00 &
  0.00 $\pm$ 0.00 & 0.00 $\pm$ 0.00 & 0.00 $\pm$ 0.00 \\
  Bunch Of Colors5 & 0.00 $\pm$ 0.00 & 0.00 $\pm$ 0.00 & 0.00 $\pm$ 0.00 & 0.00 $\pm$ 0.00 &
  0.00 $\pm$ 0.00 & 0.00 $\pm$ 0.00 & 0.00 $\pm$ 0.00 \\
  Bunch Of Colors7 & 0.00 $\pm$ 0.00 & 0.00 $\pm$ 0.00 & 0.00 $\pm$ 0.00 & 0.00 $\pm$ 0.00 &
  0.00 $\pm$ 0.00 & 0.00 $\pm$ 0.00 & 0.00 $\pm$ 0.00 \\
  Seq Of Colors3 & 0.00 $\pm$ 0.00 & 0.00 $\pm$ 0.00 & 0.00 $\pm$ 0.00 & 0.00 $\pm$ 0.00 & 0.00
   $\pm$ 0.00 & 0.00 $\pm$ 0.00 & 0.00 $\pm$ 0.00 \\
  Seq Of Colors5 & 0.00 $\pm$ 0.00 & 0.00 $\pm$ 0.00 & 0.00 $\pm$ 0.00 & 0.00 $\pm$ 0.00 & 0.00
   $\pm$ 0.00 & 0.00 $\pm$ 0.00 & 0.00 $\pm$ 0.00 \\
  Seq Of Colors7 & 0.00 $\pm$ 0.00 & 0.00 $\pm$ 0.00 & 0.00 $\pm$ 0.00 & 0.00 $\pm$ 0.00 & 0.00
   $\pm$ 0.00 & 0.00 $\pm$ 0.00 & 0.00 $\pm$ 0.00 \\
  Chain Of Colors3 & 0.00 $\pm$ 0.00 & 0.00 $\pm$ 0.00 & 0.00 $\pm$ 0.00 & 0.00 $\pm$ 0.00 &
  0.00 $\pm$ 0.00 & 0.00 $\pm$ 0.00 & 0.00 $\pm$ 0.00 \\
  Chain Of Colors5 & 0.00 $\pm$ 0.00 & 0.00 $\pm$ 0.00 & 0.00 $\pm$ 0.00 & 0.00 $\pm$ 0.00 &
  0.00 $\pm$ 0.00 & 0.00 $\pm$ 0.00 & 0.00 $\pm$ 0.00 \\
  Chain Of Colors7 & 0.00 $\pm$ 0.00 & 0.00 $\pm$ 0.00 & 0.00 $\pm$ 0.00 & 0.00 $\pm$ 0.00 &
  0.00 $\pm$ 0.00 & 0.00 $\pm$ 0.00 & 0.00 $\pm$ 0.00 \\
  \bottomrule
  \end{tabular}}
  \end{table}

\subsection{Stage-3: BC policy $\pi_\mathrm{BC}$}
\label{app:training:policy}

\textbf{Optimization budget.} The regression head
($\pi_\mathrm{BC}^{\text{SST}}$) is trained for $15{,}000$ gradient
steps; the flow-matching variant ($\pi_\mathrm{BC}^{\text{FM}}$) is
trained for $100{,}000$ gradient steps. Batch size is $128$ throughout, and the
final checkpoint is used for evaluation.

\textbf{Flow-matching scheduler.} We adopt the Euler discrete
flow-matching scheduler~\cite{lipman2022flow} with $200$ training
timesteps sampled uniformly at random. At inference we run $5$
Euler steps; we found this sufficient on every \texttt{Intercept*}
task.

\textbf{Closed-loop control.} The BC policy emits an action chunk of
length $H$ at every environment step, but only the first action of the
chunk is executed before re-querying.

\subsection{Camera-image data augmentation}
\label{app:training:augment}

We apply augmentation to all camera
images in the buffer during BC policy training. Each of the following operations
is applied independently with probability $p = 0.5$: brightness
jitter $[0.8,\,1.2]$, contrast jitter $[0.8,\,1.2]$, saturation jitter
$[0.5,\,1.5]$, hue jitter $[-0.05,\,0.05]$, sharpness adjustment
(factor $1.5$), and a random affine transform with up to $10^{\circ}$
rotation, $\pm 10\%$ translation, and $\pm 10\%$ scaling. The
proprioceptive state vector is not augmented. We find this augmentation crucial for training the BC policy to avoid causal confusion \cite{de2019causal} induced by the larger context size.

\subsection{\texttt{MIKASA-Robo} all results}
\label{app:mikasa-results}
Averaged over the 23 non-zero tasks, Keyframe Mnemonics outperforms Decision Transformer by $+40.4\%$, Conservative Q-Learning by $+42.5\%$, Behavior Cloning by $+37.1\%$, \texttt{RATE} by $+30.6\%$, the diffusion-policy baseline by $+36.6\%$, and \texttt{ELMUR} by $+13.9\%$ absolute success rate. See Table \ref{tab:mikasa_results} for complete results.

  \subsection{Hardware and wall-clock}
  \label{app:robot-hardware}
  Training runs were conducted on a mix of \textsc{NVIDIA} A$40$
  ($48$\,GB) and L$40$S ($46$\,GB) GPU instances. Per-stage
  wall-clock, averaged across the $23$ non-zero \texttt{MIKASA-Robo}
  tasks and rounded, is approximately $1.5$\,h for the proxy
  $f_\theta$ ($60$ epochs), $\sim\!10$\,h for the selector
  $\pi_\mathrm{s}$ ($5\,{\times}\,10^{6}$ environment steps), and
  $\sim\!1.5$\,h for the BC policy $\pi_\mathrm{BC}^{\text{SST}}$
  ($15{,}000$ gradient steps). End-to-end training therefore takes
  roughly half a day per SST task ($\sim\!13$\,h). The
  flow-matching variant $\pi_\mathrm{BC}^{\text{FM}}$ trains for
  $100{,}000$ gradient steps and takes proportionally longer at the
  BC stage ($\sim\!10$\,h), bringing the four \texttt{Intercept*}
  flow-matching tasks to roughly a full day each.

\section{Experimental Details}
\label{appendix:experiments}

\begin{table}[h]
\centering
\caption{Horizon-invariance results across synthetic and grid benchmarks. Cells
report mean$~\pm~$std of the relevant metric: final-step MSE for Add
($\downarrow$ better), Copy SR for Scattered-Copy and Episode SR for
T-Maze, MiniGrid-Memory and LTMB-Hallway ($\uparrow$ better). Each
benchmark lists the horizons (or environment size for
MiniGrid-Memory) evaluated. ``---'' marks cells for baselines in extreme-horizon experiments as we run it only for Keyframe Mnemonics, or for the LSTM(BPTT) variant on grid tasks, which were trained only with LSTM(TBPTT).}
\label{tab:horizon-scaling}
\providecommand{\E}[1]{{\scriptscriptstyle\,\cdot\,10^{#1}}}
\resizebox{\textwidth}{!}{%
\setlength{\tabcolsep}{3pt}
\begin{tabular}{l rrrrrrrr}
\toprule
& LSTM(B) & LSTM(T) & DLSTM & DT & DMamba & TrXL & ELMUR & \textbf{KM (Ours)} \\
\midrule
\multicolumn{9}{l}{\textbf{Add} (final-step MSE $\downarrow$)} \\
$h=10$       & $\mathbf{5.0\E{-6} \!\pm\! 2.3\E{-6}}$ & $3.4\E{-3} \!\pm\! 3.5\E{-3}$ & $0.258 \!\pm\! 0.216$ & $\mathbf{6.5\E{-4} \!\pm\! 6.2\E{-4}}$ & $\mathbf{2.5\E{-4} \!\pm\! 1.8\E{-4}}$ & $\mathbf{3.9\E{-4} \!\pm\! 2.4\E{-4}}$ & $0.143 \!\pm\! 0.114$ & $\mathbf{3.1\E{-5} \!\pm\! 3.4\E{-5}}$ \\
$h=20$       & $\mathbf{7.7\E{-5} \!\pm\! 6.9\E{-5}}$ & $6.8\E{-3} \!\pm\! 5.6\E{-3}$ & $0.330 \!\pm\! 0.366$ & $\mathbf{9.2\E{-4} \!\pm\! 7.7\E{-4}}$ & $\mathbf{2.1\E{-4} \!\pm\! 1.9\E{-4}}$ & $\mathbf{1.8\E{-4} \!\pm\! 1.4\E{-4}}$ & $0.131 \!\pm\! 0.090$ & $\mathbf{2.8\E{-4} \!\pm\! 5.7\E{-4}}$ \\
$h=30$       & $\mathbf{3.1\E{-4} \!\pm\! 1.9\E{-4}}$ & $1.1\E{-2} \!\pm\! 8.3\E{-3}$ & $0.267 \!\pm\! 0.348$ & $\mathbf{7.0\E{-4} \!\pm\! 7.8\E{-4}}$ & $\mathbf{2.2\E{-4} \!\pm\! 1.5\E{-4}}$ & $\mathbf{1.4\E{-4} \!\pm\! 6.1\E{-5}}$ & $0.141 \!\pm\! 0.052$ & $\mathbf{7.7\E{-5} \!\pm\! 1.1\E{-4}}$ \\
$h=10^{3}$   & $0.246 \!\pm\! 0.097$ & $0.127 \!\pm\! 0.027$ & $0.531 \!\pm\! 0.455$ & $0.408 \!\pm\! 0.439$ & $0.387 \!\pm\! 0.247$ & $0.296 \!\pm\! 0.216$ & $0.797 \!\pm\! 0.938$ & $\mathbf{3.8\E{-5} \!\pm\! 4.8\E{-5}}$ \\
$h=10^{4}$   & $0.494 \!\pm\! 0.320$ & $0.351 \!\pm\! 0.142$ & $0.218 \!\pm\! 0.195$ & $0.447 \!\pm\! 0.398$ & $0.594 \!\pm\! 0.359$ & $0.634 \!\pm\! 0.387$ & $0.759 \!\pm\! 0.465$ & $\mathbf{3.7\E{-5} \!\pm\! 4.0\E{-5}}$ \\
$h=10^{5}$   & --- & --- & --- & --- & --- & --- & --- & $\mathbf{3.8\E{-5} \!\pm\! 3.8\E{-5}}$ \\
$h=10^{6}$   & --- & --- & --- & --- & --- & --- & --- & $\mathbf{1.6\E{-5} \!\pm\! 3.5\E{-5}}$ \\
\midrule
\multicolumn{9}{l}{\textbf{Scattered-Copy} (Copy SR $\uparrow$)} \\
$h=10$       & $0.720 \!\pm\! 0.110$ & $0.080 \!\pm\! 0.110$ & $0.000 \!\pm\! 0.000$ & $0.000 \!\pm\! 0.000$ & $0.000 \!\pm\! 0.000$ & $0.000 \!\pm\! 0.000$ & $0.000 \!\pm\! 0.000$ & $\mathbf{1.000 \!\pm\! 0.000}$ \\
$h=20$       & $0.360 \!\pm\! 0.167$ & $0.080 \!\pm\! 0.110$ & $0.000 \!\pm\! 0.000$ & $0.480 \!\pm\! 0.390$ & $0.000 \!\pm\! 0.000$ & $0.040 \!\pm\! 0.089$ & $0.000 \!\pm\! 0.000$ & $\mathbf{1.000 \!\pm\! 0.000}$ \\
$h=30$       & $0.360 \!\pm\! 0.167$ & $0.080 \!\pm\! 0.110$ & $0.000 \!\pm\! 0.000$ & $0.040 \!\pm\! 0.089$ & $0.000 \!\pm\! 0.000$ & $0.000 \!\pm\! 0.000$ & $0.000 \!\pm\! 0.000$ & $\mathbf{1.000 \!\pm\! 0.000}$ \\
$h=10^{3}$   & $0.120 \!\pm\! 0.179$ & $0.160 \!\pm\! 0.167$ & $0.000 \!\pm\! 0.000$ & $0.000 \!\pm\! 0.000$ & $0.000 \!\pm\! 0.000$ & $0.000 \!\pm\! 0.000$ & $0.000 \!\pm\! 0.000$ & $\mathbf{1.000 \!\pm\! 0.000}$ \\
$h=10^{4}$   & $0.000 \!\pm\! 0.000$ & $0.040 \!\pm\! 0.089$ & $0.000 \!\pm\! 0.000$ & $0.000 \!\pm\! 0.000$ & $0.000 \!\pm\! 0.000$ & $0.000 \!\pm\! 0.000$ & $0.000 \!\pm\! 0.000$ & $\mathbf{1.000 \!\pm\! 0.000}$ \\
$h=10^{5}$   & --- & --- & --- & --- & --- & --- & --- & $\mathbf{1.000 \!\pm\! 0.000}$ \\
$h=10^{6}$   & --- & --- & --- & --- & --- & --- & --- & $\mathbf{1.000 \!\pm\! 0.000}$ \\
\midrule
\multicolumn{9}{l}{\textbf{T-Maze} (Episode SR $\uparrow$)} \\
$h=10$       & $\mathbf{1.000 \!\pm\! 0.000}$ & $0.360 \!\pm\! 0.219$ & $0.680 \!\pm\! 0.179$ & $\mathbf{1.000 \!\pm\! 0.000}$ & $0.560 \!\pm\! 0.261$ & $\mathbf{1.000 \!\pm\! 0.000}$ & $0.400 \!\pm\! 0.200$ & $\mathbf{1.000 \!\pm\! 0.000}$ \\
$h=20$       & $\mathbf{1.000 \!\pm\! 0.000}$ & $0.320 \!\pm\! 0.179$ & $0.560 \!\pm\! 0.329$ & $\mathbf{1.000 \!\pm\! 0.000}$ & $0.520 \!\pm\! 0.335$ & $\mathbf{1.000 \!\pm\! 0.000}$ & $0.280 \!\pm\! 0.228$ & $\mathbf{1.000 \!\pm\! 0.000}$ \\
$h=30$       & $\mathbf{1.000 \!\pm\! 0.000}$ & $0.280 \!\pm\! 0.335$ & $0.640 \!\pm\! 0.167$ & $\mathbf{1.000 \!\pm\! 0.000}$ & $0.520 \!\pm\! 0.228$ & $\mathbf{1.000 \!\pm\! 0.000}$ & $0.520 \!\pm\! 0.228$ & $\mathbf{1.000 \!\pm\! 0.000}$ \\
$h=10^{3}$   & $0.800 \!\pm\! 0.283$ & $0.320 \!\pm\! 0.228$ & $0.320 \!\pm\! 0.228$ & $0.320 \!\pm\! 0.228$ & $0.360 \!\pm\! 0.261$ & $0.520 \!\pm\! 0.303$ & $0.000 \!\pm\! 0.000$ & $\mathbf{1.000 \!\pm\! 0.000}$ \\
$h=10^{4}$   & $0.840 \!\pm\! 0.261$ & $0.320 \!\pm\! 0.110$ & $0.320 \!\pm\! 0.110$ & $0.320 \!\pm\! 0.110$ & $0.360 \!\pm\! 0.167$ & $0.440 \!\pm\! 0.219$ & $0.000 \!\pm\! 0.000$ & $\mathbf{1.000 \!\pm\! 0.000}$ \\
$h=10^{5}$   & --- & --- & --- & --- & --- & --- & --- & $\mathbf{1.000 \!\pm\! 0.000}$ \\
$h=10^{6}$   & --- & --- & --- & --- & --- & --- & --- & $\mathbf{1.000 \!\pm\! 0.000}$ \\
\midrule
\multicolumn{9}{l}{\textbf{MiniGrid-Memory} (Episode SR $\uparrow$)} \\
$h=17$   & --- & $0.767 \!\pm\! 0.404$ & $0.800 \!\pm\! 0.000$ & $0.333 \!\pm\! 0.153$ & $0.500 \!\pm\! 0.000$ & $0.600 \!\pm\! 0.000$ & $0.467 \!\pm\! 0.058$ & $\mathbf{1.000 \!\pm\! 0.000}$ \\
$h=51$   & --- & $0.667 \!\pm\! 0.321$ & $0.533 \!\pm\! 0.231$ & $0.533 \!\pm\! 0.058$ & $0.300 \!\pm\! 0.000$ & $0.600 \!\pm\! 0.000$ & $0.367 \!\pm\! 0.115$ & $\mathbf{1.000 \!\pm\! 0.000}$ \\
$h=101$  & --- & $0.600 \!\pm\! 0.265$ & $0.500 \!\pm\! 0.173$ & $0.333 \!\pm\! 0.153$ & $0.100 \!\pm\! 0.000$ & $0.500 \!\pm\! 0.100$ & $0.333 \!\pm\! 0.153$ & $\mathbf{1.000 \!\pm\! 0.000}$ \\
$h=501$  & --- & $0.500 \!\pm\! 0.173$ & $0.500 \!\pm\! 0.173$ & $0.433 \!\pm\! 0.231$ & $0.000 \!\pm\! 0.000$ & $0.300 \!\pm\! 0.100$ & $0.333 \!\pm\! 0.153$ & $\mathbf{1.000 \!\pm\! 0.000}$ \\
$h=1001$ & --- & $0.467 \!\pm\! 0.208$ & $0.500 \!\pm\! 0.173$ & $0.433 \!\pm\! 0.153$ & $0.000 \!\pm\! 0.000$ & $0.367 \!\pm\! 0.115$ & $0.300 \!\pm\! 0.173$ & $\mathbf{1.000 \!\pm\! 0.000}$ \\
\midrule
\multicolumn{9}{l}{\textbf{LTMB-Hallway} (Episode SR $\uparrow$)} \\
$h=8$        & --- & $0.033 \!\pm\! 0.058$ & $0.033 \!\pm\! 0.058$ & $0.133 \!\pm\! 0.058$ & $0.000 \!\pm\! 0.000$ & $0.100 \!\pm\! 0.000$ & $0.100 \!\pm\! 0.000$ & $\mathbf{1.000 \!\pm\! 0.000}$ \\
$h=50$       & --- & $0.000 \!\pm\! 0.000$ & $0.000 \!\pm\! 0.000$ & $0.000 \!\pm\! 0.000$ & $0.000 \!\pm\! 0.000$ & $0.000 \!\pm\! 0.000$ & $0.000 \!\pm\! 0.000$ & $\mathbf{1.000 \!\pm\! 0.000}$ \\
$h=100$      & --- & $0.000 \!\pm\! 0.000$ & $0.000 \!\pm\! 0.000$ & $0.000 \!\pm\! 0.000$ & $0.000 \!\pm\! 0.000$ & $0.000 \!\pm\! 0.000$ & $0.000 \!\pm\! 0.000$ & $\mathbf{0.933 \!\pm\! 0.058}$ \\
$h=500$      & --- & $0.000 \!\pm\! 0.000$ & $0.000 \!\pm\! 0.000$ & $0.000 \!\pm\! 0.000$ & $0.000 \!\pm\! 0.000$ & $0.000 \!\pm\! 0.000$ & $0.000 \!\pm\! 0.000$ & $\mathbf{0.967 \!\pm\! 0.058}$ \\
$h=1000$     & --- & $0.000 \!\pm\! 0.000$ & $0.000 \!\pm\! 0.000$ & $0.000 \!\pm\! 0.000$ & $0.000 \!\pm\! 0.000$ & $0.000 \!\pm\! 0.000$ & $0.000 \!\pm\! 0.000$ & $\mathbf{0.933 \!\pm\! 0.058}$ \\
\bottomrule
\end{tabular}%
}
\vspace{2pt}
\\{\scriptsize Random baseline omitted; its test SR remains
low across every (benchmark, horizon) cell, consistent with performance shown in Table~\ref{tab:results}.}
\end{table}

\subsection{Horizon-invariance experiments on synthetic and grid domains}
\label{app:horizon-scaling}

Table~\ref{tab:horizon-scaling} reports the consolidated horizon-invariance results underlying Figure~\ref{fig:horizon_all5}. We evaluated the trained
checkpoints from Appendix~\ref{appendix:benchmark_details}, by instantiating the test
environment with a target horizon $h$ taken from a pre-defined set per benchmark: Add, Scattered-Copy, T-Maze episode length $h\in\{10,20,30,1000,10000\}$; Keyframe Mnemonics is additionally evaluated at $h\in\{10^{5},10^{6}\}$ to test horizon-invariance behavior.
MiniGrid-Memory environment max size $h\in\{17,51,101,501,1001\}$.
LTMB-Hallway corridor length $h\in\{8,50,100,500,1000\}$.
Per-cell metrics are mean$~\pm~$std across $5$ random seeds for the
synthetic tasks and $3$ random seeds for the grid tasks. For each horizon, every method is evaluated on $5$
episodes for the synthetic tasks (Add, Scattered-Copy, T-Maze) and
$10$ episodes for the grid tasks (MiniGrid-Memory, LTMB-Hallway) per
seed.

Within the training horizon range, several baselines achieve strong performance; for example, LSTM(BPTT), DT, and TrXL reach $1.0$ SR on T-Maze up to $h=30$. However, baseline methods collapse once the test horizon exceeds the training horizon by even one order of magnitude. In contrast, Keyframe Mnemonics consistently exhibits horizon-invariant performance. In particular, on synthetic tasks, it generalizes up to five orders of magnitude beyond the training horizon, confirming the horizon-invariance guarantee of Theorem~2 in regimes where the underlying assumptions hold strongly.

\subsection{Horizon-invariance experiment on \texttt{RememberColor3-$\Delta T$}}
\label{app:training:hover}

\begin{figure}[t]
    \centering
    \includegraphics[width=1\linewidth]{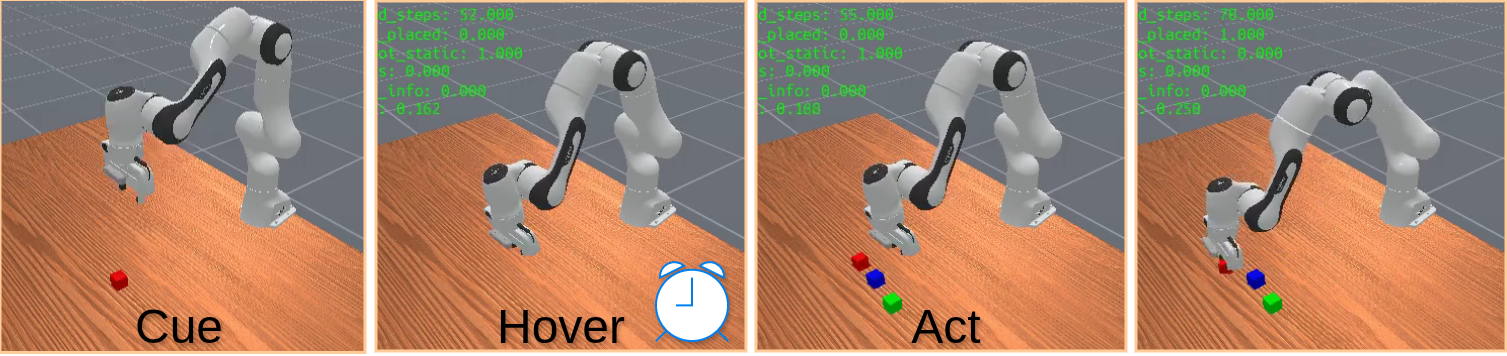}
    \caption{Hover time is varied across evaluations resulting in catastrophic failure for the baselines, while KM maintains it's performance even at $250$ timesteps of hover. Training data had 15 timesteps of hover.}
    \label{fig:hover}
\end{figure}

The horizon-invariance study (Figure~\ref{fig:horizon}) trains every
method at a fixed cue-to-action delay $\Delta T = 15$ and evaluates
the trained policies, without retraining, at delays
$\Delta T \in \{15, 35, 55, 75, 95, 115, 135, 150, 200, 250\}$. The
training set is built from the released \texttt{RememberColor3-v0}
demonstrations, whose default cue-to-action delay is
$\Delta T_{\text{orig}} = 5$.

\textbf{Hover-augmented data.}
We re-replay each successful expert demonstration in a live
\texttt{ManiSkill} simulator with the cue-to-action delay extended to
$\Delta T = 15$ and a $10$-step \emph{extra hover} phase spliced into the
original action tape immediately after the cue disappears (Figure~\ref{fig:hover}). During
the hover phase the gripper is held and small net-zero noise is added to the arm for the entire delay time.

\textbf{Training on the hover dataset.}
We run the standard three-stage KM pipeline of
Appendix~\ref{app:training:pipeline} on the hover trajectories. Every
hyperparameter matches the corresponding entries for
\texttt{RememberColor3} in
Tables~\ref{tab:arch} ($k = 3$, proxy chunk
$H = 8$, SST policy with $H = 1$, $15{,}000$ BC gradient steps).

We train \texttt{PTP} using their official implementation on our dataset with prediction horizon of $16$,
observation length of $8$ and a sub-sampling rate of $8$
to roughly cover the entire episode of length $70$. This indicates why we see steeper
deterioration in the performance of \texttt{PTP} on and after $75$ steps of time delay. We train \texttt{ELMUR} on the same trajectories using its published
\texttt{RememberColor3} recipe, with the context length extended to cover
the full $70$-step hover episode. All other hyperparameters
(architecture, optimizer, schedule, batch size, number of epochs) are
left at the authors' defaults.

\textbf{Horizon-invariance evaluation.}
The trained policy is evaluated at every target delay $\Delta T$ by re-instantiating the simulator with the requested
$\Delta T$; no retraining or fine-tuning is performed at any
$\Delta T \neq 15$. Following the \texttt{MIKASA-Robo} convention, an episode is
counted as a success only if the binary task-completion reward is
reached before the time budget elapses. We report mean $\pm$ standard error across
three independent environment seeds, $100$ episodes per seed.
Results are plotted in Figure~\ref{fig:horizon} and details can be found in Table~\ref{tab:rememberc3-dt}.

\setlength{\columnsep}{15pt}
\begin{wraptable}{r}{0.5\textwidth}
\vspace{-10pt}
\centering
\small
\caption{\texttt{RememberColor3}-$\Delta T$ success rate under
cue-to-action delays. Training delay is $\Delta T=15$; all larger
values are out-of-distribution. Mean$~\pm~$sem over $3$ seeds, $100$ episodes per seed.}
\label{tab:rememberc3-dt}

\resizebox{\linewidth}{!}{%
\begin{tabular}{r c c c}
\toprule
$\Delta T$ & PTP~\cite{torne2025learning} & ELMUR~\cite{cherepanov2025elmur} & \textbf{KM (Ours)} \\
\midrule
$15$  & $0.66 \pm 0.03$ & $0.61 \pm 0.07$ & $\mathbf{0.72 \pm 0.07}$ \\
$35$  & $0.72 \pm 0.04$ & $0.00 \pm 0.00$ & $\mathbf{0.69 \pm 0.11}$ \\
$55$  & $\mathbf{0.73 \pm 0.05}$ & $0.00 \pm 0.00$ & $0.69 \pm 0.11$ \\
$75$  & $0.55 \pm 0.07$ & $0.00 \pm 0.00$ & $\mathbf{0.69 \pm 0.08}$ \\
$95$  & $0.43 \pm 0.06$ & $0.00 \pm 0.00$ & $\mathbf{0.65 \pm 0.08}$ \\
$115$ & $0.31 \pm 0.05$ & $0.00 \pm 0.00$ & $\mathbf{0.66 \pm 0.09}$ \\
$135$ & $0.23 \pm 0.03$ & $0.00 \pm 0.00$ & $\mathbf{0.66 \pm 0.09}$ \\
$150$ & $0.17 \pm 0.03$ & $0.00 \pm 0.00$ & $\mathbf{0.65 \pm 0.09}$ \\
$200$ & $0.07 \pm 0.02$ & $0.00 \pm 0.00$ & $\mathbf{0.64 \pm 0.08}$ \\
$250$ & $0.07 \pm 0.03$ & $0.00 \pm 0.00$ & $\mathbf{0.65 \pm 0.09}$ \\
\bottomrule
\end{tabular}%
}
\end{wraptable}

Results in Table~\ref{tab:rememberc3-dt} show that KM preserves horizon-invariant performance in memory-intensive robot manipulation with high-dimensional visual data. ELMUR matches the other methods at the training delay ($\Delta T=15$), but its external memory is overwritten as the delay increases, causing success to collapse to zero for every $\Delta T \ge 35$. PTP retains its training-time performance up to roughly $\Delta T=55$ and then degrades monotonically, dropping below $10\%$ by $\Delta T=200$. In contrast, Keyframe Mnemonics remains nearly flat across the longer horizons, retaining $\approx 0.65$ success rate at $\Delta T=250$, i.e., $\sim17\!\times$ the training delay, without any architectural change or fine-tuning.

\subsection{Reproducibility}
\label{app:training:repro}
Released code, per-task configurations, and trained
checkpoints, together with launch scripts, are available at \url{https://keyframe-mnemonics.github.io}.

\section{Ablation Study}
\label{app:ablations}
We report ablation results over 5 seeds for the synthetic domains and 3 seeds for the grid and robot domains unless noted, with 100 evaluation episodes per seed. All domains are evaluated at out-of-distribution horizons.

\subsection{Learned selector against a random selector}
\label{app:abl-random}

\begin{wraptable}{r}{0.5\textwidth}
\vspace{-10pt}
\centering
\small
\caption{Learned selector against the random selector $\pi_{\mathrm{rnd}}$ on MIKASA-Robo RememberColor3-$\Delta T$. Reported mean$\pm$sem success rate (\%) $\uparrow$.}

\resizebox{\linewidth}{!}{%
\begin{tabular}{lcc}
\toprule
$\Delta T$ & Learned selector & Random selector ($\pi_{\mathrm{rnd}}$) \\
\midrule
15  & \textbf{72.3 $\pm$ 5.8} & 31.3 $\pm$ 3.1 \\
35  & \textbf{69.0 $\pm$ 9.2} & 28.7 $\pm$ 2.9 \\
55  & \textbf{69.0 $\pm$ 8.8} & 28.3 $\pm$ 2.1 \\
75  & \textbf{68.7 $\pm$ 6.6} & 35.3 $\pm$ 1.2 \\
95  & \textbf{65.0 $\pm$ 6.4} & 28.7 $\pm$ 4.0 \\
115 & \textbf{66.3 $\pm$ 7.6} & 28.3 $\pm$ 4.1 \\
135 & \textbf{65.7 $\pm$ 7.4} & 33.0 $\pm$ 1.4 \\
150 & \textbf{65.3 $\pm$ 7.6} & 29.0 $\pm$ 0.8 \\
200 & \textbf{64.3 $\pm$ 6.9} & 31.7 $\pm$ 1.7 \\
250 & \textbf{65.3 $\pm$ 7.6} & 26.7 $\pm$ 1.2 \\
\bottomrule
\end{tabular}%
}
\vspace{-20pt}
\end{wraptable}

We replace the learned selector with a random selector $\pi_{\mathrm{rnd}}$ that assigns priorities $p \sim \mathcal{U}(0,1)$ and train the BC policy using Stage 3 (Algorithm~\ref{alg:e2e_training}). The resulting policy performs worse at every delay interval, demonstrating that the learned selector is critical for the memory task.

\subsection{Selector discount factor}
\label{app:abl-gamma}

We vary the discount factor $\gamma$ of the PPO selector to optimize over a frozen proxy policy. At $\gamma = 0$ the selector greedily optimizes the per-step proxy loss. For T-Maze we report the active (default) and passive variations~\cite{ni2023transformers}. Greedy optimization fails at $\gamma = 0$ on every domain except active T-Maze, and performance recovers as $\gamma$ increases. In active T-Maze, the action changes right after the keyframe is observed, so the immediate change in proxy loss is sufficient. In the other domains, the gap between the keyframe and the decision-critical action spans multiple steps, and a higher $\gamma$ is necessary. Keyframe identification from the proxy loss is therefore a delayed credit-assignment problem, which explains our choice of using RL for selector training (Section~\ref{sec:selector_optimization}).

\begin{table}[h]
\centering
\small
\caption{Effect of the PPO discount factor $\gamma$. Reported mean$\pm$sd. }
\label{tab:abl-gamma}
\resizebox{\linewidth}{!}{%
\begin{tabular}{llcccc}
\toprule
Domain & Metric & $\gamma = 0$ & $\gamma = 0.5$ & $\gamma = 0.9$ & $\gamma = 0.99$ (default) \\
\midrule
T-Maze (active)  & Episode SR $\uparrow$  & \textbf{100.0 $\pm$ 0.0} & \textbf{100.0 $\pm$ 0.0} & \textbf{100.0 $\pm$ 0.0} & \textbf{100.0 $\pm$ 0.0} \\
T-Maze (passive) & Episode SR $\uparrow$  & 48.8 $\pm$ 6.6 & \textbf{100.0 $\pm$ 0.0} & \textbf{100.0 $\pm$ 0.0} & \textbf{100.0 $\pm$ 0.0} \\
Add              & Final MSE $\downarrow$ & 1.84e-01 $\pm$ 1.77e-03 & 1.25e-02 $\pm$ 6.86e-03 & \textbf{1.28e-05 $\pm$ 5.67e-06} & \textbf{1.28e-05 $\pm$ 5.70e-06} \\
Scattered-Copy   & Copy SR $\uparrow$     & 0.0 $\pm$ 0.0 & \textbf{100.0 $\pm$ 0.0} & \textbf{100.0 $\pm$ 0.0} & \textbf{100.0 $\pm$ 0.0} \\
\bottomrule
\end{tabular}}
\end{table}

\subsection{Surprise coefficient}
\label{app:abl-surprise}

We change the temperature $\kappa$ of the surprise coefficient in Equation~\ref{eq:reward}. Off in Table~\ref{tab:abl-surprise} corresponds to an unscaled reward, which is the $\kappa \to \infty$ limit where surprise coefficient equals 1. For every task we first train with the unscaled reward and add surprise scaling only if it improves performance. $\kappa = 0.2$ has the widest usable range and is our default for the tasks that need scaling (Add, Scattered-Copy and MiniGrid-Memory).

\begin{table}[h]
\centering
\small
\caption{Effect of the surprise coefficient across temperatures. Bold marks the best setting per domain. Reported mean$\pm$sd.}
\label{tab:abl-surprise}
\resizebox{\linewidth}{!}{%
\begin{tabular}{llcccc}
\toprule
Domain & Metric & $\kappa = 0.02$ & $\kappa = 0.2$ & $\kappa = 2$ & off \\
\midrule
T-Maze          & Episode SR $\uparrow$   & 0.0 $\pm$ 0.0 & \textbf{100.0 $\pm$ 0.0} & \textbf{100.0 $\pm$ 0.0} & \textbf{100.0 $\pm$ 0.0} \\
Add             & Final MSE $\downarrow$  & \textbf{1.28e-05 $\pm$ 5.7e-06} & \textbf{1.28e-05 $\pm$ 5.7e-06} & 1.84e-01 $\pm$ 1.8e-03 & 1.84e-01 $\pm$ 1.8e-03 \\
Scattered-Copy  & Copy SR $\uparrow$      & \textbf{100.0 $\pm$ 0.0} & \textbf{100.0 $\pm$ 0.0} & \textbf{100.0 $\pm$ 0.0} & 0.0 $\pm$ 0.0 \\
MiniGrid-Memory & Success rate $\uparrow$ & \textbf{100.0 $\pm$ 0.0} & \textbf{100.0 $\pm$ 0.0} & 3.0 $\pm$ 0.0 & 3.7 $\pm$ 0.6 \\
LTMB-Hallway    & Success rate $\uparrow$ & 0.0 $\pm$ 0.0 & 0.0 $\pm$ 0.0 & 0.0 $\pm$ 0.0 & \textbf{99.7 $\pm$ 0.6} \\
RememberColor3  & Success rate $\uparrow$ & 25.0 $\pm$ 3.3 & 29.3 $\pm$ 6.2 & 34.7 $\pm$ 6.9 & \textbf{99.0 $\pm$ 0.0} \\
\bottomrule
\end{tabular}}
\end{table}

\subsection{Current observation in proxy training}
\label{app:abl-curobs}

Including $o_t$ during proxy training collapses MiniGrid-Memory and RememberColor3 (Table~\ref{tab:abl-curobs}), as the proxy overfits to the current observation. This is consistent with Proposition~1, which requires that no low-complexity non-invariant shortcut exists in the proxy training data. The current observation is helpful for learning the final policy, but not for learning a reward function for keyframe identification.

\begin{table}[h]
\centering
\small
\caption{Effect of including the current observation $o_t$ in the proxy input. Reported mean$\pm$sd.}
\label{tab:abl-curobs}
\begin{tabular}{llcc}
\toprule
Domain & Metric & Proxy without $o_t$ (default) & Proxy with $o_t$ \\
\midrule
T-Maze          & Episode SR $\uparrow$   & 100.0 $\pm$ 0.0 & 100.0 $\pm$ 0.0 \\
Add             & Final MSE $\downarrow$  & 1.28e-05 $\pm$ 5.7e-06 & 2.20e-05 $\pm$ 1.3e-05 \\
Scattered-Copy  & Copy SR $\uparrow$      & 100.0 $\pm$ 0.0 & 100.0 $\pm$ 0.0 \\
MiniGrid-Memory & Success rate $\uparrow$ & \textbf{100.0 $\pm$ 0.0} & 33.3 $\pm$ 26.3 \\
LTMB-Hallway    & Success rate $\uparrow$ & \textbf{99.7 $\pm$ 0.6} & 93.0 $\pm$ 9.5 \\
RememberColor3  & Success rate $\uparrow$ & \textbf{99.0 $\pm$ 0.0} & 18.7 $\pm$ 5.9 \\
\bottomrule
\end{tabular}
\end{table}

\subsection{Queue tie-break rule}
\label{app:abl-queue}

The memory buffer uses a FIFO queue (Algorithm~2) during selector training and a LIFO queue (Algorithm~3) at inference. The two differ in which slot is evicted when several observations are tied at the lowest priority. We vary this rule independently at training and inference. On the synthetic and grid domains, all four combinations are identical (Table~\ref{tab:abl-queue}). The rule matters on RememberColor3-$\Delta T$, where the default FIFO $\rightarrow$ LIFO beats its reverse at 9 of 10 retention intervals (Table~\ref{tab:abl-queue-dt}).

\begin{table}[h]
\centering
\small
\caption{Queue tie-break rule, train $\rightarrow$ inference. Reported mean$\pm$sd.}
\label{tab:abl-queue}
\resizebox{\linewidth}{!}{%
\begin{tabular}{llcccc}
\toprule
Domain & Metric & FIFO $\rightarrow$ FIFO & FIFO $\rightarrow$ LIFO (default) & LIFO $\rightarrow$ FIFO & LIFO $\rightarrow$ LIFO \\
\midrule
T-Maze          & Episode SR $\uparrow$   & 100.0 $\pm$ 0.0 & 100.0 $\pm$ 0.0 & 100.0 $\pm$ 0.0 & 100.0 $\pm$ 0.0 \\
Add             & Final MSE $\downarrow$  & 1.28e-05 $\pm$ 5.7e-06 & 1.28e-05 $\pm$ 5.7e-06 & 1.28e-05 $\pm$ 5.7e-06 & 1.28e-05 $\pm$ 5.7e-06 \\
Scattered-Copy  & Copy SR $\uparrow$      & 100.0 $\pm$ 0.0 & 100.0 $\pm$ 0.0 & 100.0 $\pm$ 0.0 & 100.0 $\pm$ 0.0 \\
MiniGrid-Memory & Success rate $\uparrow$ & 100.0 $\pm$ 0.0 & 100.0 $\pm$ 0.0 & 100.0 $\pm$ 0.0 & 100.0 $\pm$ 0.0 \\
LTMB-Hallway    & Success rate $\uparrow$ & 99.6 $\pm$ 0.5 & 99.6 $\pm$ 0.5 & 99.6 $\pm$ 0.5 & 99.6 $\pm$ 0.5 \\
RememberColor3  & Success rate $\uparrow$ & 93.7 $\pm$ 4.5 & \textbf{99.0 $\pm$ 0.0} & 98.7 $\pm$ 1.3 & 96.0 $\pm$ 2.2 \\
\bottomrule
\end{tabular}}
\end{table}

\begin{table}[h!]
\centering
\small
\caption{Queue tie-break rule on RememberColor3-$\Delta T$. Reported mean$\pm$sem success rate $\uparrow$.}
\label{tab:abl-queue-dt}
\begin{tabular}{lcc}
\toprule
$\Delta T$ & FIFO $\rightarrow$ LIFO (default) & LIFO $\rightarrow$ FIFO \\
\midrule
15  & 72.3 $\pm$ 5.8 & \textbf{75.0 $\pm$ 1.6} \\
35  & \textbf{69.0 $\pm$ 9.2} & 66.0 $\pm$ 3.6 \\
55  & \textbf{69.0 $\pm$ 8.8} & 66.0 $\pm$ 2.4 \\
75  & \textbf{68.7 $\pm$ 6.6} & 59.3 $\pm$ 3.7 \\
95  & \textbf{65.0 $\pm$ 6.4} & 59.3 $\pm$ 1.7 \\
115 & \textbf{66.3 $\pm$ 7.6} & 58.3 $\pm$ 2.5 \\
135 & \textbf{65.7 $\pm$ 7.4} & 60.3 $\pm$ 1.7 \\
150 & \textbf{65.3 $\pm$ 7.6} & 59.3 $\pm$ 2.5 \\
200 & \textbf{64.3 $\pm$ 6.9} & 58.0 $\pm$ 1.6 \\
250 & \textbf{65.3 $\pm$ 7.6} & 57.7 $\pm$ 3.8 \\
\bottomrule
\end{tabular}
\end{table}

\subsection{Memory buffer size}
\label{app:abl-buffer}

We scale the memory buffer size $k$ to $0.5\times$ and $2\times$ the default value and retrain all three stages. Halving $k$ degrades performance across all domains, while doubling it helps on some domains and hurts on others. A larger $k$ can make the KM policy harder to train, and our default values serve as a minimum-capacity floor, which were chosen per domain independently.

\begin{table}[h!]
\centering
\small
\caption{Effect of scaling the memory buffer size $k$. Reported mean$\pm$sd.}
\label{tab:abl-buffer}
\begin{tabular}{llccc}
\toprule
Domain & Metric & $0.5\times$ & $1\times$ (default) & $2\times$ \\
\midrule
T-Maze          & Episode SR $\uparrow$   & 40.0 $\pm$ 23.2 & \textbf{100.0 $\pm$ 0.0} & \textbf{100.0 $\pm$ 0.0} \\
Add             & Final MSE $\downarrow$  & 9.55e-02 $\pm$ 2.38e-02 & \textbf{1.28e-05 $\pm$ 5.7e-06} & 2.33e-05 $\pm$ 2.03e-05 \\
Scattered-Copy  & Copy SR $\uparrow$      & 0.0 $\pm$ 0.0 & \textbf{100.0 $\pm$ 0.0} & 80.0 $\pm$ 44.7 \\
MiniGrid-Memory & Success rate $\uparrow$ & 35.3 $\pm$ 26.4 & \textbf{100.0 $\pm$ 0.0} & 20.7 $\pm$ 27.2 \\
LTMB-Hallway    & Success rate $\uparrow$ & 0.0 $\pm$ 0.0 & 81.7 $\pm$ 15.0 & \textbf{99.3 $\pm$ 1.2} \\
RememberColor3  & Success rate $\uparrow$ & 94.7 $\pm$ 2.5 & 99.0 $\pm$ 0.0 & \textbf{100.0 $\pm$ 0.0} \\
\bottomrule
\end{tabular}
\end{table}

\section{Additional Analysis}
\label{app:additional_analysis}

\subsection{Interpretability of discovered keyframes}
\label{app:interpretability}

For a well-trained selector, the \textit{mnemonics} buffer $\mathcal{B}_t$ should accumulate keyframes that are information-critical for the task and provide necessary context for the downstream BC policy to generate actions unambiguously. Figure~\ref{fig:mnemonics} shows the buffer contents for our visual benchmark domains. We observe that the buffer keyframes qualitatively capture the belief state for each task by retaining observations with task-relevant cues. The discovery process is fully self-supervised, as the selector is trained only from expert demonstration state-action pairs without any additional temporal keyframe annotation. These qualitative results highlight the interpretability of Keyframe Mnemonics: the buffer contains prior observations that can be inspected to understand what the BC policy relies on when taking a particular action in a partially observable environment.

\subsection{Selector priority behavior}
\label{app:selector-priority}
\begin{figure*}[t]
  \centering
  \includegraphics[width=\textwidth]{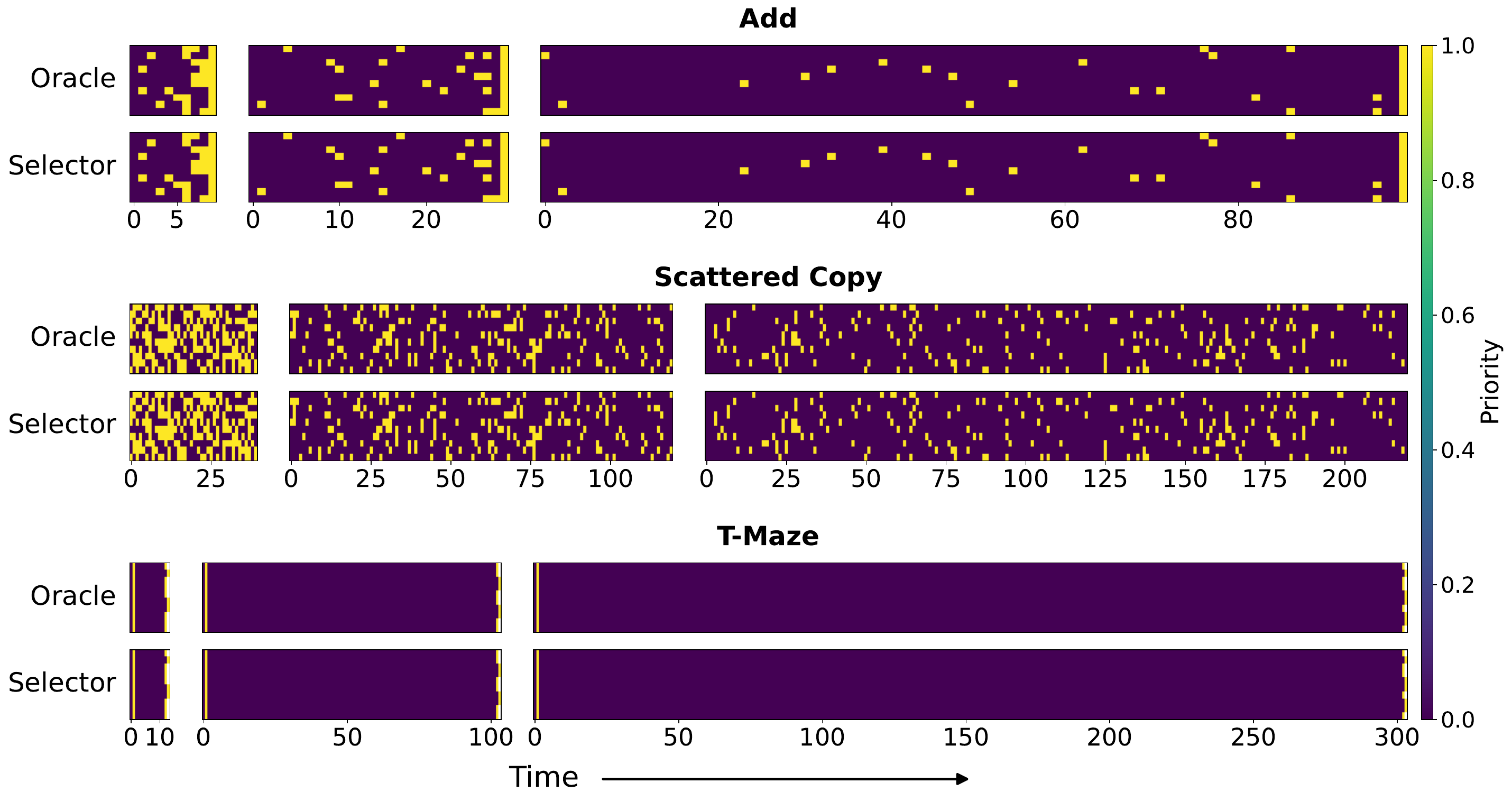}
  \caption{%
    \textbf{Trained selector recovers the keyframe oracle as a near-optimal function.}
Per-step selector priorities (bottom row of each pair) are compared against the oracle keyframe indicator (top row of each pair) on three synthetic memory benchmarks. Each benchmark is evaluated at three increasing episode horizons, where the longest horizon extends one order of magnitude beyond the training horizon. For each panel, we render $10$ evaluation episodes (rows) over time (columns), where each column corresponds to a single environment step and color encodes either the selector output ($p_t \in [0,1]$) or the binary oracle indicator ($1$ for keyframes and $0$ otherwise). The oracle definition is problem-specific, as discussed in Appendix~\ref{app:synth-env}. Across all cells, the selector priority agrees with the oracle. After squashing the raw selector outputs to $[0,1]$, the mean selector priority is exactly $1.00$ on keyframes and $0.00$ on background frames. This empirical result shows that the selector policy $\pi_{\mathrm{s}}$ learns a near-optimal keyframe selection function for domains where our assumptions hold strongly.}
  \label{fig:priority-timeline}
\end{figure*}

In Figure~\ref{fig:priority-timeline} we evaluate trained selector policy $\pi_{\mathrm{s}}$ across synthetic domains to qualitatively assess how well the learned priorities align with the oracle keyframes defined in Appendix~\ref{app:synth-env}. We find that the selector learns a near-optimal selection function that agrees with the oracle, highlighting that the proxy objective provides a strong signal for keyframe discovery, consistent with Corollary~1. During Stage~2 RL training (Algorithm~\ref{alg:selector-env}), the selector explores priority assignments and learns to commit high priorities to decision-relevant keyframes. This qualitatively explains the horizon-invariant results in Table~\ref{tab:horizon-scaling}, where BC policies achieve perfect performance over horizons orders of magnitude beyond training. Since the selector is Markovian and depends only on the current observation, its behavior remains invariant to horizon length, allowing it to admit only informative keyframes into the buffer at inference time.

\subsection{BC Policy activation map}
\label{app:activation-map}
\begin{figure}[t]
    \centering
    \includegraphics[width=0.8\linewidth]{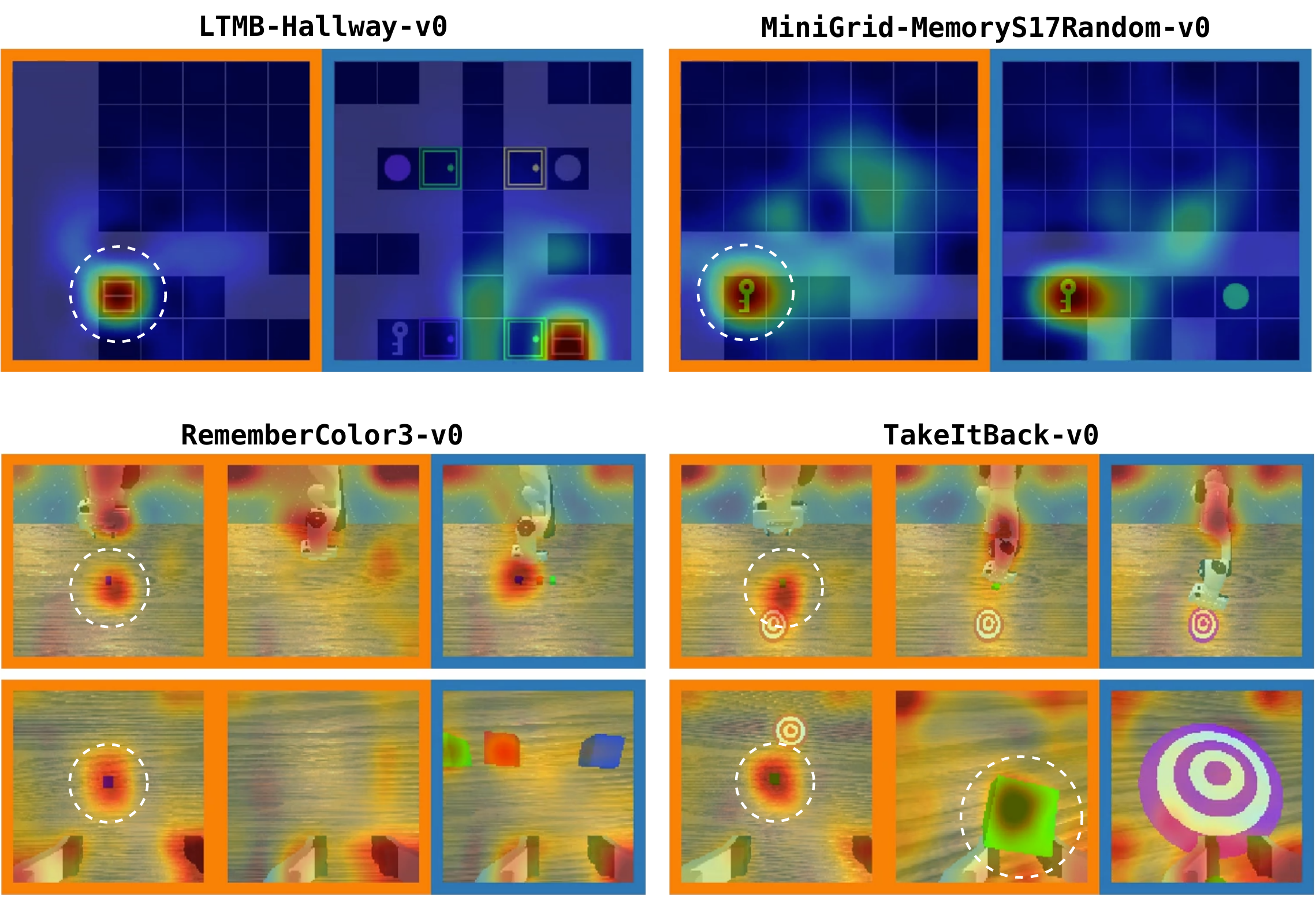}
\caption{\textbf{BC policy activation map on mnemonics.} Grad-CAM~\cite{selvaraju2020grad, chattopadhay2018grad} activation maps of the behavior cloning policy $\pi_{\mathrm{BC}}$ on buffer images at decision-critical steps, where hot regions indicate inputs driving the predicted action. Frames with \swatch{orange!80} boundaries represent discovered keyframes, while the frame with a \swatch{NavyBlue!80} boundary shows the agent's current observation. Across both grid and high-dimensional visual domains, the BC policy uses task-specific cue information from the \textit{discovered} keyframes to generate actions. This highlights the utility of mnemonics for disambiguating policy actions under partial observability. See \url{https://keyframe-mnemonics.github.io} for activation map videos over policy rollouts.}
\label{fig:activation_map}
\end{figure}

To understand whether the discovered keyframes are useful to the behavior cloning policy, we use Grad-CAM~\cite{selvaraju2020grad, chattopadhay2018grad}, an explainable AI technique commonly used to visualize which parts of the input a policy focuses on when generating actions. In Figure~\ref{fig:activation_map}, we visualize activation maps at decision-critical timesteps in grid and robot manipulation domains, where the BC policy must rely on memory-buffer information to generate the correct action. We observe that the BC policy attends to information cues from both the buffer and the current observation to infer the POMDP belief state and act correctly. In the grid domain, the heatmap is concentrated around the matching object cue. In \texttt{RememberColor3}, activations concentrate around the blue colored cube in both the keyframe and the current observation. In \texttt{TakeItBack}, activations appear around the initial cube location in the first keyframe, while the second keyframe captures temporal task-progress information just before the robot begins pushing the cube, helping disambiguate the state so that the policy pulls the cube back to its initial location. These results show that the keyframes discovered by our method are decision-relevant and provide necessary context for BC policies to disambiguate actions under partial observability, even in high-dimensional robot visual domains.

\subsection{Regret on low memory budget \texorpdfstring{$k$}{k}}
\label{app:memory-budget}

\begin{figure}[t]
    \centering
    \includegraphics[width=0.8\linewidth]{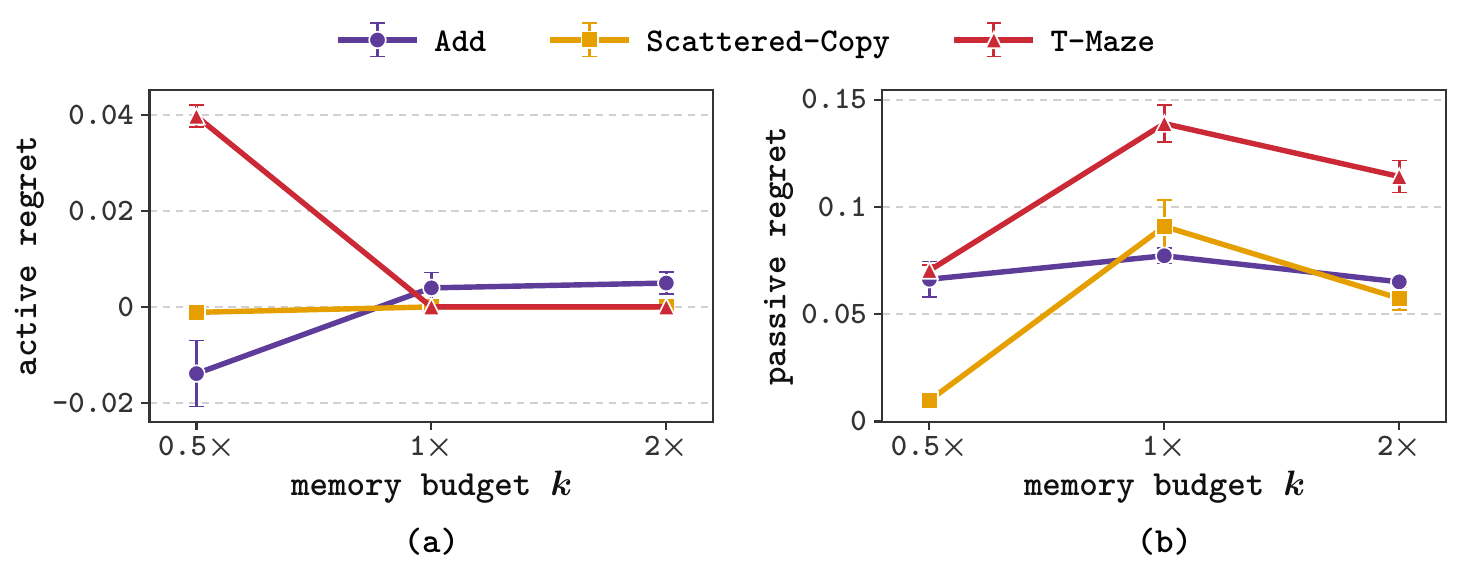}
    \caption{\textbf{Memory regret as a function of buffer capacity $k$.} The $1\times$ setting corresponds to the proxy training memory budget $k$ in Table~\ref{tab:km-arch}. \textbf{(a)} Active regret decreases as buffer capacity becomes sufficient to retain the true keyframes, with domain-specific behavior when $k$ is below the true keyframe count. \textbf{(b)} Passive regret remains low under insufficient capacity and decreases again for larger buffers, where random selection is more likely to retain keyframes.}
    \label{fig:cross-buffer}
    \vspace{-5pt}
\end{figure}

Figure~\ref{fig:cross-buffer} analyzes how memory regret (Section~\ref{sec:empirical_analysis}) changes with buffer capacity $k$ for synthetic domains. In T-Maze, active regret remains high when $k=1$ ($\lceil 0.5 \times 2 \rceil$), which is below the true number of keyframes ($2$), because the buffer cannot retain all required keyframes. In Add, active regret becomes negative when $k=2$ ($\lceil 0.5 \times 3 \rceil$), also below the true keyframe count ($3$). This occurs because the selector prioritizes the two marked numbers over the emit step, whereas the oracle assigns equal priority to all three keyframes. Under the low memory budget, the oracle evicts the first marked step, leading to higher proxy loss. It shows that the selector can learn a non-uniform priority policy that favors keyframes with higher mutual information with the action. For Scattered-Copy, active regret remains close to zero since the oracle and selector perform similarly across memory budgets. Passive regret remains low when the buffer size is smaller than the true keyframe count because the proxy cannot converge to an invariant hypothesis under constrained memory. At the $2\times$ buffer size, passive regret is lower than at $1\times$, since a larger buffer increases the probability that a random selector retains keyframes.

\subsection{PPO reward plots}
\label{app:reward-plots}

\begin{figure}[t]
\centering
\includegraphics[width=0.85\linewidth]{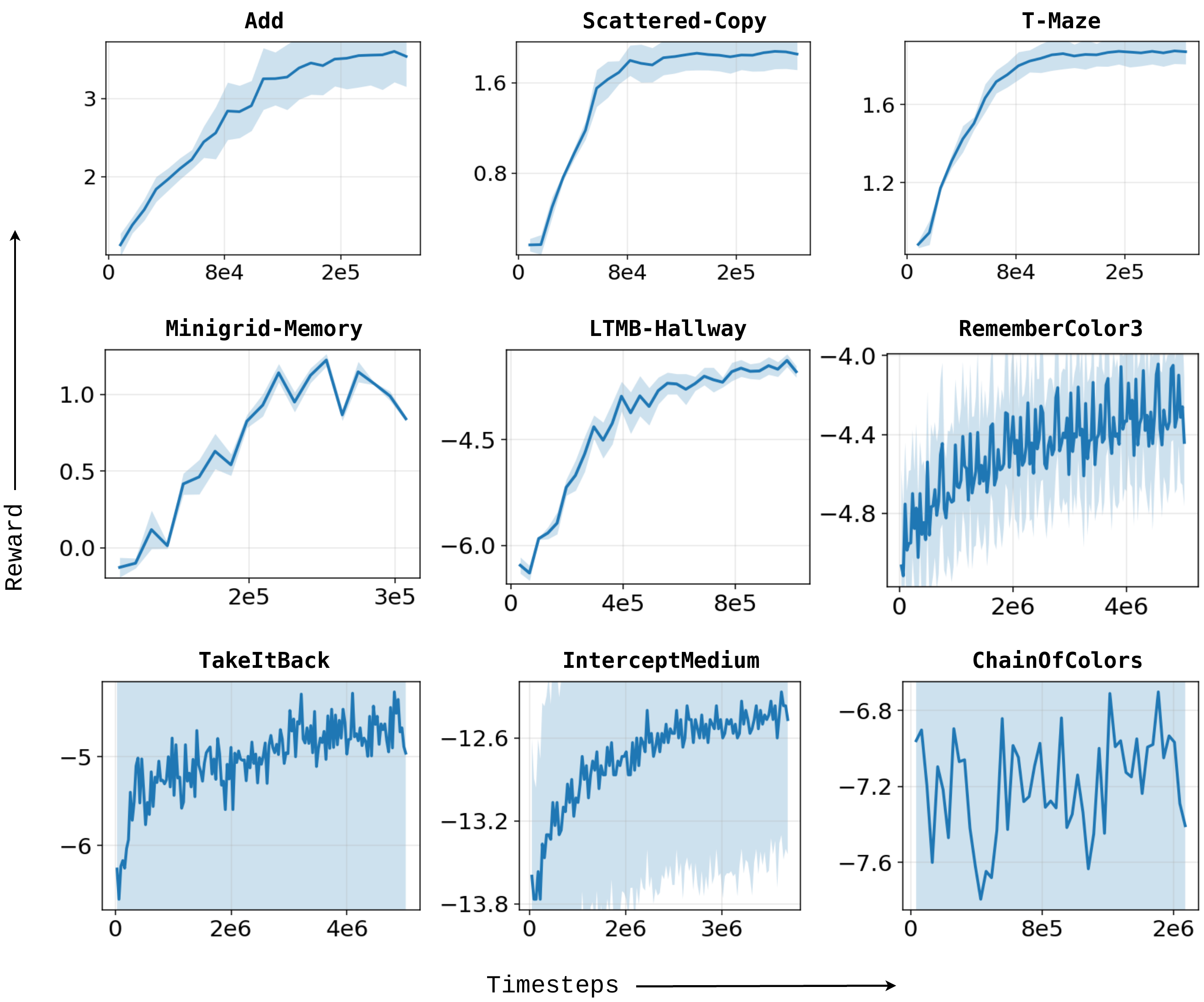}
\caption{\textbf{Selector training PPO learning curves.}
The selector policy converges within $200$K environment steps for synthetic domains and $5$M steps for robot manipulation domains. For tasks where optimization is difficult, e.g., \texttt{ChainOfColors}, the learning curve is noisy and shows modest reward improvement, indicating that some of our assumptions may not hold strongly in the environment.}
\label{fig:reward-plot}
\end{figure}

Figure~\ref{fig:reward-plot} shows the learning curves for training the selector policy $\pi_{\mathrm{s}}$ on the frozen proxy objective (Section~\ref{sec:selector_optimization}) using PPO~\cite{schulman2017proximal}. For synthetic domains where our assumptions hold strongly, we observe clear reward improvement, with the selector converging within $200$K environment (Algorithm~\ref{alg:selector-env}) steps. For robot manipulation domains, the selector converges within $5$M steps. The plot for \texttt{ChainOfColors} illustrates a task that is more difficult to optimize, exhibiting a noisy learning curve. We observed that such noise typically appears when the proxy policy does not converge properly, either because it is undertrained or because our assumptions do not hold well in the environment.

\subsection{Limitation discussion}
\label{appendix:extended_limitations}
Our Assumption~1 is restrictive, as it requires keyframes to be individually identifiable from a single observation because the selector policy $\pi_\mathrm{s}(p_t \mid o_t)$ is Markovian. This formulation allows selector training to scale without recurrent modules, thereby avoiding backpropagation through time. Hence, our method is best suited for sparse, frame-localizable memory tasks. However, problems that do not satisfy the assumption may require multiple keyframes to represent the same underlying information, leading to multiple information-equivalent buffer states that achieve similar objective values under the proxy policy. In such cases, it becomes challenging to retain only one mnemonic set among several equivalent alternatives.

Our Assumption~2 requires action ambiguity to be strongly correlated with a sparse set of keyframes. In other words, the action distribution must exhibit sufficient multimodality that can be disambiguated by the remembered keyframes. In tasks where actions are mostly local and exhibit low variance, actions may carry little mutual information with the buffer observations. In such a regime, the mutual information gap in Theorem~1 becomes negligible, and the proxy policy may fail to converge because the invariant hypothesis is difficult to learn from a sparse memory buffer. We observed this failure mode on the VizDoom-Two-Colors~\cite{sorokin2022explain} task, where the agent observes a color cue and must later choose the corresponding target, but the discrete navigation actions remain largely local.

We found that the proxy policy struggles to converge to the true invariant hypothesis when expert actions at decision-critical timesteps depend simultaneously on multiple unique keyframes. This occurs because the probability that masked-sampling selects all relevant keyframes during proxy training decreases as the number of action-relevant keyframes increases, making it difficult to learn tasks requiring high memory capacity. A similar issue arises in very long demonstrations with few keyframes, where keyframe sparsity further reduces the probability of sampling the relevant frames. A straightforward way to address this issue is to increase the training memory budget, which makes sampling keyframes into the buffer more likely and also increases the support size $M$, thereby enforcing Theorem~1 more effectively. Another promising direction is to train proxy policies on shorter low-memory-capacity tasks, learn specialized selectors for each, and compose them to support high-memory-capacity BC policies, which we leave for future work.

The PPO training time dominates the computational cost in robot-domain experiments. Our current implementation uses SB3~\cite{stable-baselines3}, which introduces nontrivial CPU-GPU overhead and becomes the main bottleneck. The selector training environment (Algorithm~4) primarily requires proxy policy inference, which can in principle be executed entirely on the GPU. In future work, we plan to use a GPU-native PPO implementation to accelerate selector training.

\section{Extended Related Works}
\label{appendix:ext-rel-works}
\textbf{Language as a proxy for state tracking.} Recent methods have used segment-level language-annotated demonstrations to finetune vision-language models (VLMs) that generate autoregressive subgoal commands for policy steering \cite{sridhar2025memer, torne2026mem}. While language provides high-level semantic guidance, it lacks the spatio-temporal detail present in visual observations. In contrast, our approach does not strictly require language annotations for training. Instead, we view Keyframe Mnemonics as complementary to such methods, with informative keyframes addressing partial observability and language supporting abstract reasoning over past events.

\textbf{Masking-induced invariant feature learning.} Prior works have used masking in self-supervised learning over language tokens~\cite{devlin2019bert} and spatial image patches~\cite{he2022masked} to learn invariant features from input data. Our keyframe discovery process instead applies masking over the temporal history of individual observations. We hypothesize that this induces a similar regularization effect: random input sampling increases the difficulty of learning complex, non-invariant features, thereby biasing the model toward invariant hypotheses that better capture true correlations over spurious ones.

\textbf{Instance-wise feature selection.} Prior work learns to select task-relevant input features with a selector–predictor architecture, using a Gumbel-softmax relaxation to sample a fixed number of features~\cite{chen2018learning} or an actor-critic objective against a full-feature baseline~\cite{yoon2018invase}. ATK~\cite{zhang2025atk} applies a Gumbel-softmax mask with an $L1$ penalty to select task-relevant keypoints for robot policies. These methods select from the current input, assuming it suffices for decision-making, and do not transfer to non-Markovian settings where selection over the full history is intractable. Keyframe Mnemonics instead uses masked sampling over history subsets bounded by the memory budget $k$.

\clearpage

%% file: checklist.tex
\section*{NeurIPS Paper Checklist}

\begin{enumerate}

\item {\bf Claims}
    \item[] Question: Do the main claims made in the abstract and introduction accurately reflect the paper's contributions and scope?
    \item[] Answer: \answerYes{} 
    \item[] Justification: We have clearly stated the contributions in the introduction and have ensured that the Methodology, and the Experiments and Results Analysis sections back up the three contributions clearly.  
    \item[] Guidelines:
    \begin{itemize}
        \item The answer \answerNA{} means that the abstract and introduction do not include the claims made in the paper.
        \item The abstract and/or introduction should clearly state the claims made, including the contributions made in the paper and important assumptions and limitations. A \answerNo{} or \answerNA{} answer to this question will not be perceived well by the reviewers. 
        \item The claims made should match theoretical and experimental results, and reflect how much the results can be expected to generalize to other settings. 
        \item It is fine to include aspirational goals as motivation as long as it is clear that these goals are not attained by the paper. 
    \end{itemize}

\item {\bf Limitations}
    \item[] Question: Does the paper discuss the limitations of the work performed by the authors?
    \item[] Answer: \answerYes{} 
    \item[] Justification: The paper has a limitations section where we have stated clearly the limitations of the current work. 
    \item[] Guidelines:
    \begin{itemize}
        \item The answer \answerNA{} means that the paper has no limitation while the answer \answerNo{} means that the paper has limitations, but those are not discussed in the paper. 
        \item The authors are encouraged to create a separate ``Limitations'' section in their paper.
        \item The paper should point out any strong assumptions and how robust the results are to violations of these assumptions (e.g., independence assumptions, noiseless settings, model well-specification, asymptotic approximations only holding locally). The authors should reflect on how these assumptions might be violated in practice and what the implications would be.
        \item The authors should reflect on the scope of the claims made, e.g., if the approach was only tested on a few datasets or with a few runs. In general, empirical results often depend on implicit assumptions, which should be articulated.
        \item The authors should reflect on the factors that influence the performance of the approach. For example, a facial recognition algorithm may perform poorly when image resolution is low or images are taken in low lighting. Or a speech-to-text system might not be used reliably to provide closed captions for online lectures because it fails to handle technical jargon.
        \item The authors should discuss the computational efficiency of the proposed algorithms and how they scale with dataset size.
        \item If applicable, the authors should discuss possible limitations of their approach to address problems of privacy and fairness.
        \item While the authors might fear that complete honesty about limitations might be used by reviewers as grounds for rejection, a worse outcome might be that reviewers discover limitations that aren't acknowledged in the paper. The authors should use their best judgment and recognize that individual actions in favor of transparency play an important role in developing norms that preserve the integrity of the community. Reviewers will be specifically instructed to not penalize honesty concerning limitations.
    \end{itemize}

\item {\bf Theory assumptions and proofs}
    \item[] Question: For each theoretical result, does the paper provide the full set of assumptions and a complete (and correct) proof?
    \item[] Answer: \answerYes{} 
    \item[] Justification: In the methodology section and the appendix we have attached mathematical proofs of our claims which are correct in our opinion. 
    \item[] Guidelines:
    \begin{itemize}
        \item The answer \answerNA{} means that the paper does not include theoretical results. 
        \item All the theorems, formulas, and proofs in the paper should be numbered and cross-referenced.
        \item All assumptions should be clearly stated or referenced in the statement of any theorems.
        \item The proofs can either appear in the main paper or the supplemental material, but if they appear in the supplemental material, the authors are encouraged to provide a short proof sketch to provide intuition. 
        \item Inversely, any informal proof provided in the core of the paper should be complemented by formal proofs provided in appendix or supplemental material.
        \item Theorems and Lemmas that the proof relies upon should be properly referenced. 
    \end{itemize}

    \item {\bf Experimental result reproducibility}
    \item[] Question: Does the paper fully disclose all the information needed to reproduce the main experimental results of the paper to the extent that it affects the main claims and/or conclusions of the paper (regardless of whether the code and data are provided or not)?
    \item[] Answer: \answerYes{} 
    \item[] Justification: We are submitting self contained code to run our framework on two of the domains demonstrated in our results. We have aimed to provide parameters for our architectures where appropriate. We have not collected any data as part of this paper. 
    \item[] Guidelines:
    \begin{itemize}
        \item The answer \answerNA{} means that the paper does not include experiments.
        \item If the paper includes experiments, a \answerNo{} answer to this question will not be perceived well by the reviewers: Making the paper reproducible is important, regardless of whether the code and data are provided or not.
        \item If the contribution is a dataset and\slash or model, the authors should describe the steps taken to make their results reproducible or verifiable. 
        \item Depending on the contribution, reproducibility can be accomplished in various ways. For example, if the contribution is a novel architecture, describing the architecture fully might suffice, or if the contribution is a specific model and empirical evaluation, it may be necessary to either make it possible for others to replicate the model with the same dataset, or provide access to the model. In general. releasing code and data is often one good way to accomplish this, but reproducibility can also be provided via detailed instructions for how to replicate the results, access to a hosted model (e.g., in the case of a large language model), releasing of a model checkpoint, or other means that are appropriate to the research performed.
        \item While NeurIPS does not require releasing code, the conference does require all submissions to provide some reasonable avenue for reproducibility, which may depend on the nature of the contribution. For example
        \begin{enumerate}
            \item If the contribution is primarily a new algorithm, the paper should make it clear how to reproduce that algorithm.
            \item If the contribution is primarily a new model architecture, the paper should describe the architecture clearly and fully.
            \item If the contribution is a new model (e.g., a large language model), then there should either be a way to access this model for reproducing the results or a way to reproduce the model (e.g., with an open-source dataset or instructions for how to construct the dataset).
            \item We recognize that reproducibility may be tricky in some cases, in which case authors are welcome to describe the particular way they provide for reproducibility. In the case of closed-source models, it may be that access to the model is limited in some way (e.g., to registered users), but it should be possible for other researchers to have some path to reproducing or verifying the results.
        \end{enumerate}
    \end{itemize}

\item {\bf Open access to data and code}
    \item[] Question: Does the paper provide open access to the data and code, with sufficient instructions to faithfully reproduce the main experimental results, as described in supplemental material?
    \item[] Answer: \answerYes{} 
    \item[] Justification: We have released two self contained domains with our architecture that are runnable. This should allow the reviewers to verify our claims. We have also attached our webpage with more information and access to github.
    \item[] Guidelines:
    \begin{itemize}
        \item The answer \answerNA{} means that paper does not include experiments requiring code.
        \item Please see the NeurIPS code and data submission guidelines (\url{https://neurips.cc/public/guides/CodeSubmissionPolicy}) for more details.
        \item While we encourage the release of code and data, we understand that this might not be possible, so \answerNo{} is an acceptable answer. Papers cannot be rejected simply for not including code, unless this is central to the contribution (e.g., for a new open-source benchmark).
        \item The instructions should contain the exact command and environment needed to run to reproduce the results. See the NeurIPS code and data submission guidelines (\url{https://neurips.cc/public/guides/CodeSubmissionPolicy}) for more details.
        \item The authors should provide instructions on data access and preparation, including how to access the raw data, preprocessed data, intermediate data, and generated data, etc.
        \item The authors should provide scripts to reproduce all experimental results for the new proposed method and baselines. If only a subset of experiments are reproducible, they should state which ones are omitted from the script and why.
        \item At submission time, to preserve anonymity, the authors should release anonymized versions (if applicable).
        \item Providing as much information as possible in supplemental material (appended to the paper) is recommended, but including URLs to data and code is permitted.
    \end{itemize}

\item {\bf Experimental setting/details}
    \item[] Question: Does the paper specify all the training and test details (e.g., data splits, hyperparameters, how they were chosen, type of optimizer) necessary to understand the results?
    \item[] Answer: \answerYes{} 
    \item[] Justification: The results and Appendix B aim to provide complete experimental details for reproducibility. 
    \item[] Guidelines:
    \begin{itemize}
        \item The answer \answerNA{} means that the paper does not include experiments.
        \item The experimental setting should be presented in the core of the paper to a level of detail that is necessary to appreciate the results and make sense of them.
        \item The full details can be provided either with the code, in appendix, or as supplemental material.
    \end{itemize}

\item {\bf Experiment statistical significance}
    \item[] Question: Does the paper report error bars suitably and correctly defined or other appropriate information about the statistical significance of the experiments?
    \item[] Answer: \answerYes{} 
    \item[] Justification: The error bars are visible in each of the figures and claims are made with respect to these error bars. 
    \item[] Guidelines:
    \begin{itemize}
        \item The answer \answerNA{} means that the paper does not include experiments.
        \item The authors should answer \answerYes{} if the results are accompanied by error bars, confidence intervals, or statistical significance tests, at least for the experiments that support the main claims of the paper.
        \item The factors of variability that the error bars are capturing should be clearly stated (for example, train/test split, initialization, random drawing of some parameter, or overall run with given experimental conditions).
        \item The method for calculating the error bars should be explained (closed form formula, call to a library function, bootstrap, etc.)
        \item The assumptions made should be given (e.g., Normally distributed errors).
        \item It should be clear whether the error bar is the standard deviation or the standard error of the mean.
        \item It is OK to report 1-sigma error bars, but one should state it. The authors should preferably report a 2-sigma error bar than state that they have a 96\% CI, if the hypothesis of Normality of errors is not verified.
        \item For asymmetric distributions, the authors should be careful not to show in tables or figures symmetric error bars that would yield results that are out of range (e.g., negative error rates).
        \item If error bars are reported in tables or plots, the authors should explain in the text how they were calculated and reference the corresponding figures or tables in the text.
    \end{itemize}

\item {\bf Experiments compute resources}
    \item[] Question: For each experiment, does the paper provide sufficient information on the computer resources (type of compute workers, memory, time of execution) needed to reproduce the experiments?
    \item[] Answer: \answerYes{} 
    \item[] Justification:  Appendix~\ref{appendix:toy_implementation} and  \ref{appendix:robot_implementation} specified compute times for our methods and computational resources used. 
    \item[] Guidelines:
    \begin{itemize}
        \item The answer \answerNA{} means that the paper does not include experiments.
        \item The paper should indicate the type of compute workers CPU or GPU, internal cluster, or cloud provider, including relevant memory and storage.
        \item The paper should provide the amount of compute required for each of the individual experimental runs as well as estimate the total compute. 
        \item The paper should disclose whether the full research project required more compute than the experiments reported in the paper (e.g., preliminary or failed experiments that didn't make it into the paper). 
    \end{itemize}
    
\item {\bf Code of ethics}
    \item[] Question: Does the research conducted in the paper conform, in every respect, with the NeurIPS Code of Ethics \url{https://neurips.cc/public/EthicsGuidelines}?
    \item[] Answer: \answerYes{} 
    \item[] Justification: This work has not collected any human data. We have listed computational costs and released runnable code to demonstrate reproducibility. Our work is in introductory stages so does not have comparable societal impact to existing generative AI technologies as of now. The goal is to improve memory of neural networks to eventually perform long horizon tasks on robots which would be more directly useful to our society.
    \item[] Guidelines:
    \begin{itemize}
        \item The answer \answerNA{} means that the authors have not reviewed the NeurIPS Code of Ethics.
        \item If the authors answer \answerNo, they should explain the special circumstances that require a deviation from the Code of Ethics.
        \item The authors should make sure to preserve anonymity (e.g., if there is a special consideration due to laws or regulations in their jurisdiction).
    \end{itemize}

\item {\bf Broader impacts}
    \item[] Question: Does the paper discuss both potential positive societal impacts and negative societal impacts of the work performed?
    \item[] Answer: \answerNA{} 
    \item[] Justification: The work on improving memory for neural networks is not practical yet for direct societal interaction. This work is on artificial benchmark and simulated robot tasks which is not meaningful to society at large in its current state. 
    \item[] Guidelines:
    \begin{itemize}
        \item The answer \answerNA{} means that there is no societal impact of the work performed.
        \item If the authors answer \answerNA{} or \answerNo, they should explain why their work has no societal impact or why the paper does not address societal impact.
        \item Examples of negative societal impacts include potential malicious or unintended uses (e.g., disinformation, generating fake profiles, surveillance), fairness considerations (e.g., deployment of technologies that could make decisions that unfairly impact specific groups), privacy considerations, and security considerations.
        \item The conference expects that many papers will be foundational research and not tied to particular applications, let alone deployments. However, if there is a direct path to any negative applications, the authors should point it out. For example, it is legitimate to point out that an improvement in the quality of generative models could be used to generate Deepfakes for disinformation. On the other hand, it is not needed to point out that a generic algorithm for optimizing neural networks could enable people to train models that generate Deepfakes faster.
        \item The authors should consider possible harms that could arise when the technology is being used as intended and functioning correctly, harms that could arise when the technology is being used as intended but gives incorrect results, and harms following from (intentional or unintentional) misuse of the technology.
        \item If there are negative societal impacts, the authors could also discuss possible mitigation strategies (e.g., gated release of models, providing defenses in addition to attacks, mechanisms for monitoring misuse, mechanisms to monitor how a system learns from feedback over time, improving the efficiency and accessibility of ML).
    \end{itemize}
    
\item {\bf Safeguards}
    \item[] Question: Does the paper describe safeguards that have been put in place for responsible release of data or models that have a high risk for misuse (e.g., pre-trained language models, image generators, or scraped datasets)?
    \item[] Answer: \answerNA{} 
    \item[] Justification: The work is too preliminary to have a high risk of misuse in a societal scale.
    \item[] Guidelines:
    \begin{itemize}
        \item The answer \answerNA{} means that the paper poses no such risks.
        \item Released models that have a high risk for misuse or dual-use should be released with necessary safeguards to allow for controlled use of the model, for example by requiring that users adhere to usage guidelines or restrictions to access the model or implementing safety filters. 
        \item Datasets that have been scraped from the Internet could pose safety risks. The authors should describe how they avoided releasing unsafe images.
        \item We recognize that providing effective safeguards is challenging, and many papers do not require this, but we encourage authors to take this into account and make a best faith effort.
    \end{itemize}

\item {\bf Licenses for existing assets}
    \item[] Question: Are the creators or original owners of assets (e.g., code, data, models), used in the paper, properly credited and are the license and terms of use explicitly mentioned and properly respected?
    \item[] Answer: \answerYes{} 
    \item[] Justification: We have cited the papers whose code and/or data we have used to run our experiments. All other code is our own. 
    \item[] Guidelines:
    \begin{itemize}
        \item The answer \answerNA{} means that the paper does not use existing assets.
        \item The authors should cite the original paper that produced the code package or dataset.
        \item The authors should state which version of the asset is used and, if possible, include a URL.
        \item The name of the license (e.g., CC-BY 4.0) should be included for each asset.
        \item For scraped data from a particular source (e.g., website), the copyright and terms of service of that source should be provided.
        \item If assets are released, the license, copyright information, and terms of use in the package should be provided. For popular datasets, \url{paperswithcode.com/datasets} has curated licenses for some datasets. Their licensing guide can help determine the license of a dataset.
        \item For existing datasets that are re-packaged, both the original license and the license of the derived asset (if it has changed) should be provided.
        \item If this information is not available online, the authors are encouraged to reach out to the asset's creators.
    \end{itemize}

\item {\bf New assets}
    \item[] Question: Are new assets introduced in the paper well documented and is the documentation provided alongside the assets?
    \item[] Answer: \answerYes{} 
    \item[] Justification: We are releasing code as part of the supplementary material and on our webpage. No human subjects data was collected or used as part of this project. 
    \item[] Guidelines:
    \begin{itemize}
        \item The answer \answerNA{} means that the paper does not release new assets.
        \item Researchers should communicate the details of the dataset\slash code\slash model as part of their submissions via structured templates. This includes details about training, license, limitations, etc. 
        \item The paper should discuss whether and how consent was obtained from people whose asset is used.
        \item At submission time, remember to anonymize your assets (if applicable). You can either create an anonymized URL or include an anonymized zip file.
    \end{itemize}

\item {\bf Crowdsourcing and research with human subjects}
    \item[] Question: For crowdsourcing experiments and research with human subjects, does the paper include the full text of instructions given to participants and screenshots, if applicable, as well as details about compensation (if any)? 
    \item[] Answer: \answerNA{} 
    \item[] Justification: We did not collect any human subjects data. 
    \item[] Guidelines:
    \begin{itemize}
        \item The answer \answerNA{} means that the paper does not involve crowdsourcing nor research with human subjects.
        \item Including this information in the supplemental material is fine, but if the main contribution of the paper involves human subjects, then as much detail as possible should be included in the main paper. 
        \item According to the NeurIPS Code of Ethics, workers involved in data collection, curation, or other labor should be paid at least the minimum wage in the country of the data collector. 
    \end{itemize}

\item {\bf Institutional review board (IRB) approvals or equivalent for research with human subjects}
    \item[] Question: Does the paper describe potential risks incurred by study participants, whether such risks were disclosed to the subjects, and whether Institutional Review Board (IRB) approvals (or an equivalent approval/review based on the requirements of your country or institution) were obtained?
    \item[] Answer: \answerNA{} 
    \item[] Justification: We did not collect any human subjects data.
    \item[] Guidelines:
    \begin{itemize}
        \item The answer \answerNA{} means that the paper does not involve crowdsourcing nor research with human subjects.
        \item Depending on the country in which research is conducted, IRB approval (or equivalent) may be required for any human subjects research. If you obtained IRB approval, you should clearly state this in the paper. 
        \item We recognize that the procedures for this may vary significantly between institutions and locations, and we expect authors to adhere to the NeurIPS Code of Ethics and the guidelines for their institution. 
        \item For initial submissions, do not include any information that would break anonymity (if applicable), such as the institution conducting the review.
    \end{itemize}

\item {\bf Declaration of LLM usage}
    \item[] Question: Does the paper describe the usage of LLMs if it is an important, original, or non-standard component of the core methods in this research? Note that if the LLM is used only for writing, editing, or formatting purposes and does \emph{not} impact the core methodology, scientific rigor, or originality of the research, declaration is not required.
    \item[] Answer: \answerNA{} 
    \item[] Justification: LLMs were not used to develop the core idea or experimental design of this research. 
    \item[] Guidelines:
    \begin{itemize}
        \item The answer \answerNA{} means that the core method development in this research does not involve LLMs as any important, original, or non-standard components.
        \item Please refer to our LLM policy in the NeurIPS handbook for what should or should not be described.
    \end{itemize}

\end{enumerate}